\documentclass[man,floatsintext,oneside]{apa7}
\usepackage{graphicx}
\usepackage{adjustbox}
\usepackage{mathtools}
\usepackage{amssymb}
\usepackage{subcaption}
\usepackage[american]{babel}
\usepackage{csquotes}
\usepackage{needspace}
\usepackage[style=apa,backend=biber,natbib=true,sortcites=true,sorting=nyt]{biblatex}
\usepackage[section]{placeins}
\usepackage{appendix}
\usepackage{listings}
\usepackage{xcolor}
\usepackage{hyperref}
\usepackage{caption}
\usepackage{float}

\definecolor{codeblue}{rgb}{0.10,0.10,0.55}
\definecolor{codegreen}{rgb}{0.00,0.40,0.00}
\definecolor{codegray}{gray}{0.45}

\lstdefinestyle{paper}{
  language=Python,
  basicstyle=\ttfamily\small,
  keywordstyle=\bfseries\color{codeblue},
  commentstyle=\itshape\color{codegray},
  stringstyle=\color{codegreen},
  numbers=left,
  numberstyle=\tiny\color{codegray},
  frame=single,
  breaklines=true,
  columns=fullflexible,
  showstringspaces=false,
  xleftmargin=2em,
  aboveskip=1em,
  belowskip=1em
}

\DeclareLanguageMapping{american}{american-apa}
\title{How to predict creativity ratings from written narratives: A comparison of co-occurrence and textual forma mentis networks} 
\shorttitle{Predicting creativity ratings}
\leftheader{Passaro}

\authorsnames{Roberto Passaro\textsuperscript{1,2}, Edith Haim\textsuperscript{2}, Massimo Stella\textsuperscript{2}}
\authorsaffiliations{
  \textsuperscript{1}Center for Mind/Brain Sciences (CIMeC), University of Trento\\
  \textsuperscript{2}CogNosco Lab, Department of Psychology and Cognitive Science, University of Trento
}

\abstract{This tutorial paper provides a step-by-step workflow for building and analysing semantic networks from short creative texts. We introduce and compare two widely used text-to-network approaches: word co-occurrence networks and textual forma mentis networks (TFMNs). We also demonstrate how they can be used in machine learning to predict human creativity ratings.
Using a corpus of 1029 short stories, we guide readers through text preprocessing, network construction, feature extraction (structural measures, spreading-activation indices, and emotion scores), and application of regression models. We evaluate how network-construction choices influence both network topology and predictive performance. Across all modelling settings, TFMNs consistently outperformed co-occurrence networks through lower prediction errors (best MAE = 0.581 for TFMN, vs 0.592 for co-occurrence with window size 3). Network-structural features dominated predictive performance (MAE = 0.591 for TFMN), whereas emotion features performed worse (MAE = 0.711 for TFMN) and spreading-activation measures contributed little (MAE = 0.788 for TFMN).
This paper offers practical guidance for researchers interested in applying network-based methods for cognitive fields like creativity research. We show when syntactic networks are preferable to surface co-occurrence models, and provide an open, reproducible workflow accessible to newcomers in the field, while also offering deeper methodological insight for experienced researchers.}


\authornote{
Correspondence concerning this article should be addressed to Massimo Stella, massimo.stella-1@unitn.it 
Conflicts of interest: We have no conflicts of interest to declare.
}

\begin{document}
\maketitle

\section{Introduction}

Understanding how humans evaluate creativity in written narratives is a key challenge in cognitive psychology. Short story creativity involves multiple interacting components, such as conceptual richness, flexible recombination of concepts within human memory, and emotional tone \citep{distefano2025automatic,beaty2023associative}. Yet, researchers often lack practical, interpretable tools for quantifying these properties directly from text. This article therefore offers both a tutorial and an empirical evaluation of two widely used approaches for constructing semantic networks from short stories: word co-occurrence networks and textual forma mentis networks.

In terms of creative storytelling, a crucial component of human memory which is mediating and supporting creative achievement is the so-called mental lexicon \citep{aitchison2012words}. It is a large-scale associative structure that supports meaning construction, inference, and generation of any experience representable through language. In the last few years, network science has become a key modelling paradigm for representing this structure, linking computational representations of text to cognitive theories of conceptual organisation \citep{siew2019cognitive,beaty2023associative,stella2024cognitive,cancho2001small}. Cognitive networks derived from text provide a compact and analysable approximation of how concepts co-activate, spread, and structurally organise during human thinking \citep{cancho2001small,stella2020text}. These networks capture statistical regularities in language and have been used to model feelings expressed on social media \citep{joseph2023cognitive,colladon2025social,fronzetti2023forecasting}, writing styles \citep{quispe2021using,amancio2015complex}, and creative storytelling \citep{haim2024forma}.

A dominant approach, in both artificial intelligence and computational linguistics, is the construction of co-occurrence networks, where words are connected if they appear within a fixed-size sliding window \citep{cancho2001small, goni2011semantic, colladon2025social}. Early work showed that simple adjacency statistics extracted from large corpora produce small-world and scale-free lexical networks \citep{cancho2001small,watts1998collective}, reflecting broad organisational principles of human language. These networks underlie classic distributional semantic models, including count-based embeddings \citep{turney2010frequency}, where proximity in co-occurrence space approximates semantic similarity. In Natural Language Processing (NLP), co-occurrence networks have supported a variety of tasks, like: (i) keyword extraction \citep{mihalcea2004textrank}, e.g. identifying pivotal concepts in a dialogue or textual corpus; (ii) topic modelling \citep{misra2011text}, e.g. identifying recurrent ideas or groups of ideas being discussed in a corpus; (iii) text summarisation \citep{erkan2004lexrank}, e.g. producing a short text presenting the most central ideas or elements of a longer text or corpus; and (iv) word-sense induction \citep{widdows2004geometry}, e.g. understanding the meanings encoded in a single word from its associates in a text. Co-occurrence networks have also been used in AI for common sense reasoning \citep{speer2012representing} (e.g. identifying the context in which a word was mentioned) and in cognitive science to build network representations of associations in animal verbal fluency tasks \citep{goni2011semantic} (e.g. identifying semantic similarities between animals recalled in sequences). Despite their versatility, co-occurrence networks face some crucial limitations: edges/links reflect linear adjacency rather than syntactic or semantic roles; their topology varies sharply with window size; and short texts often produce sparse or disconnected networks \citep{biemann2013creating}. Recent studies highlight that while co-occurrence captures local contextual cues, it struggles with compositional structure, negation, and long-range dependency phenomena that are central to human meaning-making and essential for modelling creativity or narrative complexity \citep{stella2020text, mihalcea2011graph}.

To address the structural shortcomings of surface-based models, recent work in cognitive network science has introduced textual forma mentis networks (or TFMNs, where forma mentis means "mindset shape" in Latin;  \cite{stella2020text,semeraro2025emoatlas}). TFMNs represent a newer class of cognitively inspired lexical networks that explicitly integrate syntactic structure, affective information, and negation into a unified network representation. Introduced by Stella \citep{stella2020text}, TFMNs leverage dependency parsing to connect words that lie within a bounded syntactic radius, capturing subject–verb–object relations, adjective–noun attachments, and other grammatical dependencies that co-occurrence models often miss. This design aligns the representation more closely with cognitive theories of the mental lexicon, where conceptual activation spreads along structural and functional relations rather than surface adjacency \citep{collins1975spreading,vitevitch2021cognitive}. TFMNs have been applied across domains in computational social science, education, and AI-driven text analysis: to map misconceptions about gender balance in social media posts \citep{stella2020text}; to track the evolution of public sentiment and scientific understanding during crises such as COVID-19 \citep{semeraro2022emotional}; to analyse traumatic narratives of self-disclosed sexual assaults reported on social media \citep{abramski2024voices}; and to quantify how emotions propagate through social media trends \citep{stella2022cognitive}. The integration of valence, explicit negation handling, and syntactic connectivity in textual forma mentis networks has yielded improvements in tasks requiring fine-grained semantic discrimination and conceptual mapping, such as: (i) detecting subtle framing (e.g. "bed" in 1-star hotel reviews was framed more negatively compared to "bed" in 5-start hotel reviews, see \cite{semeraro2025emoatlas}), (ii) modelling collective sense making (e.g. "woman" being associated with professional jargon about "success" and "career", but also "loneliness" in 10K tweets on the gender gap, see \cite{stella2020text}), and analysing author-level differences in story writing (e.g. more creative stories being richer in positive emotions when written by Large Language Models, see \cite{haim2024forma}). In AI and computational linguistics, TFMNs thus offer a compelling alternative to surface-based models, providing richer structural cues for downstream tasks such as text classification, narrative analysis, and creativity assessment.

Both co-occurrence networks and textual forma mentis networks can be enriched with dynamic signals derived from spreading-activation \citep{siew2019spreadr}, a mechanism rooted in classic cognitive models of semantic memory. In the influential theory of Collins and Loftus \citep{collins1975spreading}, conceptual retrieval is modelled as activation spreading from a set of seed concepts through associative or relational links, decaying with distance, and accumulating along convergent pathways. This process has inspired numerous computational implementations for modelling semantic priming, lexical access, and associative creativity \citep{siew2019spreadr}. Recent work by Citraro and colleagues formalises spreading-activation dynamics for single-layer and multiplex cognitive networks within the SpreadPy framework \citep{citraro2025spreadpy}, enabling controlled diffusion processes in cognitive networks. In our setting, both co-occurrence networks and TFMNs serve as substrates over which activation originating from the prompt words of a creative story propagates until reaching a stationary distribution. The resulting steady-state activation levels quantify how structurally accessible each prompt is within the story-specific lexical organisation. For example, consider a creative story, like those gathered in \cite{johnson2023divergent}, written by participants stimulated with the prompts "violin", "storm" and "memory". A sample narrative might be: “As the storm raged outside, Mira tightened her grip on the violin, hoping its familiar melodies would anchor a memory that kept slipping away.” In a co-occurrence network, "violin" and "memory" may be connected only if they appear within a narrow sliding window, and "storm" may remain relatively distant if the narrative places it at the beginning of a sentence or clause. Spreading-activation in this network will flow primarily along these surface adjacencies, amplifying prompts that happen to appear near one another in text. In contrast, a TFMN will connect "violin" to its adjectival modifiers and to the verb "grip", "storm" to syntactically related actions, and "memory" to predicates expressing its recovery or loss. Dependency-based paths may position "violin" and "memory" closer through shared verbs or modifiers, even if they are not adjacent in the sentence. When spreading-activation is applied to this network, activation may reach the node "memory" more efficiently from "violin" because the network's syntactic structure provides multiple short dependency routes between the concepts. 

The resulting steady-state activation levels quantify how structurally accessible each prompt is within the story-specific lexical organisation. This approach allows dynamic, cognitively motivated features to complement static network measures: in co-occurrence networks, activation primarily traces surface adjacency patterns, while in TFMNs it navigates syntactic and semantically richer pathways. Although spreading-activation does not modify the underlying network, it provides an additional layer of information reflecting how efficiently conceptual material can flow through associative knowledge. This makes spreading activation-based simulations \citep{citraro2025spreadpy} an aspect relevant to recent theories, cf. \cite{kenett2019semantic} and \cite{beaty2023associative}, linking creativity to flexible associative traversal within the mental lexicon.

Creativity assessment offers a compelling testing ground for comparing co-occurrence networks and TFMNs. Creative writing tasks require the integration of distant concepts or emotional cues within semantic constraints - which is naturally observable in the structure of a lexical network \citep{johnson2023divergent,beaty2023associative}. Prior research shows that creative texts often display large but sparsely clustered networks, longer semantic paths, and flexible conceptual jumps \citep{haim2024forma,semeraro2025emoatlas}. Yet, it remains unclear whether surface co-occurrence or syntactically grounded TFMNs better capture these properties, particularly in extremely short stories where structural detail is limited.

\subsection{Manuscript Scope and Research Questions}

This article offers both a tutorial and an empirical evaluation of two methods for building semantic networks from short stories:
(i) word co-occurrence networks, and (ii) textual forma mentis networks. 

Our goal is twofold: (i) to provide researchers with a practical, step-by-step workflow for turning short texts into interpretable semantic networks, and (ii) to assess how these different network-building strategies influence the ability of machine-learning models to predict human creativity ratings.

For the empirical evaluation, we conduct the first systematic, controlled comparison between co-occurrence networks and TFMNs. Using a corpus of over 1,000 stories written under the constraints of three-word prompts, we build TFMN and co-occurrence network representations per story (six co-occurrence variants differing by window size and pronoun handling, plus a TFMN). We compute structural network measures, prompt-seeded spreading-activation features, as well as emotion profiles, and evaluate predictive performance of human creativity ratings across ten machine-learning regressors under cross-validation.  

Our results show that TFMNs consistently achieve higher predictive accuracy than any co-occurrence model. Structural features dominate prediction; emotions yield incremental gains; spreading-activation adds little beyond static topology. These findings support the hypothesis that dependency-based lexical networks provide a cognitively and computationally richer representation of conceptual organisation in short creative texts.

By clarifying when and why dependency-based network models outperform surface co-occurrence, this work offers both methodological guidance and theoretical insight. It equips behavioural researchers with a transparent, reproducible workflow for analysing creativity through text-based semantic networks. In addition, it informs the design of network-based text representations in AI systems that aim to analyse, generate, or assess creative language.
\vspace{1cm}

\begin{figure}[!htbp]
    \centering
    \includegraphics[width=\linewidth,height=0.75\textheight,keepaspectratio]{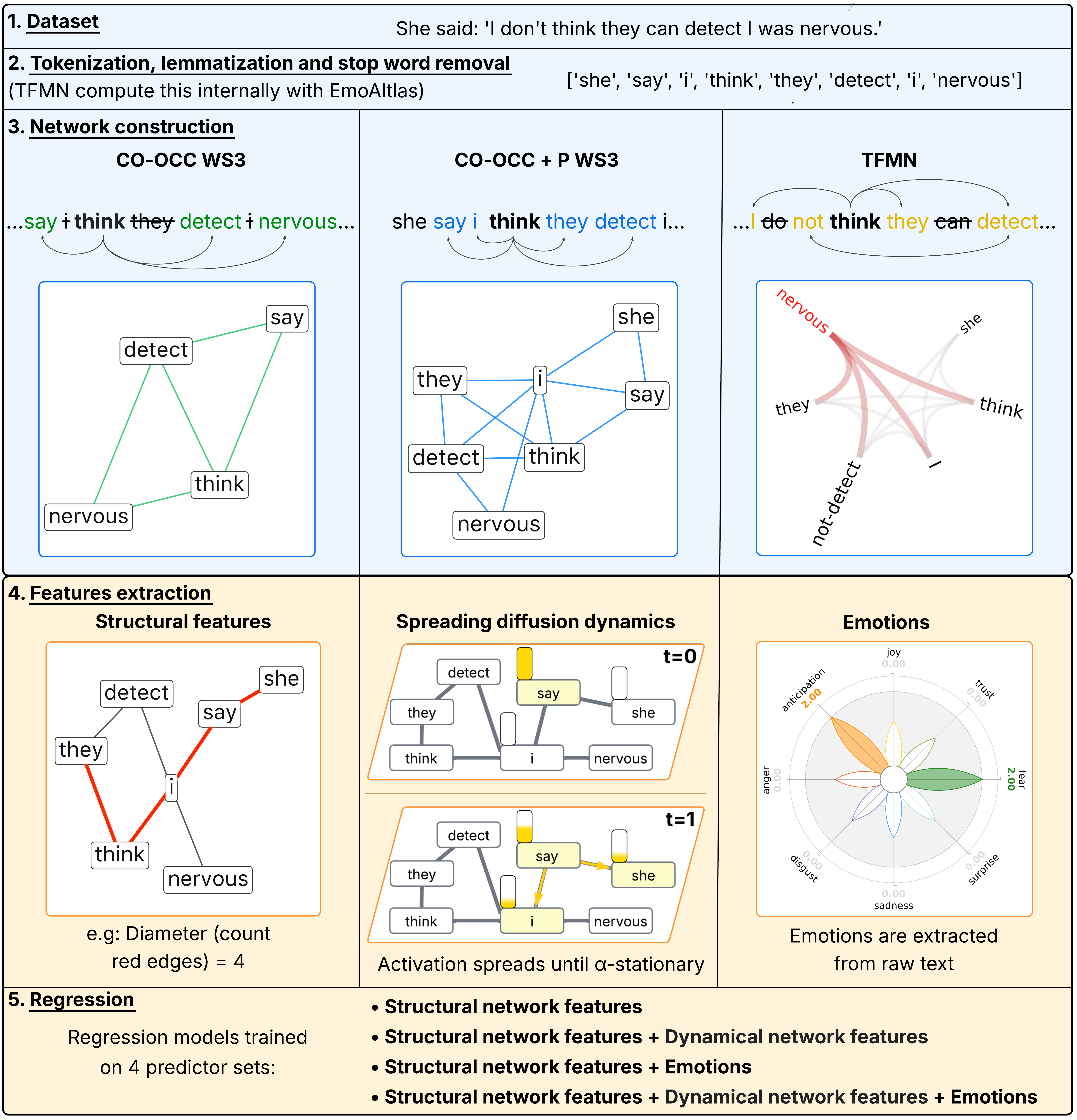}
    \caption{\textit{Overview of the analysis pipeline.} Raw stories are lemmatised and tokenised (Steps 1–2), then converted into three network types (Step 3). TFMNs also capture the valence of words, which is encoded as red for negative, cyan for positive, and grey for neutral. From each network, we extract structural measures, stationary spreading-activation values seeded by the prompt words, and emotion scores from the raw text (Step 4). Regression models are then trained on four feature sets to predict human creativity ratings (Step 5).}

    \label{fig:analysis-pipeline}
\end{figure}

\section{Tutorial: Building and analysing text-based semantic networks for creativity research}
\label{sec:tutorial}
This tutorial provides a step-by-step introduction to constructing semantic networks from texts using two prominent approaches: word co-occurrence and textual forma mentis networks. We also outline how to derive structural, spreading-activation, and emotional features from short narratives. Finally, we introduce the process of using these features in a machine learning pipeline to predict creativity ratings and assess which features contribute in which ways to higher or lower creativity ratings.

This tutorial is written for the following exemplary dataset in mind: a corpus of more than 1,000 short stories collected by \cite{johnson2023divergent}, each narrative being 4-6 sentences long, written by individuals following certain prompt words that needed to be included in the text. Each story has a numeric rating value assigned to it by human raters, who evaluated the stories on a scale from 1 (low creative) to 5 (high creative). The ratings are needed for our machine learning pipeline to predict creativity evaluations and assess which features of the stories are most relevant for predicting creativity levels. 
For datasets where no creativity ratings are available, researchers can still follow steps 1-5 of this tutorial which describe text preprocessing, network construction and extracting structural, spreading activation \citep{collins1975spreading, citraro2025spreadpy}, and emotional features \citep{semeraro2025emoatlas} from the narratives. Instead of predicting creativity levels, researchers may also compare the values from the extracted features between individuals or, in cases of different participant groups, do so as group comparisons.

In the following steps, we guide the reader through a sequential workflow to transform raw, apparently unstructured texts into network structure and extract predictive features from them that can be compared across stories. Step 1 explains text preprocessing and how to get from a raw text to a set of cleaned, lemmatised linguistic units. In Step 2 we cover network construction, where we introduce two alternative approaches for turning these tokens into semantic networks: word co-occurrence networks (Step 2.1) and textual forma mentis networks (Step 2.2). In the present study, we use both approaches for the same texts in order to compare their differing structures and usefulness. In an optional step, we show how the resulting networks can be visualised to gain intuitive insights about the global structure. In Steps 3-5, we describe how different classes of features can be extracted from the networks: network measures (Step 3), spreading-activation values (Step 4), and emotion-related features (Step 5). Finally, Step 6 briefly introduces predictive modelling as a tool for predicting creativity levels from short texts. We explain why such models are useful, and discuss methodological details in the subsequent Analysis section of the paper.  

We provide some example code snippets to illustrate the key operations that researchers need to perform themselves.  We rely on widely used tools implemented in Python (version 3.12.2) such as \texttt{spaCy} (version 3.8.7; \cite{montani2023explosion}) for linguistic preprocessing, \texttt{NetworkX} (version 3.5; \cite{hagberg2008exploring}) for network generation and analysis, and \texttt{EmoAtlas} \citep{semeraro2025emoatlas} for constructing syntactically grounded textual forma mentis networks and extracting emotion features.

Thus, this tutorial is useful for two audiences simultaneously: researchers who are new to text-based network construction can follow a ready-to-use pipeline, while more experienced network scientists can gain a deeper understanding of the methodological assumptions and design choices embedded in these tools. All our code and Python notebooks are freely available at \href{https://osf.io/5cn2y/overview}{OSF repository}.

\subsection{Step 1: Text preprocessing}
\label{tutorial:preprocessing}
The first step in constructing a semantic network from text is to transform raw stories into a standardised set of linguistic units. Text preprocessing reduces superficial variability by normalising different word inflections (e.g., lemmatising "walked", "walking" to "walk"), letter casing (e.g., "School", "SCHOOL" becomes "school") and removing punctuation (which otherwise might be considered as separate concepts). The output from this stage is a set of comparable tokens that can later be treated as nodes in a network representation. 
For text preprocessing we rely on \texttt{spaCy} \citep{montani2023explosion}, which handles most linguistic operations automatically once the correct language model is imported (for instance, \texttt{en\_core\_web\_sm}\ for English texts). Different language models are available on \texttt{spaCy}'s website (https://spacy.io/usage/models/, Last Access 12.12.2025). A typical preprocessing pipeline includes:
\begin{enumerate}
    \item \textbf{Sentence segmentation}: Each story is first segmented into sentences. \texttt{spaCy} identifies sentence boundaries automatically.
    \item \textbf{Tokenisation}: Sentences are split into tokens. Typically, non-alphabetic tokens (numbers, punctuation) are filtered out, which can be done with simple attribute checks.
    \item \textbf{Stop-word removal}: We want to retain alphabetic content words from the texts, so stop-words need to be removed. Content words carry internal semantic meaning and are typically nouns ("house", "creativity"), adjectives ("green", "important") and verbs ("find", "say"). In contrast, stop-words are functional words with little internal semantic meaning, like articles ("a", "the"), conjunctions ("or", "but"), prepositions ("on", "in"), and pronouns ("you", "her"). Stop-words can be removed according to \texttt{spaCy}'s built-in stop-list (\texttt{token.is\_stop}) for the selected language model. Depending on the analysis, researchers may remove all stop-words or choose to retain pronouns, as they impact network structure. To retain pronouns in co-occurrence networks, one can define a fixed set of pronouns (subject, object, possessive, and reflexive forms such as \emph{i}, \emph{me}, \emph{my}, \emph{we}, \emph{our}, \emph{you}, \emph{he}, \emph{she}, \emph{it}, \emph{their}), and allow these items to bypass the stop-word filter.
    \item \textbf{Lemmatisation}: All surviving tokens are then converted to lemmas (e.g., "children"\ $\rightarrow$\ "child"; "played"\ $\rightarrow$\ "play"). Typically, lemmas are lowercased (e.g., "Many"\ $\rightarrow$\ "many") and sentences that contain no remaining valid tokens are discarded.
\end{enumerate}

From the researcher's perspective, the required code in this stage is minimal. The user only has to load the correct language model and pass each story to the pipeline, while \texttt{spaCy} internally performs sentence segmentation, tokenisation, part-of-speech tagging, and lemmatisation. The code snippet in listing \ref{lst:spacy_minimal} illustrates a minimal \texttt{spaCy}-based preprocessing pipeline. Note that this example code is not needed by the user for TFMNs, because they handle text preprocessing internally.
\begin{lstlisting}[
  language=Python,
  caption={Preprocessing with spaCy to convert a story into lemmatised tokens.},
  label={lst:spacy_minimal}
]
python -m spacy download en_core_web_sm
import spacy

nlp = spacy.load("en_core_web_sm", disable=["ner", "parser"])
nlp.add_pipe("sentencizer", first=True)
doc = nlp(text)
tokens = [[t.lemma_.lower() for t in sent if t.is_alpha and not t.is_stop] 
    for sent in doc.sents]
\end{lstlisting}
First, the user downloads and imports \texttt{spaCy} with the appropriate language model for the text (in this case, the English model \texttt{en\_core\_web\_sm}). The named-entity recogniser (\texttt{"ner"}) and dependency parser (\texttt{"parser"}) are disabled as they are not required for basic tokenisation and lemmatisation at this stage. Next, a rule-based \texttt{sentencizer} component is added to the pipeline. This ensures that sentence boundaries are identified. The input text is then processed by the \texttt{nlp} object, which applies tokenisation and lemmatisation automatically and returns a \texttt{Doc} object containing tokens and sentence spans. The final list comprehension extracts the preprocessed tokens. For each sentence, it iterates over the contained tokens and retains only those that are alphabetic content words (\texttt{t.is\_alpha}) and not part of \texttt{spaCy}’s built-in stop-word list (\texttt{not t.is\_stop}). The remaining tokens are converted to their lemma forms and lowercased (\texttt{t.lemma\_.lower}).
The output of this stage is a list of lemmatised tokens for each sentence. Both co-occurrence networks and TFMNs follow these preprocessing steps. The only intentional variation concerns the handling of pronouns. Co-occurrence networks may exclude or include pronouns if desired, whereas TFMNs always retain pronouns by design \citep{semeraro2025emoatlas}. Furthermore, for constructing TFMNs the user does not need to perform these steps explicitly as they are handled by \texttt{EmoAtlas} \citep{semeraro2025emoatlas} internally.

\subsection{Step 2.1: Constructing word co-occurrence networks}
\label{tutorial:cooc}
Co-occurrence networks capture local, surface-level relationships between words by linking terms that appear near one another in the text. They are simple to construct and have long been used in language networks \citep{amancio2015complex,tohalino2020language}. A co-occurrence network can be built via the following steps:
\begin{enumerate}
    \item  \textbf{Preprocessed token sequence:} For each sentence, the sequence of preprocessed, lemmatised tokens is taken (see Step 1). 
    \item \textbf{Choose a window size:} At this step, researchers can decide which window size $WS$ to use for linking adjacent words together. The choice of window size depends on the underlying assumptions of the data and research question. A common choice is to use a window size of 2 (yielding so-called adjacency networks; see \citealp{quispe2021using,stanisz2019linguistic,antiqueira2007some,roxas2010prose,egan2023would}), though slightly larger windows are often used when the goal is to incorporate a wider local co-occurrence context \citep{garg2021survey,tohalino2020language}.
    \item \textbf{Link words based on a sliding window:} For a chosen window size $WS$, each word is linked to the next $WS-1$ words. For instance, if the window size is 3, a word $i$ is linked to the two words on its left and the two words on its right.  
    \item \textbf{Collapse repeated pairs:} How repeated pairs are handled depends on the network characteristics:
\begin{itemize}
        \item \textbf{Unweighted, undirected networks:} All repeated occurrences of the same word pair are collapsed into a single edge. Repeated pairs do not influence the structure of the network through their recurrence. Furthermore, if the network is undirected, the ordering of the words in the word pair does not make a difference. The edge [i, k] is equivalent with [k, i] and both will be collapsed into a single edge.
        \item \textbf{Weighted networks}: In a weighted network, each repetition of a word pair increases the weight of the edge. This reflects the frequency of this co-occurrence.

    \end{itemize}
\end{enumerate}

Constructing co-occurrence networks can be implemented with relatively little custom code (see Listing \ref{lst:coocc_ws3}), because the underlying logic (linking words within a fixed sliding window) is straightforward.
\begin{lstlisting}[
  language=Python,
  caption={Constructing a co-occurrence network with window size $WS=3$.},
  label={lst:coocc_ws3}
]
import networkx as nx

def build_cooccurrence_ws3(lemmas_by_sent):
    G = nx.Graph()
    for sent in lemmas_by_sent:
        for w in sent:
            G.add_node(w)
        for i in range(len(sent)):
            for j in range(i + 1, min(i + 3, len(sent))):   # WS=3
                a, b = sent[i], sent[j]
                if a != b:
                    G.add_edge(a, b)
    return G
\end{lstlisting}

In this code, we define a new function to build a co-occurrence network using the window size 3 (\texttt{build\_cooccurrence\_ws3}). This function takes the list of lemmas extracted from the sentences in the previous step (\texttt{lemmas\_by\_sent}). We rely on the Python library \texttt{NetworkX} to create a network \texttt{G}. Then the function iterates over all sentences and connects words that occur next to each other within a window size of $WS=3$. We prevent self-loops by adding the condition that both words in the word pair need to be different from each other (\texttt{if a != b}). The output from this stage is an undirected, unweighted co-occurrence network $G$.

\subsection{Step 2.2: Constructing textual forma mentis networks}
\label{tutorial:tfmn}
Textual forma mentis networks (TFMNs) connect words according to their syntactic dependencies rather than the distance of tokens in the surface text \citep{semeraro2025emoatlas,haim2024forma,stella2020text}. Unlike co-occurrence networks, TFMNs encode grammatical structure that may better reflect conceptual relationships. In a sentence like "Lucy, despite her immense fear of heights, loves hiking", co-occurrence models fail to connect "Lucy" to "loves hiking" due to the intervening phrase. In contrast, TFMNs are able to link the concepts due to their grammatical relationship. This enables TFMNs to capture meaningful dependencies even in long or structurally complex sentences.

TFMNs not only contain dependency-based edges, but are attributed networks in which each node represents a lexical concept that can be enriched with external annotations. Nodes in the network are assigned a valence score, which captures whether concepts are perceived as positive, negative or neutral (see \cite{semeraro2025emoatlas,mohammad2013crowdsourcing}). Furthermore, users can apply semantic enrichment, where the edge set is extended with synonym relations (e.g. "dog", "canine") or hypernym/hyponym relations ("mammal", "mouse"). These are added from the lexical resource \texttt{WordNet} \citep{miller1995wordnet} to complement rather than override the grammatical structure captured by the edge set that is anchored in syntactic dependencies. 

TFMNs have been successfully applied to capture and analyse public perceptions on Social Media regarding topics such as the STEM gender gap \citep{stella2020text} as well as for predicting creativity levels in short narratives \citep{haim2024forma}. TFMNs are constructed using the \texttt{EmoAtlas} library \citep{semeraro2025emoatlas}, which performs the following steps:
\begin{enumerate}
    \item \textbf{Tokenisation and normalisation:} Text preprocessing follows the procedure described in Step 1 of this Tutorial. Texts are segmented into sentences, tokenised, and lemmatised. 
    \item \textbf{Syntactic parsing:} Each sentence is parsed with \texttt{spaCy}'s dependency parser. This results in a syntactic tree that encodes grammatical relations between words. 
    \item \textbf{Connecting words:} All non-stop words (typically nouns, verbs, adjectives) that lie within a distance threshold of three dependency steps on the syntax tree are connected. This "dependency window" differs fundamentally from the sliding window used in co-occurrence networks. While co-occurrence models define distance in terms of word position in the sentence string, TFMNs define distance in terms of grammatical relations on the syntax tree. As a result, in TFMNs words can be linked even when they are far apart in the surface text but closely related syntactically.
    \item \textbf{Network construction:} For each sentence, a network is built by linking syntactically related tokens identified in the previous steps. These sentence-level networks are then merged into a unified network for the entire text.
    \item \textbf{Semantic enrichment:} Optionally, the resulting network can be enriched with synonym or hypernym/hyponym relations, connecting conceptually similar words even when they are not directly linked by syntax \citep{semeraro2025emoatlas,miller1995wordnet}.
    \item \textbf{Assigning valence:} In addition to syntactic structure, TFMNs also encode affect. Each word in the network is annotated with valence labels (positive, negative, neutral) using the EmoLex lexicon which is a psychological lexicon of emotion words \citep{mohammad2013crowdsourcing}. Because the network is built on syntactic dependencies, negations can be handled correctly and considered for attributing valence labels. When a word carrying an emotion is negated ("not angry"), that word is interpreted as its opposite ("not angry" → opposite of "angry"; see \cite{semeraro2025emoatlas, haim2024forma}).
\end{enumerate}

These steps are naturally implemented in the \texttt{EmoAtlas} library, which makes its use intuitive and easy. Thus, the users only need minimal code to transform a text into a TFMN (see Listing \ref{lst:tfmn}).
\begin{lstlisting}[language=Python,
caption={Constructing a textual forma mentis network (TFMN) using EmoAtlas.},
label={lst:tfmn}]
pip install git+https://github.com/MassimoStel/emoatlas
from emoatlas import EmoScores
import emoatlas

emo = EmoScores(language = "english")
G = emo.formamentis_network(text)
\end{lstlisting}
First, the user downloads and imports the EmoAtlas package from GitHub. The correct language is selected, in this case English. \texttt{EmoAtlas} has also been tested for Italian and supports further languages available on \texttt{spaCy}. Finally, the code \texttt{emo-formamentis\_network()} creates a dependency-based network $G$.  \texttt{EmoAtlas} also supports more features, such as enriching the network with synonyms or plotting an emotion flower (see on the \href{https://github.com/MassimoStel/emoatlas/wiki/0-%E2%80%90-Home}{EmoAtlas GitHub} page).

\subsection{Optional: Visualisation of Co-occurrence Networks and TFMNs}
\subsubsection{Standard visualisation}
After creating the networks from the texts, one can visualise the networks (see Listing \ref{lst:viz_simple}). Nodes correspond to word lemmas and edges encode either co-occurrence or syntactic relations.

Visualising networks is useful for getting an intuition for conceptual organisation, and for inspecting structural properties that are not apparent from summary statistics alone. Visual layouts can help identify central concepts, thematic clusters, and disconnected components (cf. \cite{siew2019cognitive}). Furthermore, network visualisation can make the node labels of the network explicit, supporting qualitative interpretability and relating the network structure back to the underlying text \citep{haim2023cognitive}.

\begin{lstlisting}[language=Python,
caption={Visualising semantic networks using a force-directed layout.},
label={lst:viz_simple}]
import pandas as pd
import networkx as nx
import matplotlib.pyplot as plt

# load edge list
edges = pd.read_csv(edge_csv)
G = nx.from_pandas_edgelist(edges, "source", "target")
# compute layout and draw full network
pos = nx.spring_layout(G, seed=7)
nx.draw(G, pos, with_labels=True, node_size=500, font_size=10)
plt.show()
\end{lstlisting}

First, the code reads a CSV file containing an edge list, which lists pairs of connected words. Each row in this file specifies a connection between a source and a target node. All connections taken together define the structure of the network. The code then computes a force-directed layout (\texttt{spring\_layout}), which positions nodes in a way that connected nodes are drawn closer together while minimising edges crossing over each other \citep{hills2024behavioral}. Finally, the network is drawn with node labels (indicating the word represented by each node), and the figure is displayed.

\begin{figure}[!htbp]
    \centering
    \includegraphics[width=0.32\textwidth]{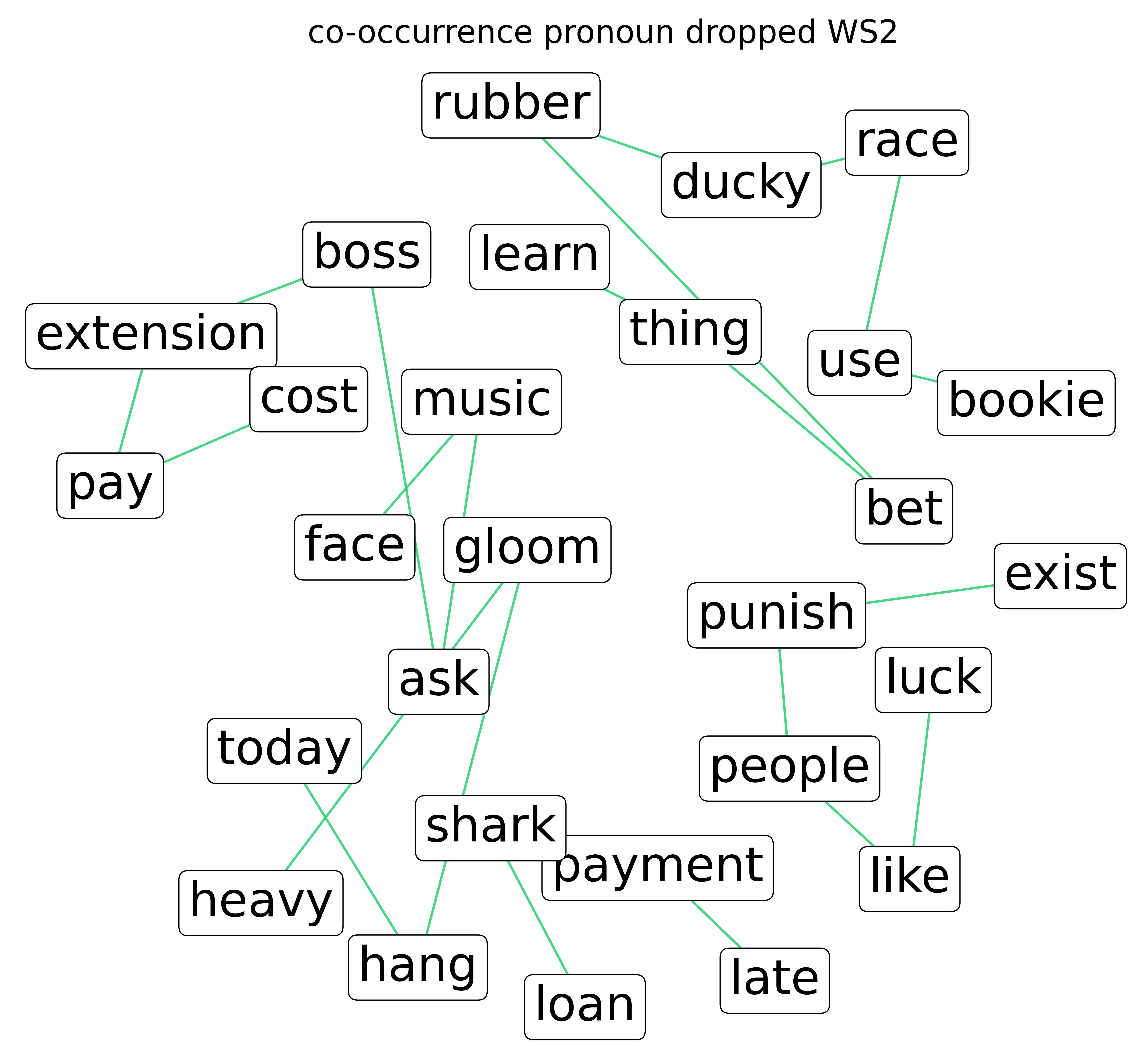}
    \includegraphics[width=0.32\textwidth]{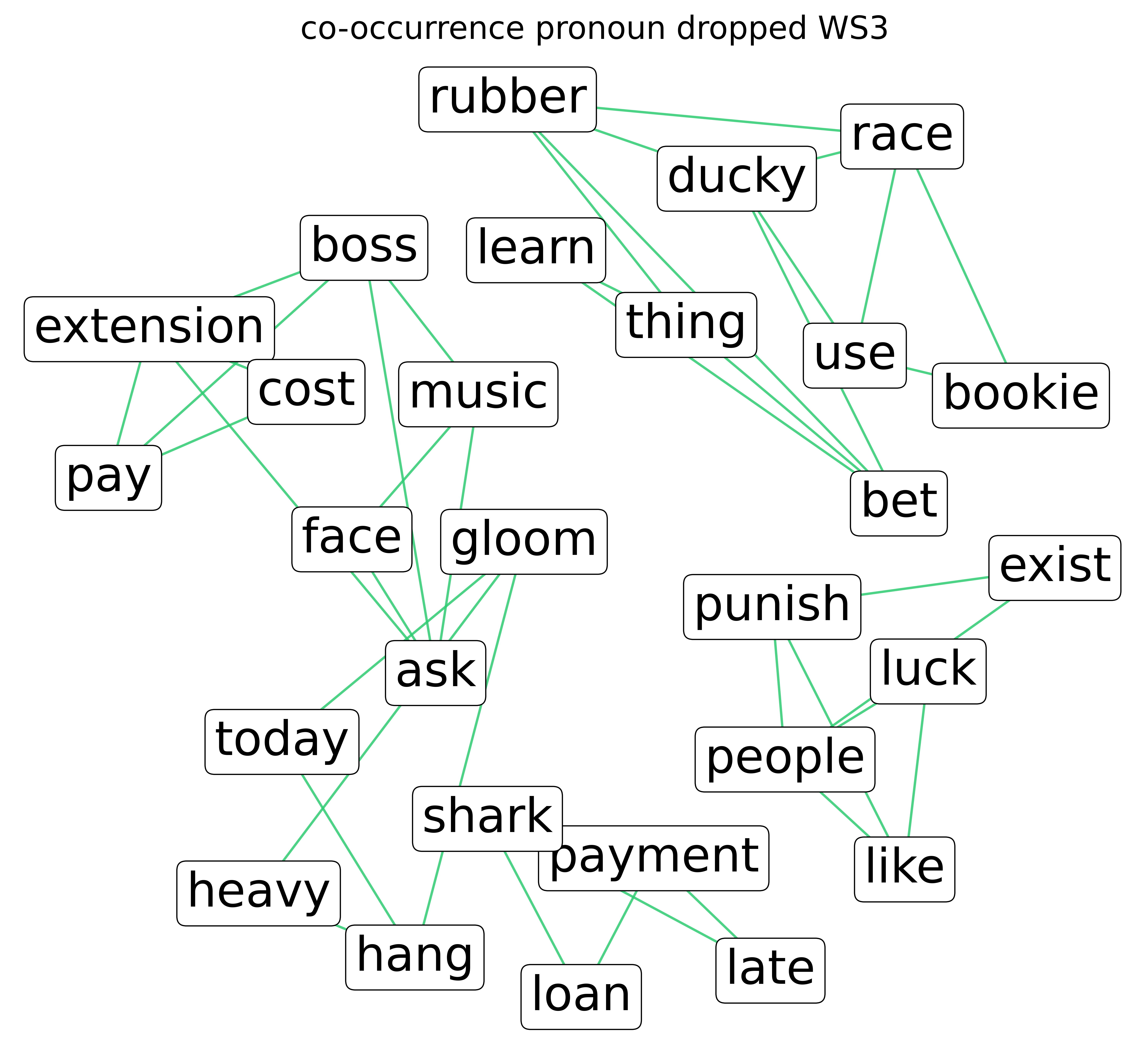}
    \includegraphics[width=0.32\textwidth]{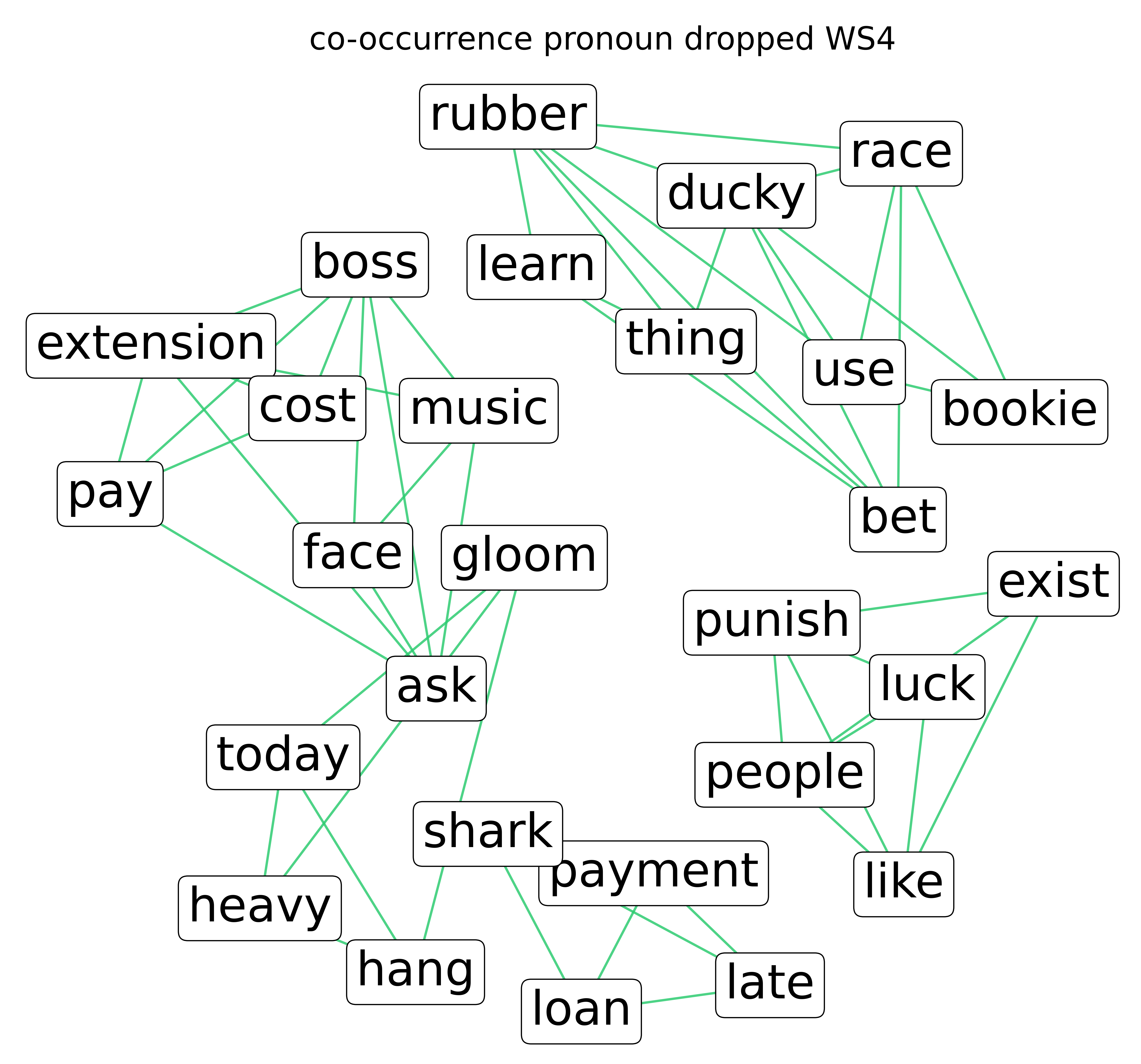}

    \includegraphics[width=0.32\textwidth]{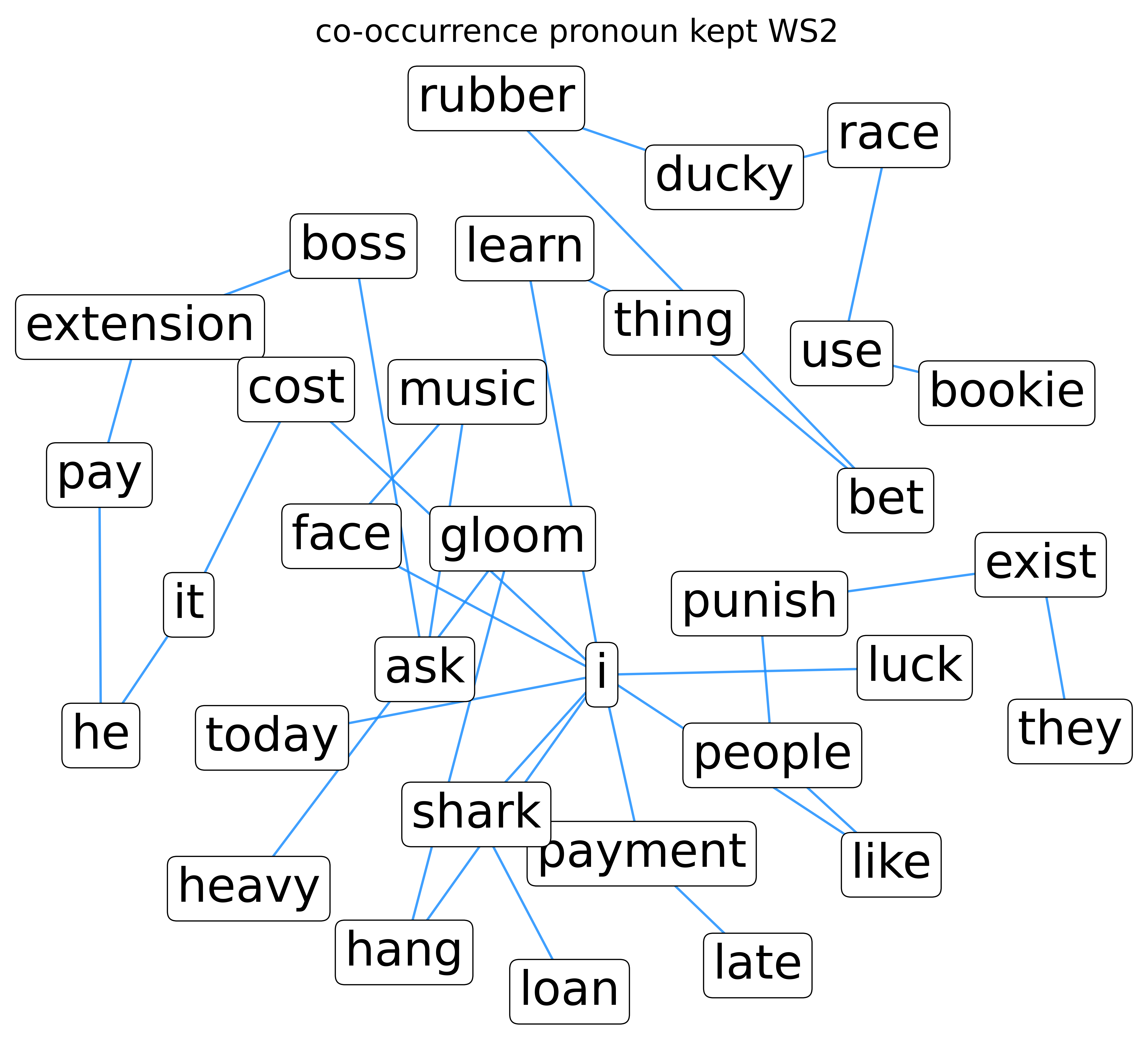}
    \includegraphics[width=0.32\textwidth]{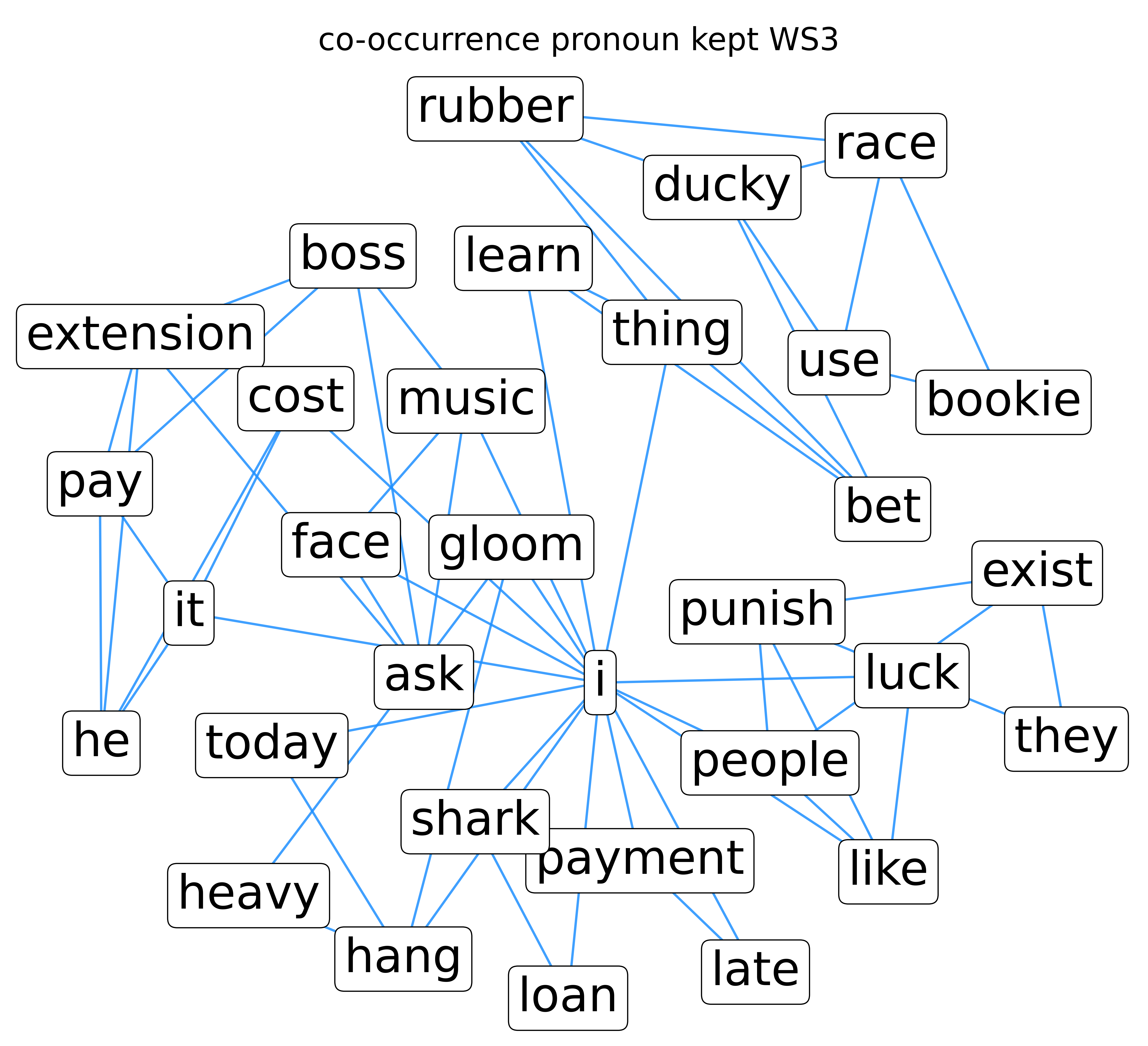}
    \includegraphics[width=0.32\textwidth]{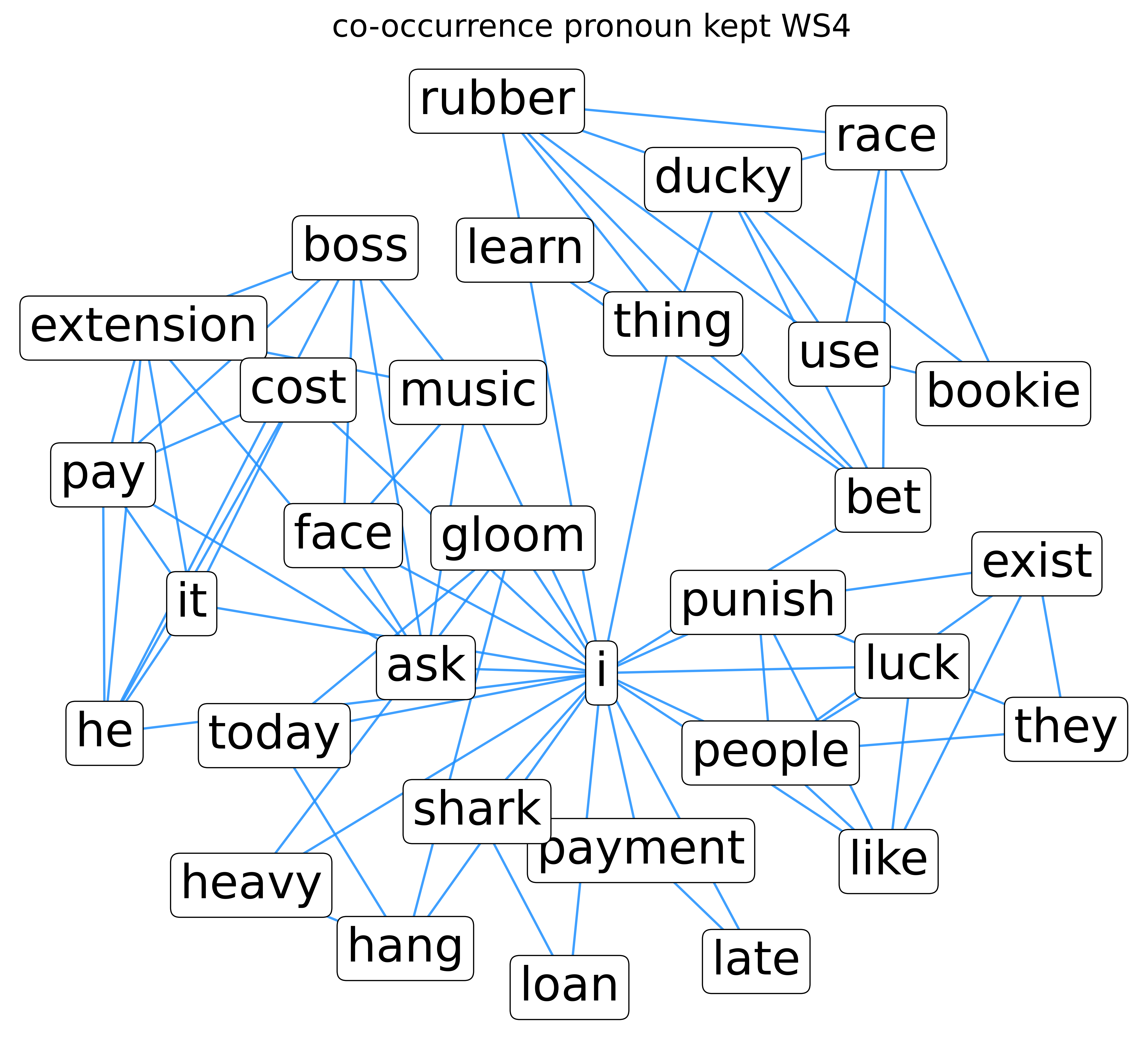}\\[1em]
    
    \adjustbox{valign=c}{\includegraphics[width=0.32\textwidth]{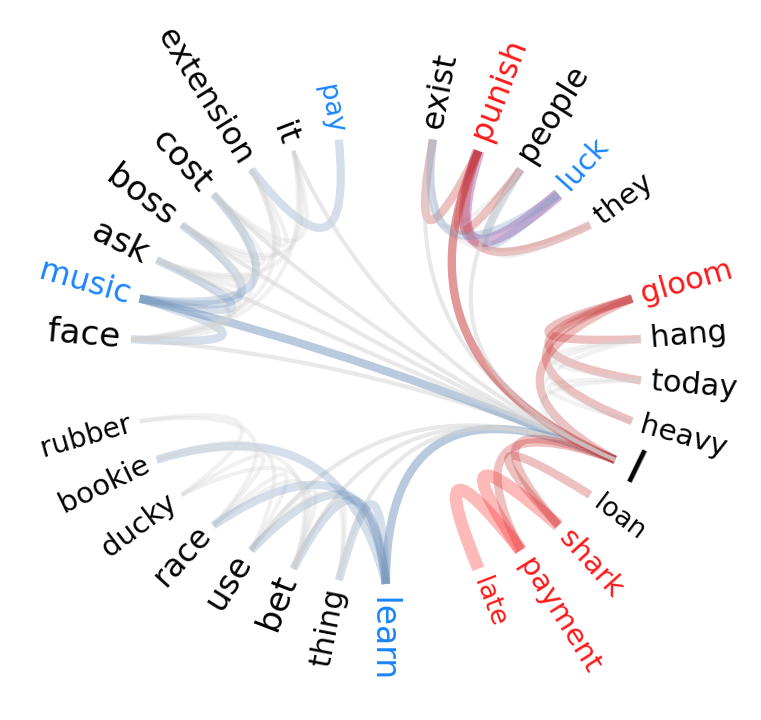}}
    \adjustbox{valign=c}{\includegraphics[width=0.33\textwidth]{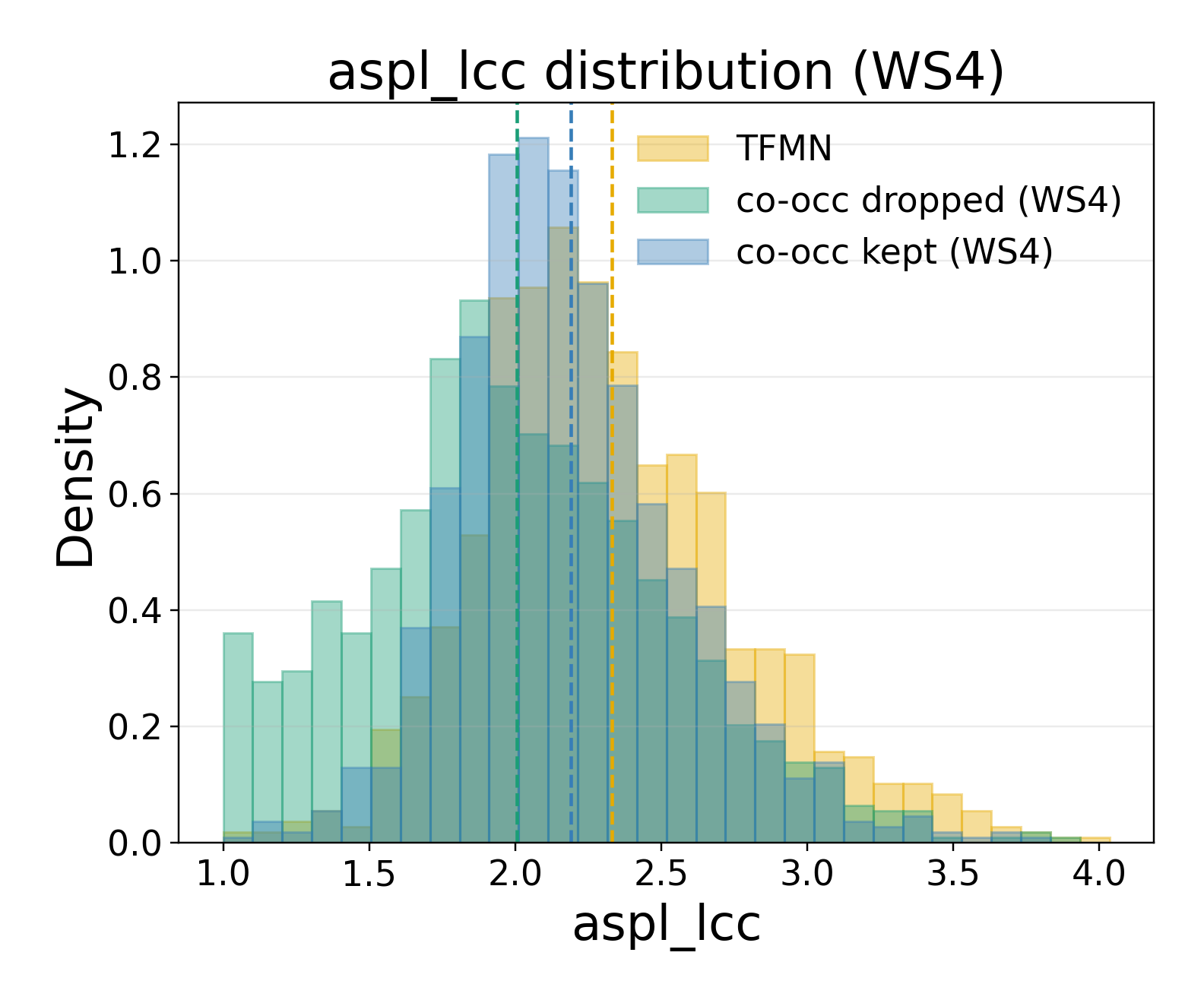}}
    \adjustbox{valign=c}{\includegraphics[width=0.33\textwidth]{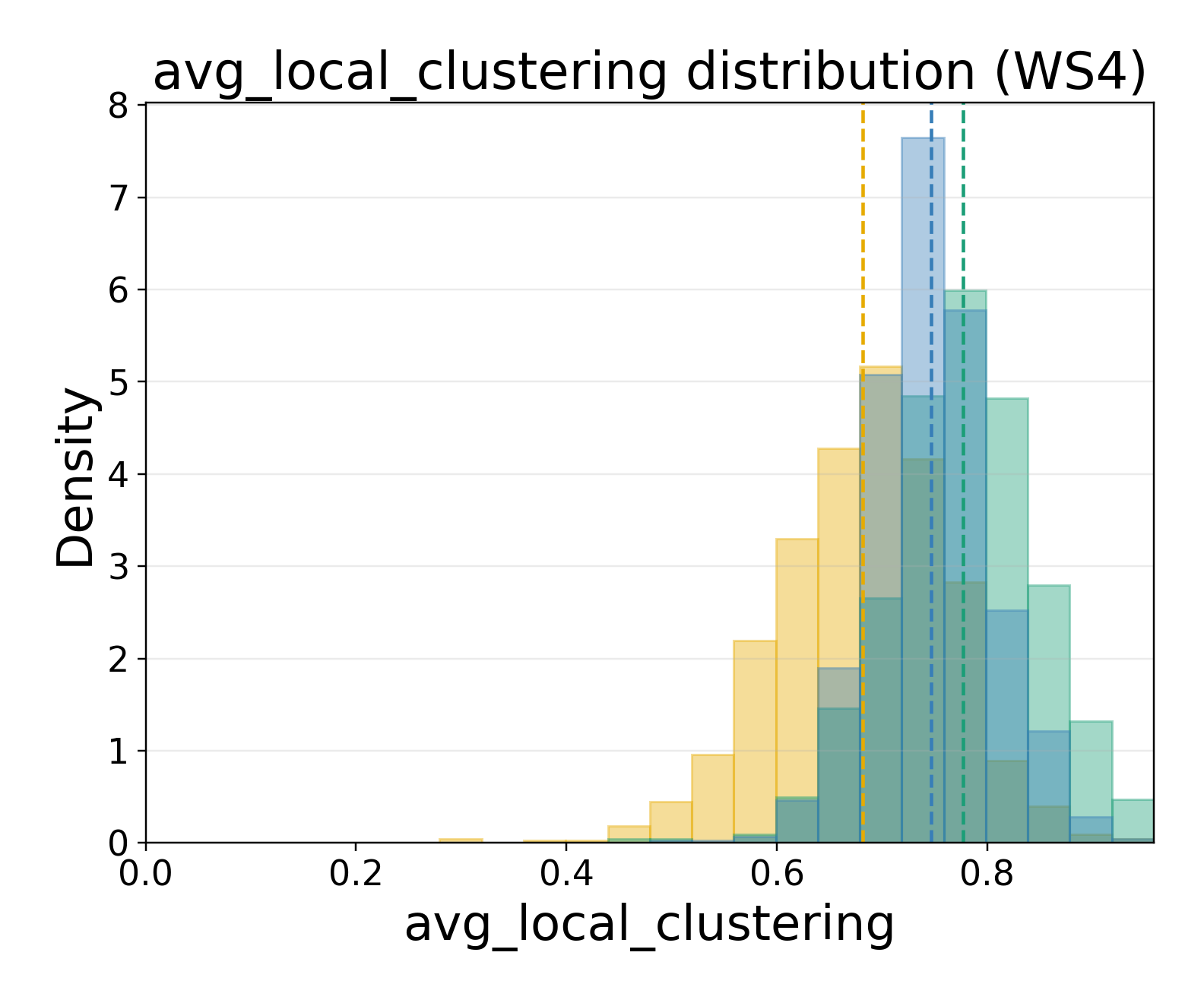}}\\[1em]

\caption{\textit{Story:} "There is a heavy gloom hanging over me today. The loan sharks are after me because I have a payment that is late. They exist to punish people like me, down own there luck. I might as well face the music and ask the boss for an extension on paying him back even though it will cost me more. I have learned one thing though, never bet on a rubber ducky race and use a bookie." The panels show co-occurrence networks without pronouns (top), with pronouns (middle), and the textual forma mentis network (bottom). The final panels show ASPL and clustering coefficient distributions for WS4, including TFMN for comparison.}

    \label{fig:graphs}
\end{figure}

Example networks created from the same story can be seen in Figure \ref{fig:graphs}. The figure visualises all seven network configurations examined in this study: co-occurrence networks without pronouns (window sizes 2, 3, and 4), co-occurrence networks with pronouns retained (window sizes 2, 3, and 4), and the corresponding textual forma mentis network. These visualisations provide an intuitive illustration of how methodological choices in network construction shape the resulting network structure. Increasing the co-occurrence window size systematically adds edges, resulting in denser and more connected networks. Small window sizes produce sparse networks with many disconnected components, particularly when pronouns are removed. Retaining pronouns substantially alters this structure. In the present example, the pronoun “I” acts as a central hub that connects otherwise separate parts of the network, markedly reducing fragmentation. This effect highlights how seemingly minor preprocessing decisions, such as removing pronouns, can have strong consequences for network connectivity in short texts. The TFMN, shown in the bottom-left panel of Fig. \ref{fig:graphs}, differs strongly from co-occurrence variants with small window sizes and excluding pronouns. Because edges are derived from syntactic dependencies rather than sentence proximity, it produces a network that is both denser and more structurally coherent. The bottom-right panels of Fig. \ref{fig:graphs} further compare distributions of average shortest path length (ASPL) and local clustering coefficient for the largest connected component of the WS4 co-occurrence network and the TFMN. These distributions illustrate that the two network-building approaches differ systematically in how words are organised in the networks. In particular, they highlight differences in how far apart concepts are (global distance) and how tightly groups of related words cluster together (local cohesion). This illustrates that networks built from syntactic relations capture different patterns of conceptual organisation than networks based on surface word proximity. 

\subsubsection{TFMN visualisation using EmoAtlas}
When networks are constructed as textual forma mentis networks (TFMNs), \texttt{EmoAtlas} provides a dedicated visualisation routine that reproduces its standard graphical encoding for syntactic structure and affective valence \citep{semeraro2025emoatlas}. Nodes are coloured according to their valence (positive, negative, neutral) as provided by EmoLex \citep{turney2019natural}: positive concepts are shown in cyan, negative concepts in red, and neutral concepts in black \citep{semeraro2025emoatlas}. In addition to syntactic dependencies, EmoAtlas can optionally overlay lexicon-derived semantic relations (e.g., synonym and hypernym/hyponym relations from WordNet) on top of the syntactic scaffold. These semantic links can be toggled for clarity (and, when highlighted, are rendered in green), ensuring that semantic augmentation complements rather than replaces dependency structure \citep{miller1995wordnet,semeraro2025emoatlas}. Listing~\ref{lst:viz_tfmn} shows a minimal example for visualising the full TFMN extracted from a text.

\begin{lstlisting}[language=Python,
caption={Visualising a textual forma mentis network (TFMN) with EmoAtlas.},
label={lst:viz_tfmn}]
from emoatlas import EmoScores
import matplotlib.pyplot as plt

emo = EmoScores(language="english")
fmnt = emo.formamentis_network(text)
emo.draw_formamentis(
    fmn=fmnt,
    alpha_syntactic=0.4, # Transparency of syntactic links.
    alpha_hypernyms=0.4, # Transparency of hypernym links
    alpha_synonyms=0.4,  # Transparency of synonym links.
    thickness=2
)
plt.show()
\end{lstlisting}

For further details on \texttt{EmoAtlas} functions and visual conventions, see the\href{https://github.com/MassimoStel/emoatlas/wiki/0-%E2%80%90-Home}{EmoAtlas GitHub} page.

\subsection{Step 3: Extracting network-structural features}
\label{tutorial:structure}
From the networks built in Step 2.1 and 2.2, one can extract structural features describing network size, local structure, path organisation, and centrality. The features described below capture different aspects of how concepts are organised in the narrative, reflecting potential information flow that may be relevant for creative ideation \citep{haim2024forma}.

\vspace{0.2cm}
\textbf{Network size and network order}: Network size is captured by the total number of edges  \(m = |E|\). Network order refers to the total number of nodes \(n = |V|\). Network size and network order provide a basic measure of the complexity and connectivity of the network \citep{newman2010networks,diestel2025graph}. Larger networks with more edges indicate richer associations between concepts.

\vspace{0.2cm}
\textbf{Diameter (D)}: The diameter represents the longest distance between any two nodes in the largest connected component of the network \citep{newman2010networks}. It is defined as \[
D = \max_{i, j \in \mathrm{LCC}} d(i, j),
\]
where \(d(i, j)\) denotes the shortest-path distance between nodes \(i\) and \(j\).

\vspace{0.2cm}
\textbf{Average shortest path length (ASPL)}: This reflects the mean of the shortest number of steps needed to connect every pair of nodes in the network \citep{newman2010networks,watts1998collective}. Lower values indicate that concepts are more easily reachable, allowing more efficient information flow or activation across the network. In relation to creativity, shorter ASPL in networks based on fluency data has been linked to higher creative performance, as they facilitate rapid transitions between concepts \citep{kenett2019semantic,siew2023investigating,wang2025fluency}. At the same time, in story-based networks, longer paths have been related to higher creativity of short narratives, as they may reflect broader thematic structure \citep{haim2024forma}. ASPL can be calculated on the largest connected component (LCC) as \[
ASPL = \frac{1}{n_{\mathrm{LCC}}(n_{\mathrm{LCC}} - 1)}
\sum_{\substack{i \ne j \\ i, j \in \mathrm{LCC}}} d(i, j),
\]

\vspace{0.2cm}
\textbf{Density}: This measure represents the ratio of actual connections to the total possible number of connections in the network \citep{newman2010networks}. Higher density indicates a more interconnected network where concepts are more closely linked. The density $d$ is calculated as \[
d = \frac{2m}{n(n-1)}\] where $m$ represents the network size (total number of edges) and $n$ is the network order (total number of nodes).

\vspace{0.2cm}
\textbf{Clustering coefficient}: This measure quantifies the local density of an undirected network by measuring the extent to which a node’s neighbours are interconnected to form triangles \citep{newman2010networks,watts1998collective}. High clustering indicates a cohesive local structure, which can facilitate the spread of activation between related concepts \citep{vitevitch2012complex}. For a node $i$ with a set of neighbours $N_i$ and degree $k_i = |N_i|$, the local clustering coefficient $C_i$ is defined as the ratio of actual edges between neighbours to the maximum possible number of edges that could exist between them:
\[
C_i = \frac{2\,|\{\,e_{jk} : v_j, v_k \in N_i,\ e_{jk} \in E\,\}|}{k_i (k_i - 1)}.
\]
where the numerator represents the number of existing edges between the neighbours of node $i$.
To avoid disproportionate influence from sparsely connected nodes with only one neighbouring node (degree $k$=1), it is common to compute \(C_i\) only for nodes with \(k_i\ge 2\) and report the mean value over all such nodes.

\vspace{0.2cm}
\textbf{PageRank}: Finally, we compute PageRank values. The PageRank value $r_i$ of a node represents the asymptotic probability that a random walker, navigating links with probability $1 - \alpha$ or teleporting with probability $\alpha$, will be at that specific node \citep{griffiths2007google,page1999pagerank}. To quantify the structure of the network, we calculate a PageRank Centralisation index $C_{\mathrm{PR}}$. Higher values of $C_{\mathrm{PR}}$ indicate that a few hub nodes dominate in importance, highlighting influential concepts that may guide the flow of activation. Let $r_i$ denote the PageRank value of node $i$, and let $u = 1 / n_{\mathrm{LCC}}$ be the uniform baseline. We define
\[
S = \sum_{i \in \mathrm{LCC}} \lvert r_i - u \rvert,
\]
and the normalised index as
\[
C_{\mathrm{PR}} = \frac{S}{n_{\mathrm{LCC}}}.
\]

These measures can be extracted with \texttt{NetworkX}. They form the core of our predictive modelling in the study described in the following and provide an interpretable description of narrative structure.

\subsection{Step 4: Extracting spreading-activation features}
\label{tutorial:activation}
Spreading activation models conceptual relationships according to how a fictional activation signal flows through a given cognitive network \citep{collins1975spreading,anderson1983spreading}. One or more seed nodes is initially activated and, subsequently, every activated node shares a portion of its activation with its neighbours, iteratively. Extensive empirical research has shown that spreading activation can highlight cognitive patterns of priming in the mental lexicon, cf. \citep{aitchison2012words,vitevitch2021cognitive,siew2019spreadr}. In the study at hand, we use \texttt{SpreadPy} \citep{citraro2025spreadpy} to simulate spreading activation on cognitive networks and, subsequently, compute stationary activation values, i.e. the activation level remaining over specific nodes after several iterations.

The \texttt{SpreadPy} package follows the steps below to simulate activation spreading:
\begin{enumerate}
\item Activation is injected at one or more seed nodes with a user-defined activation level (e.g. total number of nodes). 
\item From these seed nodes activation can diffuse iteratively along the network's links. 
\item In \texttt{SpreadPy}, spreading dynamics can be controlled by a retention parameter \(r \in (0,1)\). This determines how much activation is preserved at each iteration. A lower retention parameter causes less activation to be retained by each node and more activation to flow to its neighbouring nodes. A higher retention parameter causes more activation to be "swallowed" by a node with less activation remaining to be spread on. The choice of retention parameter depends on the experimental design and underlying assumptions of the network. A standard practice is to adopt a default value of \(r = 0.5\) (meaning 50 percent of activation get retained by the node and 50 percent are spread on; see \cite{abramski2025word,vitevitch2021cognitive,vitevitch2021exploring}).
\item Once the diffusive process has explored every node several time, the stationary activation values indicate how strongly each node remains active in a dynamical equilibrium, i.e. the amount of activation leaving the node equals the amount of activation incoming on the node. This reflects both the node's position in the network and its connection to the initial seeds \citep{citraro2025spreadpy}.
\end{enumerate}
The stationary spreading-activation values can be interpreted as an activity-based centrality measure, that highlights the relative importance of each node in the spreading-activation process \citep{shabahang2024latent,koponen2021systemic,abarghouei2020random}.

\subsection{Step 5: Extracting emotion features}
\label{tutorial:emotions}
Emotion features can be extracted from the full text using the \texttt{EmoAtlas} Python package \citep{semeraro2025emoatlas}.
\texttt{EmoAtlas} provides a quantitative profile of the eight primary emotions in Plutchik’s wheel of emotions: \textit{joy}, \textit{trust}, \textit{fear}, \textit{surprise}, \textit{sadness}, \textit{disgust}, \textit{anger}, \textit{anticipation} \citep{plutchik1980general}. The package provides z-scores to quantify the intensity for each of these emotions in a given text. The observed frequency of words carrying a specific emotion is compared to the frequency expected under a null model in which words are sampled at random from the EmoLex lexicon \citep{mohammad2013crowdsourcing,stella2020text}. Hence, a z-score provided by \texttt{EmoAtlas} reflects whether an emotion is over- or under-represented relative to chance \citep{semeraro2025emoatlas}. Following the framework used by \cite{semeraro2025emoatlas}, and fixing a significance value of 0.05, values above 1.96 indicate that a given emotion is more frequent than expected by chance, while values below -1.96 indicate lower frequency than expected from the null model.

\subsection{Step 6: Building predictive models}
\label{tutorial:modelling}
The features extracted in the previous steps (network measures, spreading-activation values, emotion scores) can be used in machine-learning models to predict creativity ratings assigned to the stories, e.g. \citep{semeraro2025emoatlas}. In this paper, machine learning is used as a practical tool to test whether the extracted features capture information that is relevant for how creativity is judged by human raters.

Each story is represented by a set of numeric features (e.g. number of nodes, density, emotion scores), and is associated with a creativity score provided by human raters. Then, regression models are trained on a subset of the data to learn the relationship between these features and the creativity ratings. Their performance is then evaluated on stories the model had not seen during training. This procedure allows researchers to assess how well creativity can be predicted from different feature sets.

To ensure interpretability and reproducibility, we recommend the following best practices:
\begin{itemize}
    \item using cross-validation for model evaluation;
    \item comparing feature sets systematically (network-only, network+emotion, etc.);
    \item performing exploratory analyses of hyperparameter settings for each model;
    \item computing SHAP values to interpret how each feature influences model outputs.
\end{itemize}

In the remainder of this paper, we apply this modelling framework to compare co-occurrence networks and TFMNs, using machine learning to assess which network representation better captures creativity-relevant structure in short stories. We further assess which of the extracted features (network measures, spreading-activation values, emotion scores) is most predictive of higher or lower creativity ratings.

\vspace{1cm}

\section{Methods}
We outline our methodological pipeline in Figure~\ref{fig:analysis-pipeline}. Our goal is to derive structural features of short narratives and evaluate how well these representations predict human creativity ratings. We begin by lemmatising and tokenising the raw stories, then construct three types of networks: co-occurrence networks with and without pronouns (across window sizes $WS \in \{2,3,4\}$), and textual forma mentis networks. From each network, we extract structural features and stationary spreading-activation values based on the prompt tokens. In parallel, we compute emotion features directly from the full text. We use these features as predictors and combine them into four predictor sets, which we use in regression models to assess their respective contributions to predicting creativity.

\subsection{Dataset}
\label{sub:dataset}
The present study uses a corpus of 1071 short narratives collected by \cite{johnson2023divergent} in their second study on divergent semantic integration. They recruited 153 participants through Amazon’s Mechanical Turk (mean age 38.62, range 22-70; 54\% female, 2\% non-binary). The participants were predominantly native English speakers (97\% with English as first language) and completed a creative writing task. In this, each participant produced seven short narratives, each prompted by a triad of cue words that had to be included in the story. The following seven prompts were presented in randomised order to participants: belief-faith-sing; gloom-payment-exist; organ-empire-comply; petrol-diesel-pump; stamp-letter-send; statement-stealth-detect; year-week-embark. For these three-word-prompts, participants were asked to write 4-6 sentences within a four-minute time limit \citep{johnson2023divergent}. 

After collecting the short stories, 4 raters evaluated the creativity of each story using a five-point scale (1 = least creative; 5 = most creative). Raters were instructed to rate the creativity based on how emotive and vivid the story was, and how creatively the prompt words were used within the narrative \citep{johnson2023divergent}. 

We excluded stories from our analysis in which the writer failed to include one of the three required prompt words. This was necessary to ensure our analyses, like spreading-activation initiated at the prompt-words, could be executed for all stories. When checking if the prompt words were present in a story, we accepted any morphological or grammatical variants (e.g., plural forms, different verb conjugations, minor spelling errors). After this criterion, we removed 42 stories from the original dataset and retained 1029 stories.

\subsection{Text preprocessing}
\label{sub:processing}
As outlined in the Tutorial Section, we performed text preprocessing (including sentence segmentation, tokenisation, and lemmatisation) in Python using \texttt{spaCy} \citep{montani2023explosion}. Specifically, we applied the \texttt{en\_core\_web\_sm} model with the dependency parser and named entity recogniser disabled for efficiency, and we added \texttt{spaCy}'s rule-based \texttt{sentencizer} component to obtain sentence boundaries.

Using \texttt{spaCy}, each story was first segmented into sentences. Within each sentence, we iterated over tokens and applied the following filters. We discarded numbers, punctuation, and other non-alphabetic strings. We then removed stop-words according to \texttt{spaCy}'s built-in English stop-list (\texttt{token.is\_stop}), with a controlled exception for personal and possessive pronouns. Concretely, we defined a fixed set of English pronouns (subject, object, possessive, and reflexive forms such as \emph{i}, \emph{me}, \emph{my}, \emph{we}, \emph{our}, \emph{you}, \emph{he}, \emph{she}, \emph{it}, \emph{they}), and allowed these items to bypass the stop-word filter when pronouns were meant to be retained. All surviving tokens were converted to lemmas and then lowercased (e.g., "Many"\ $\rightarrow$\ "many"; "children"\ $\rightarrow$\ "child"; "playing"/\ "played"\ $\rightarrow$\ "play"). Sentences for which no token passed these filters were discarded.

We used the preprocessing pipeline described in the Tutorial Section in two alternative configurations depending on the network construction condition. In the co-occurrence condition without pronouns (\texttt{coocc}), we discarded pronouns during stop-word removal. In the pronoun-inclusive co-occurrence condition (\texttt{coocc\_p}) we instead enabled pronoun retention, allowing pronouns to appear as nodes and to participate in edges.
Retaining pronouns in \texttt{coocc\_p} serves to control for their presence in downstream comparisons with TFMNs, which keep pronouns by design \citep{semeraro2025emoatlas}. This ensures that any differences between co-occurrence networks and TFMNs are not merely artefacts of different token-filtering choices.

\subsection{Different network construction methods}
\label{sub:net_construction}
\vspace{0.2cm}

\subsubsection*{Co-occurrence networks}
We constructed unweighted co-occurrence networks in Python using \texttt{NetworkX} \citep{hagberg2008exploring}. For each sentence and window size $WS \in \{2, 3, 4\}$, we linked each lemma to the following lemmas in the same sentence, if present (Figure \ref{fig:analysis-pipeline}). For instance, with $WS=3$, the lemmatised sequence "child play football game" contains the edges (child--play), (child--football), (play--football), (play--game), and (football--game).
We applied this procedure to all stories and all window sizes, and created six co-occurrence variations per story: three networks without pronouns (\texttt{coocc}\_WS2, \texttt{coocc}\_WS3, \texttt{coocc}\_WS4) and three networks with pronouns retained (\texttt{coocc\_p}\_WS2, \texttt{coocc\_p}\_WS3, \texttt{coocc\_p}\_WS4). These networks capture short-range syntagmatic associations (i.e. relationships between words based on their horizontal arrangement in a sentence) in the text. However, they do not encode full syntactic dependency structures or semantic roles within a story \citep{amancio2015complex,tohalino2020language}, unlike textual forma mentis networks \citep{stella2020text,semeraro2022emotional}.

\subsubsection*{Textual forma mentis networks}
TFMNs were constructed using the \texttt{EmoAtlas} library \citep{semeraro2025emoatlas}. After text preprocessing, TFMNs derived syntax trees from the texts, thereby capturing the grammatical dependencies within a sentence. All non-stop words (nouns, verbs, adjectives, adverbs) that lay within a radius of three steps on the syntax tree were connected. We refer to the Tutorial Section for more details on both textual forma mentis networks and co-occurrence networks.

\subsection{Three types of features: Network features, alpha stationary, emotion features}
\label{sub:features}

\vspace{0.2cm}
\subsubsection*{A) Network features}
For each story and each network representation, we constructed an undirected, unweighted network \(G = (V, E)\). In these networks, word lemmas are represented by nodes and relations (co-occurrence or syntactic) between them are encoded as edges. We then extracted the following structural features as predictors in our regression model: number of nodes and edges, density, clustering coefficient (CC), average shortest path length (ASPL), diameter, and PageRank centralisation. As outlined also in the Tutorial Section of this manuscript, such network measures provide an interpretable description of narrative structure and may encode features relative to spreading activation during creative writing \citep{haim2024forma}.

\subsubsection*{B) Spreading-activation features}
With the aim of testing whether text-based networks encode features of spreading activation related with creativity levels, we computed spreading-activation features \citep{collins1975spreading,anderson1983spreading}. These features quantify how strongly the prompt concepts can activate other concepts. 

Each prompt word (in the three-word cues used for story generation) was taken as a seed node where activation was initialised. For each seed, we ran a \texttt{SpreadPy} \texttt{BaseSpreading} model \citep{citraro2025spreadpy,siew2019spreadr} on the full network.

Given a seed node $s$, we initialised the activation vector by setting the status of all nodes to zero and assigning to $s$ an initial mass equal to the number of nodes $N$ in the network. The activation then spread iteratively until it reached approximate stationarity, which means that changes in activation between iterations became minimal. The final stationary activation of each seed node produced three scalar features per story, \(\alpha_{\mathrm{prompt}1}\), \(\alpha_{\mathrm{prompt}2}\), and \(\alpha_{\mathrm{prompt}3}\). The spreading dynamics were controlled by a retention parameter \(r \in (0,1)\), which determined how much activation was preserved at each iteration. We tested multiple values \(r \in \{0.2, 0.4, 0.5, 0.6, 0.8\}\) and re-ran the full regression pipeline for each value to assess the effect of retention on the results.
Varying the retention parameter \(r\) had only minimal impact on predictive performance across different models and network types. In line with standard practice \citep{abramski2025word,vitevitch2021cognitive,vitevitch2021exploring}, we therefore adopted the default value \(r = 0.5\) in all reported analyses. If a prompt token was disconnected from the components of the network, its corresponding \(\alpha\) value was set to the initial activation level.

\subsubsection{C) Emotion features}
We extracted emotion features from the texts using the \texttt{EmoAtlas} Python package \citep{semeraro2025emoatlas}. The z-scores for Plutchik's eight basic emotions \citep{plutchik1980general} are added as separate predictors to our regression models.
\vspace{1cm}

\section{Analysis}
This Section presents a sequence of complementary analyses. We first assess whether the seven network builders produce systematically different structural profiles for the same stories. We then characterise the behaviour of the stationary spreading-activation procedure across builders and report the global emotion profile of the corpus. Next, we evaluate the predictive contribution of different feature classes (network features, spreading-activation, and emotion scores) to creativity ratings and compare performance across builders and learning algorithms, followed by a permutation baseline to verify that observed effects reflect genuine signal rather than overfitting. Finally, we analyse SHAP values to identify the predictors driving model behaviour, both for mean creativity ratings and at the level of individual raters.

\subsection{Statistical comparison of network builders}
To test whether different builders produce meaningfully different network structures, we compared all shared network features across co-occurrence and TFMN networks using paired permutation tests \citep{good2005permutation}. For each measure, we paired observations at the level of individual stories and evaluated whether the mean difference between two builders deviated from zero under a null model generated by random sign-flips of the paired differences (10,000 permutations). This procedure provides a non-parametric, distribution-free assessment of whether two builders yield systematically higher or lower values on a given structural property. To account for the large number of feature x builder comparisons, p-values were adjusted with the Benjamini–Hochberg FDR procedure. This analysis establishes whether the observed structural profiles of the seven network variants differ reliably before examining their predictive consequences.

\subsection{Machine learning models}
\label{subsec:ml_models}

For each network builder (TFMN, co-occurrence networks with and without pronouns, and window sizes $WS \in \{2,3,4\}$), we trained a set of supervised regression models to predict the mean human creativity rating of each story from four predictor configurations: (i) structural network features only; (ii) structural features plus $\alpha$-stationary spreading-activation indices; (iii) structural features plus emotion scores; and (iv) the full combination of structural features, spreading-activation, and emotion features.

To capture a range of functional forms from simple linear relationships to highly non-linear interactions, we considered ten regression algorithms as implemented in Python (v.3.12.2) with \texttt{scikit-learn} (v.1.7.2): (1) linear regression; (2) depth-limited decision tree; (3) random forest of decision trees; (4) gradient boosting regressor; (5) $k$-nearest neighbours regressor with distance weighting; (6) bagging ensemble of decision trees; (7) AdaBoost regressor; (8) feed-forward multi-layer perceptron with one or more hidden layers; (9) stochastic gradient descent regressor; and (10) XGBoost gradient boosting model on decision trees.

Hyperparameters for all non-linear models (e.g., tree depth, number of estimators, learning rates, regularisation strengths, and hidden-layer sizes) were selected in a preliminary randomised cross-validation search. For each algorithm, we defined a model-specific hyperparameter space and drew a number of random configurations proportional to the logarithm of the size of that space, resulting in roughly 1200 trials across all models. Each configuration was evaluated with $K$-fold cross-validation on the training data (with $K=4$), using mean absolute error (MAE) as the primary optimisation criterion and Spearman correlation with human ratings as a tie-breaker. The best configuration for each algorithm (minimising MAE and, in case of ties, maximising Spearman’s $\rho$) was then fixed and reused in all subsequent analyses across network types and predictor sets. For the regression setup including structural features, spreading-activation features, and emotion variables, the selected hyperparameter configurations for each regressor are reported in Appendix~\ref{tab:hyperparams_cv4}.

For the main analyses, we used shuffled $4$-fold cross-validation for every combination of network builder, feature configuration, and regression algorithm. Predictive performance was summarised by mean absolute error (MAE) and Spearman correlations between predicted and observed creativity ratings. These features allow us to assess which of the network construction methods leads to better performances in predicting creativity ratings.

\subsection{Permutation baseline and significance testing}
To assess whether cross-validated predictive performance reflects genuine signal rather than overfitting, we constructed a permutation baseline \citep{ojala2010permutation}. For each dataset and algorithm, we generated a null model by randomly permuting all input features, while leaving the creativity ratings unchanged, and trained the same regression models on these permuted targets. The permuted models were evaluated using the identical $4$-fold cross-validation protocol applied in the main analyses, ensuring a strictly comparable evaluation procedure. This allows us to quantify how unlikely the observed performance would be under a scenario in which there are no systematic relationships between predictors and creativity. To formally compare the distributions of errors obtained under the true and permuted labels, we planned a Wilcoxon signed-rank test on mean absolute error (MAE), using the directional hypothesis $\mathrm{MAE}_{\text{real}} < \mathrm{MAE}_{\text{perm}}$ \citep{wilcoxon1945individual}.

In addition to MAE-based comparisons, we also examined whether the null models produced correlations that collapsed toward zero. For each dataset–model pair, we computed Pearson and Spearman correlations between the predictions trained on permuted labels and the true creativity ratings, and verified that their mean values were near zero, as expected under the absence of signal. This correlation sanity check ensures that the permuted baseline behaves as a genuine null model and that any positive correlations in the real-label condition cannot be attributed to artefacts of the modelling pipeline.

\subsection{Model interpretability and feature importance}
To interpret how different predictors contribute to the regressions, we used a well-established method in explainable AI (xAI), computing model-based feature importance estimates with SHAP (SHapley Additive exPlanations) \citep{lundberg2017unified,adadi2018peeking}. For each explainable model class we computed SHAP values on cross-validated predictions, as done in past works using XAI to predict creativity ratings \citep{haim2024forma}. These analyses allow us to characterise which aspects of network structure, spreading-activation, and emotion content exert the strongest and most consistent influence on predicting creativity ratings.

\subsection{Rater-specific SHAP analyses}
In addition to interpreting the models optimised for predicting the mean creativity rating, we also examined SHAP feature-attribution profiles for regressions targeting each individual rater. This rater-specific interpretability analysis provides a psychological lens on the predictive pipeline, testing whether the same network-structural, spreading-activation, and emotional features support creativity judgements across evaluators or whether different raters rely on distinct cues. For each rater, we identified the best-performing model--builder pair under the complete features setup and then computed SHAP importances for that model, enabling direct comparison of attribution patterns across raters and against the mean-rating models.

\begin{table}[t]
\centering
\resizebox{\linewidth}{!}{%
\begin{tabular}{lccccccc}
\toprule
Network builder & $n_{\text{nodes}}$ & $n_{\text{edges}}$ &
$\mathrm{ASPL}_{\text{LCC}}$ & $\mathrm{CC}$ & $d$ &
$\mathrm{D}_{\text{LCC}}$ & $C_{\mathrm{PR}}$ \\
\midrule
TFMN              & 25 $\pm$ 8 & 61 $\pm$ 30 & 2.3 $\pm$ 0.5 & 0.68 $\pm$ 0.08 & 0.21 $\pm$ 0.07 & 5 $\pm$ 1  & 0.015 $\pm$ 0.006 \\
coocc\_WS2            & 24 $\pm$ 8 & 22 $\pm$ 8  & 4 $\pm$ 1      & 0.02 $\pm$ 0.04 & 0.09 $\pm$ 0.04 & 9 $\pm$ 4  & 0.02 $\pm$ 0.01   \\
coocc\_WS3            & 24 $\pm$ 8 & 40 $\pm$ 20 & 2.5 $\pm$ 0.7  & 0.67 $\pm$ 0.07 & 0.16 $\pm$ 0.06 & 5 $\pm$ 2  & 0.02 $\pm$ 0.01   \\
coocc\_WS4            & 24 $\pm$ 8 & 52 $\pm$ 20 & 2.0 $\pm$ 0.5  & 0.78 $\pm$ 0.07 & 0.21 $\pm$ 0.08 & 4 $\pm$ 2  & 0.017 $\pm$ 0.008 \\
coocc\_p\_WS2  & 27 $\pm$ 8 & 30 $\pm$ 10 & 4 $\pm$ 1      & 0.05 $\pm$ 0.06 & 0.09 $\pm$ 0.03 & 10 $\pm$ 3 & 0.013 $\pm$ 0.007 \\
coocc\_p\_WS3  & 27 $\pm$ 8 & 54 $\pm$ 20 & 2.6 $\pm$ 0.6  & 0.64 $\pm$ 0.06 & 0.16 $\pm$ 0.05 & 6 $\pm$ 2  & 0.013 $\pm$ 0.006 \\
coocc\_p\_WS4  & 27 $\pm$ 8 & 72 $\pm$ 30 & 2.2 $\pm$ 0.4  & 0.75 $\pm$ 0.06 & 0.21 $\pm$ 0.07 & 4 $\pm$ 1  & 0.013 $\pm$ 0.005 \\
\bottomrule
\end{tabular}%
}
\caption{Mean ($\pm$ SD) structural features across network builders, averaged over all stories.}
\label{tab:mean_network_metrics}
\end{table}

\vspace{1cm}

\section{Results}

\subsection{Structural differences between network Builders}
\label{subsec:met_differences}
We first assessed whether different network construction methods produce systematically distinct structural profiles for the same stories (Table \ref{tab:mean_network_metrics}; Appendix \ref{fig:ws2_network_distributions}, \ref{fig:ws3_network_distributions} \ref{fig:ws4_network_distributions}). Paired permutation tests on all shared network features revealed reliable differences for the vast majority of builder pairs. As expected, $n_{\text{nodes}}$ was invariant across window sizes within each co-occurrence family but increased when pronouns were included. TFMNs showed $n_{\text{nodes}}$ that fell between co-occurrence networks with and without pronouns. Larger windows produced denser networks, with higher $d$, lower ASPL$_{\text{LCC}}$, higher CC, and smaller $\text{D}_{\text{LCC}}$ within each co-occurrence family. Pronoun inclusion primarily increased $n_{\text{nodes}}$ and $n_{\text{edges}}$ while leaving $d$ essentially unchanged; it also yielded slightly higher ASPL$_{\text{LCC}}$ and lower $C_{\mathrm{PR}}$ compared with pronoun-excluded networks. TFMNs produced dense, highly clustered networks with ASPL$_{\text{LCC}}$, $d$, and $C_{\mathrm{PR}}$ that were intermediate between pronoun-free and pronoun-inclusive co-occurrence networks, in line with the descriptive means reported in Table~\ref{tab:mean_network_metrics}. Beyond these invariances in node counts, only three feature–builder comparisons failed to reach significance after correction: $C_{\mathrm{PR}}$ for \texttt{co-occ\_p}\_WS2 vs \texttt{co-occ}\_p\_WS3 ($p = 0.390$), $d$ for \texttt{co-occ\_p}\_WS2 vs \texttt{co-occ}\_WS2 ($p = 0.076$), and $d$ for \texttt{co-occ}\_WS4 vs TFMN ($p = 0.960$). All remaining feature–builder pairs showed reliable differences ($p < 0.001$). 

\subsection{Spreading activations differences between network builders}
Across the full set of $1{,}071$ stories, we observed occasional isolated-seed assignments, in which a prompt token was attested in the story text but absent from the corresponding network and was therefore assigned the maximal stationary value equal to the total number of nodes ($\alpha = N$). These cases were rare but non-negligible (TFMN: $20$ isolated seeds; \texttt{coocc}\_WS2/WS3/WS4: $18$ each; \texttt{coocc\_p}\_WS2/WS3/WS4: $4$ each) and can generate extreme $\alpha$ outliers unrelated to diffusion dynamics. At the descriptive level, stationary $\alpha$ magnitudes also differed substantially by builder family (see Table \ref{tab:alpha_stationary_table}): pronoun-excluded co-occurrence networks yielded the largest stationary values (\texttt{coocc}\_WS2/WS3/WS4: $3 \pm 3$), TFMNs were intermediate ($2 \pm 3$), and pronoun-retained co-occurrence networks were smallest (\texttt{coocc\_p}\_WS2/WS3/WS4: $1 \pm 2$). Taken together, these diagnostics indicate that the stationary spreading-activation procedure is sensitive to seed availability and tokenisation (non-trivial seed failures and occasional isolated-seed assignments), injecting noise and occasional extreme $\alpha$ values that are plausibly unrelated to semantic dynamics. For an illustration of the resulting activation trajectories over time under the spreading procedure, see Appendix fig.~\ref{fig:activation_dynamics_all_builders}.

\begin{table}[t]
\centering
\begin{tabular}{@{}cc@{}}
\toprule
\textbf{Builder} & \textbf{$\alpha$-stationary} \\
\midrule
TFMN                 & $2 \pm 3$ \\
coocc\_WS   & $3 \pm 3$ \\
coocc\_WS   & $3 \pm 3$ \\
coocc\_WS   & $3 \pm 3$ \\
coocc\_p\_WS2 & $1 \pm 2$ \\
coocc\_p\_WS3 & $1 \pm 2$ \\
coocc\_p\_WS4 & $1 \pm 2$ \\
\bottomrule
\end{tabular}
\caption{Stationary spreading-activation magnitude ($\alpha$) across network builders, reported as mean $\pm$ SD.}
\label{tab:alpha_stationary_table}
\end{table}

\subsection{Emotion profile of the stories}
Emotion z-scores are extracted from the full story texts and therefore remain identical across network builders. These scores show that the corpus is characterised by higher mean intensities for \textit{anticipation}, \textit{joy}, \textit{trust}, and \textit{surprise} than for \textit{fear}, \textit{anger}, \textit{disgust}, or \textit{sadness}.
Across $N=1{,}029$ stories, the highest mean intensities were observed for \textit{anticipation} ($\bar{z}=1.64$), \textit{joy} ($\bar{z}=1.25$), and \textit{trust} ($\bar{z}=0.82$), followed by \textit{surprise} ($\bar{z}=0.56$); in contrast, negative mean values were obtained for \textit{sadness} ($\bar{z}=-0.14$), \textit{fear} ($\bar{z}=-0.47$), \textit{disgust} ($\bar{z}=-0.64$), and \textit{anger} ($\bar{z}=-0.67$). Consistent with the $z>1.96$ criterion for over-representation, \textit{anticipation} and \textit{joy} most frequently exceeded threshold (47.8\% and 37.0\% of stories, respectively), whereas under-representation ($z<-1.96$) was comparatively rare and concentrated in \textit{fear} (7.0\%) and \textit{anger} (5.2\%). Together, these results indicate that the corpus is, on average, characterised more by prospective and positive emotional language than by threat- or aversion-related affect.

\begin{table}[t]
\centering
\resizebox{\linewidth}{!}{%
\begin{tabular}{lccccccc}
\toprule
Feature configuration & coocc\_WS2 & coocc\_WS3 & coocc\_WS4 & coocc\_p\_WS2 & coocc\_p\_WS3 & coocc\_p\_WS4 & TFMN \\
\midrule
Spread         & 0.779\,$\mid$\,0.046 & 0.772\,$\mid$\,0.074 & \textbf{0.766}\,$\mid$\,0.043 & 0.783\,$\mid$\,0.070 & 0.779\,$\mid$\,0.097 & 0.781\,$\mid$\,\textbf{0.138} & 0.788\,$\mid$\,-0.008 \\
Emo+Spread     & 0.717\,$\mid$\,0.358 & 0.716\,$\mid$\,0.366 & \textbf{0.712}\,$\mid$\,0.367 & 0.719\,$\mid$\,0.367 & 0.715\,$\mid$\,0.368 & 0.715\,$\mid$\,\textbf{0.374} & 0.718\,$\mid$\,0.360 \\
Emotions       & 0.711\,$\mid$\,0.376 & 0.711\,$\mid$\,0.376 & 0.711\,$\mid$\,0.376 & 0.711\,$\mid$\,0.376 & 0.711\,$\mid$\,0.376 & 0.711\,$\mid$\,0.376 & 0.711\,$\mid$\,0.376 \\
NetStr         & 0.599\,$\mid$\,0.612 & 0.602\,$\mid$\,0.606 & 0.601\,$\mid$\,0.612 & 0.599\,$\mid$\,0.614 & 0.602\,$\mid$\,0.611 & 0.601\,$\mid$\,0.609 & \textbf{0.591}\,$\mid$\,\textbf{0.629} \\
NetStr+Spread  & 0.602\,$\mid$\,0.612 & 0.601\,$\mid$\,0.612 & 0.601\,$\mid$\,0.617 & 0.599\,$\mid$\,0.611 & 0.600\,$\mid$\,0.613 & 0.597\,$\mid$\,0.610 & \textbf{0.592}\,$\mid$\,\textbf{0.629} \\
NetStr+Emo     & 0.589\,$\mid$\,0.629 & 0.593\,$\mid$\,0.627 & 0.592\,$\mid$\,0.629 & 0.588\,$\mid$\,0.629 & 0.590\,$\mid$\,0.623 & 0.588\,$\mid$\,0.624 & \textbf{0.582}\,$\mid$\,\textbf{0.642} \\
All            & 0.590\,$\mid$\,0.629 & 0.592\,$\mid$\,0.628 & 0.591\,$\mid$\,0.629 & 0.588\,$\mid$\,0.628 & 0.589\,$\mid$\,0.626 & 0.589\,$\mid$\,0.626 & \textbf{0.581}\,$\mid$\,\textbf{0.644} \\
\bottomrule
\end{tabular}%
}
\caption{Mean absolute error (MAE) and Spearman rank correlation across feature configurations and network builders.}
\label{tab:mae_feature_sets}
\end{table}

\subsection{Predictive contributions of feature sets across builders}
\label{subsec:pred_builders}

All reported MAE values reflect averages across the full set of regression models described in the Analysis and evaluated for each feature configuration. These cross-model averages, summarised in Table~\ref{tab:mae_feature_sets}, provide a robust estimate of the predictive value contributed by each feature class.

Across all network builders, network-structural features alone provide substantial predictive power for human creativity ratings (Table~\ref{tab:mae_feature_sets}). Adding $\alpha$-stationary spreading-activation features on top of network features does not lead to systematic improvements. Although four of the seven builders show significant correlations, the MAEs in the \textit{NetStr+Spread} configuration remain effectively indistinguishable from the network-features-only baseline. The paired Wilcoxon tests corroborate this pattern, indicating no reliable performance change (mean $\Delta$MAE $\approx -0.0004$, $p = 0.812$; Table~\ref{tab:wilcoxon_feature_sets}). Emotion scores used in isolation yield substantially higher MAEs and lower Spearman's correlation than the network-features-only baseline for every builder. However, when emotion features are added to network features, they provide a small but statistically reliable improvement across builders (mean $\Delta$MAE $\approx -0.011$, $p = 0.016$). The full model combining network features, spreading-activation, and emotion scores achieves a similar improvement over network features alone (mean $\Delta$MAE $\approx -0.011$, $p = 0.016$), and its performance is indistinguishable from the \textit{NetStr+Emotions} configuration (mean $\Delta$MAE $\approx -0.0001$, $p = 0.984$), indicating that spreading-activation features do not add incremental predictive value beyond structural and emotional information.

\begingroup
\sloppy

Ablation configurations further highlight the central role of network structure: models that omit network features, whether relying on spreading-activation alone (\textit{Spread}), emotion scores alone (\textit{Emotions}), or their combination (\textit{Emotions+Spread}), all perform substantially worse than the NetStr-only baseline. On average, MAE increases by approximately $+0.18$ for \textit{Spread-only}, $+0.11$ for \textit{Emotions-only}, and $+0.12$ for \textit{Emotions+Spread}. These results indicate that neither spreading-activation nor emotion scores, individually or jointly, can replace network-structural features. Importantly, these patterns are highly consistent across all seven builders: each builder benefits similarly from adding emotion features to network features, none benefits from spreading-activation beyond this, and the relative impact of the different feature classes is effectively builder-invariant. TFMNs occupy the upper end of overall predictive performance in absolute terms. Overall, these findings suggest that human creativity ratings are systematically related to the structural organisation of the underlying semantic networks, and that network-theoretic measures provide the primary source of predictive signal for this task.

\endgroup

\begin{table}[t]
\centering
\begin{tabular}{lccc}
\toprule
Comparison & $\Delta$MAE & $W$ statistic & $p$-value \\
\midrule
NetStr $\rightarrow$ NetStr+Spread   & -0.0004 & 12.0 & 0.812 \\
NetStr $\rightarrow$ NetStr+Emo & -0.0106 &  0.0 & 0.016 \\
NetStr $\rightarrow$ All                   & -0.0107 &  0.0 & 0.016 \\
NetStr+Emo $\rightarrow$ All        & -0.0001 & 13.5 & 0.984 \\
\bottomrule
\end{tabular}
\caption{Wilcoxon signed-rank tests comparing mean absolute error (MAE) across feature configurations.}
\label{tab:wilcoxon_feature_sets}
\end{table}

\begin{figure}[!htbp]
\centering
\begin{subfigure}[t]{0.48\linewidth}
\centering
\includegraphics[width=\linewidth]{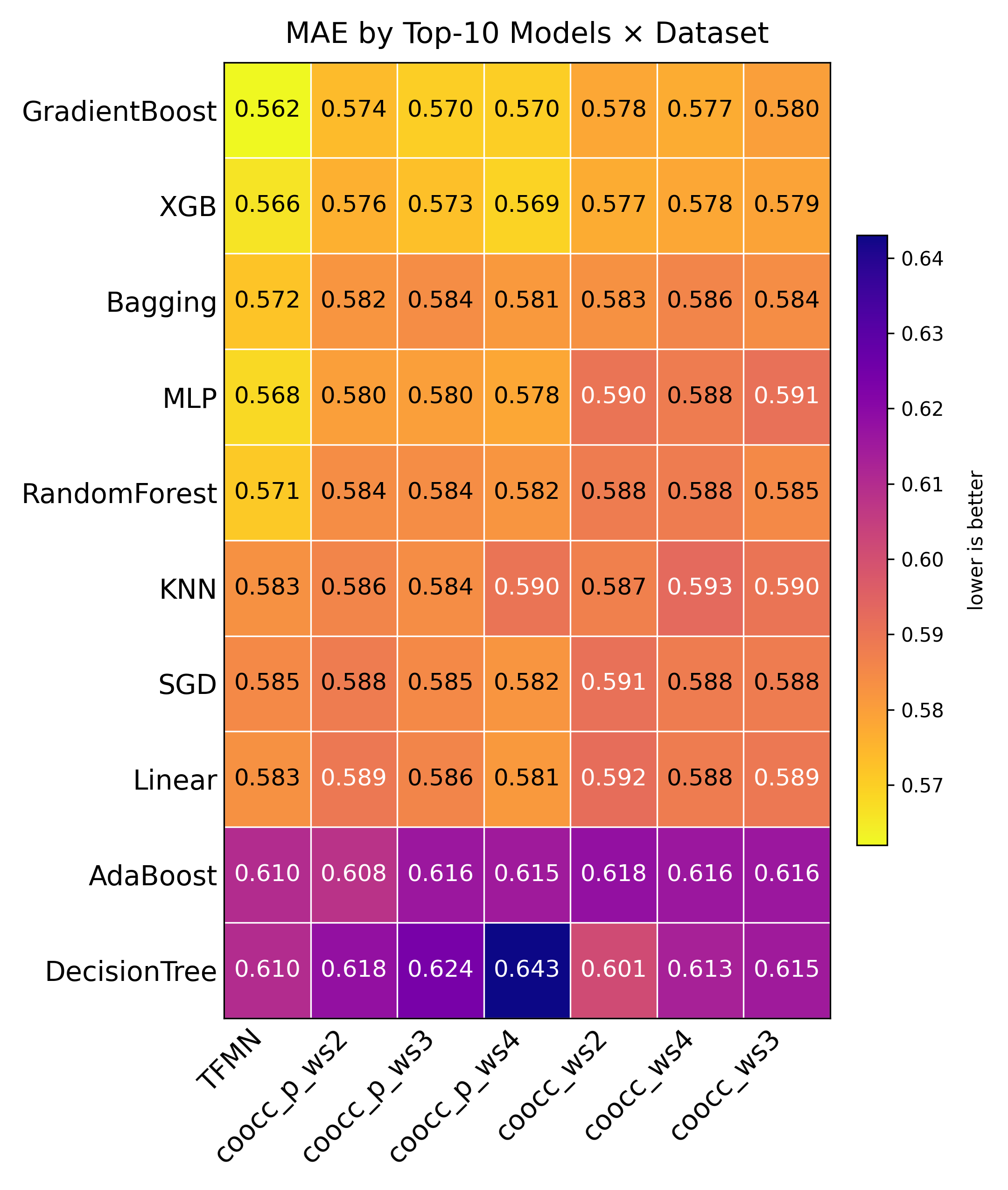}
\end{subfigure}
\hfill
\begin{subfigure}[t]{0.48\linewidth}
\centering
\includegraphics[width=\linewidth]{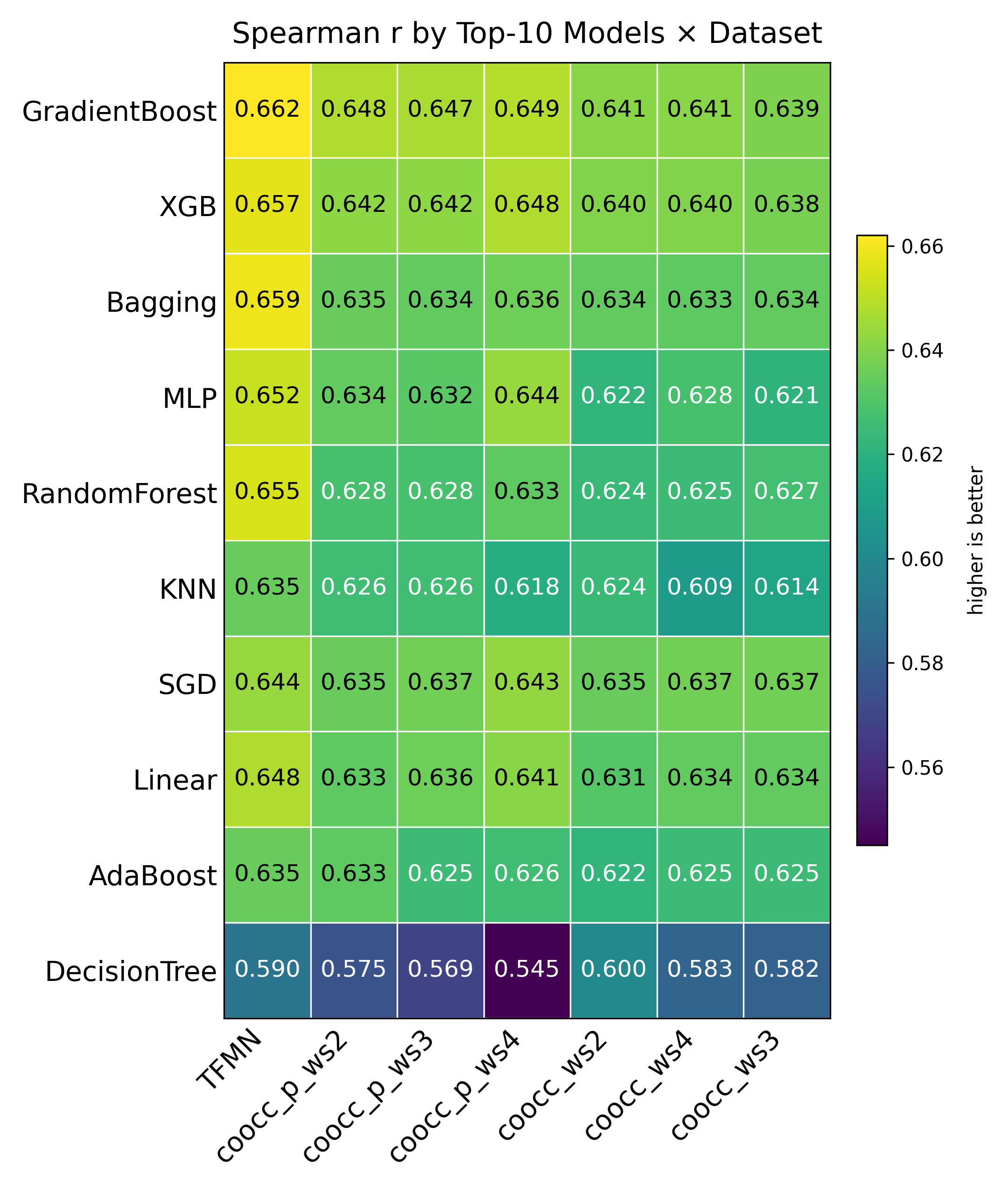}
\end{subfigure}

\caption{\textit{Predictive performance across models and network builders.}
\newline\textit{Note.} Panel (a) shows mean absolute error (MAE), where lower values indicate better predictive accuracy. Panel (b) shows Spearman rank correlations, where higher values indicate stronger association with human creativity ratings. Models (rows) and builders (columns) are ordered according to the MAE rank (lowest to highest).}
\label{fig:heatmaps_comparison}
\end{figure}

\subsection{Comparative performance of network builders under the full feature configuration}
\label{subsec:all_config}

Given that the full model, combining network features, $\alpha$-stationary spreading-activation, and emotion scores, yields the best overall predictive performance, we focus our comparison of network builders on the \textit{All} setup (\textit{Network Structure + Emotions + Spreading-activation}; Figure~\ref{fig:heatmaps_comparison}). We return to other feature sets (\textit{NetStr+Emo} and \textit{NetStr-only}) only to assess the robustness of these patterns.

Across builders, tree-based ensemble regressors systematically achieve the best scores under the \textit{All} configuration. Gradient Boosting yield the lowest mean MAE and the highest Spearman correlations, closely followed by XGBoost, MLP, Random Forest and Bagging. Linear models and SGD regressors perform slightly worse on average, while AdaBoost and single Decision Trees occupy the lower end of the performance range. Importantly, the relative ordering of builders is largely invariant across these regressors, indicating that performance differences primarily reflect properties of the underlying network representations rather than idiosyncrasies of specific learning algorithms.

In terms of builders, TFMNs consistently emerge as the best-performing representation in the setup considering all features. In the heatmaps (Figure~\ref{fig:heatmaps_comparison}), the leftmost TFMN column concentrates the lowest MAE values across almost all regressors, with Gradient Boosting achieving the single best configuration (MAE \(\approx 0.562\), Spearman \(r \approx 0.662\)). Rank correlations follow the same pattern: the brightest Spearman cells are again clustered in the TFMN column, indicating the highest alignment with human creativity ratings. 

By contrast, the co-occurrence builders cluster tightly together: within each family, differences across window sizes (WS2--WS4) are negligible, and the pronoun-handling contrast (\texttt{coocc\_p} vs.\ \texttt{coocc}) is minimal and not directionally uniform across metrics. Averaged across WS2--WS4, \texttt{coocc\_p} achieves a marginally lower MAE (mean $\approx 0.589$, range $0.588$--$0.589$) but also a marginally lower rank-correlation (mean $\rho \approx 0.627$, range $0.626$--$0.628$) than \texttt{coocc} (mean MAE $\approx 0.591$, range $0.590$--$0.592$; mean $\rho \approx 0.629$, range $0.628$--$0.629$). Practically, these values indicate that pronoun inclusion does not induce a robust performance tier; the dominant separation is instead between TFMNs and all co-occurrence instantiations.

This ranking of builders is stable across feature sets. In the \textit{NetStr+Emo}, \textit{NetStr-only} and \textit{NetStr+Spread} configurations, TFMNs again attain on average the lowest mean MAEs and highest mean Spearman correlations (\textit{NetStr} MAE \(\approx 0.591\), \(\rho \approx 0.629\);  \textit{NetStr+Spread} MAE \(\approx 0.592\), \(\rho \approx 0.629\); \textit{NetStr+Emo} MAE \(\approx 0.582\), \(\rho \approx 0.642\)). Absolute differences in MAE and Spearman across builders are understandably small, given that all story networks exhibit similar ranges of global feature values. However, the advantage of TFMN are remarkably consistent: they appear across different regressors, across all four feature configurations, and in both error-based and rank-based evaluations. This convergence suggests that pattern of structural features extracted from TFMNs captures aspects of narrative organisation that are reliably more predictive of human creativity ratings than those encoded in standard co-occurrence networks.

\subsection{Comparison with the permutation baseline}
\label{subsec:perm_baseline}

To validate that the predictive differences observed across builders do not arise from spurious correlations or model overfitting, we conducted a permutation-based control analysis. The results of this baseline confirm that the observed predictive signal is genuine. Under column-wise permutation, the models produce correlations that collapse tightly around zero (Pearson: mean $r = -0.016$; Spearman: mean $\rho = -0.019$). These values reflect negligible effect sizes and are fully consistent with the absence of any systematic structure linking the shuffled predictors to the creativity ratings. Crucially, MAEs increase substantially relative to the real-data condition, confirming that the predictive performance reported above does not arise from overfitting to noise.

A global paired Wilcoxon test \citep{wilcoxon1945individual} directly comparing the real-label MAEs with those obtained under the shuffled-predictor baseline shows a highly reliable separation: real scores are lower in every comparison ($W = 0.0$, $p < .001$, $n = 70$).


\begin{figure}[!htbp]
\centering
\begin{subfigure}[b]{0.48\linewidth}
\centering
\includegraphics[width=\linewidth]{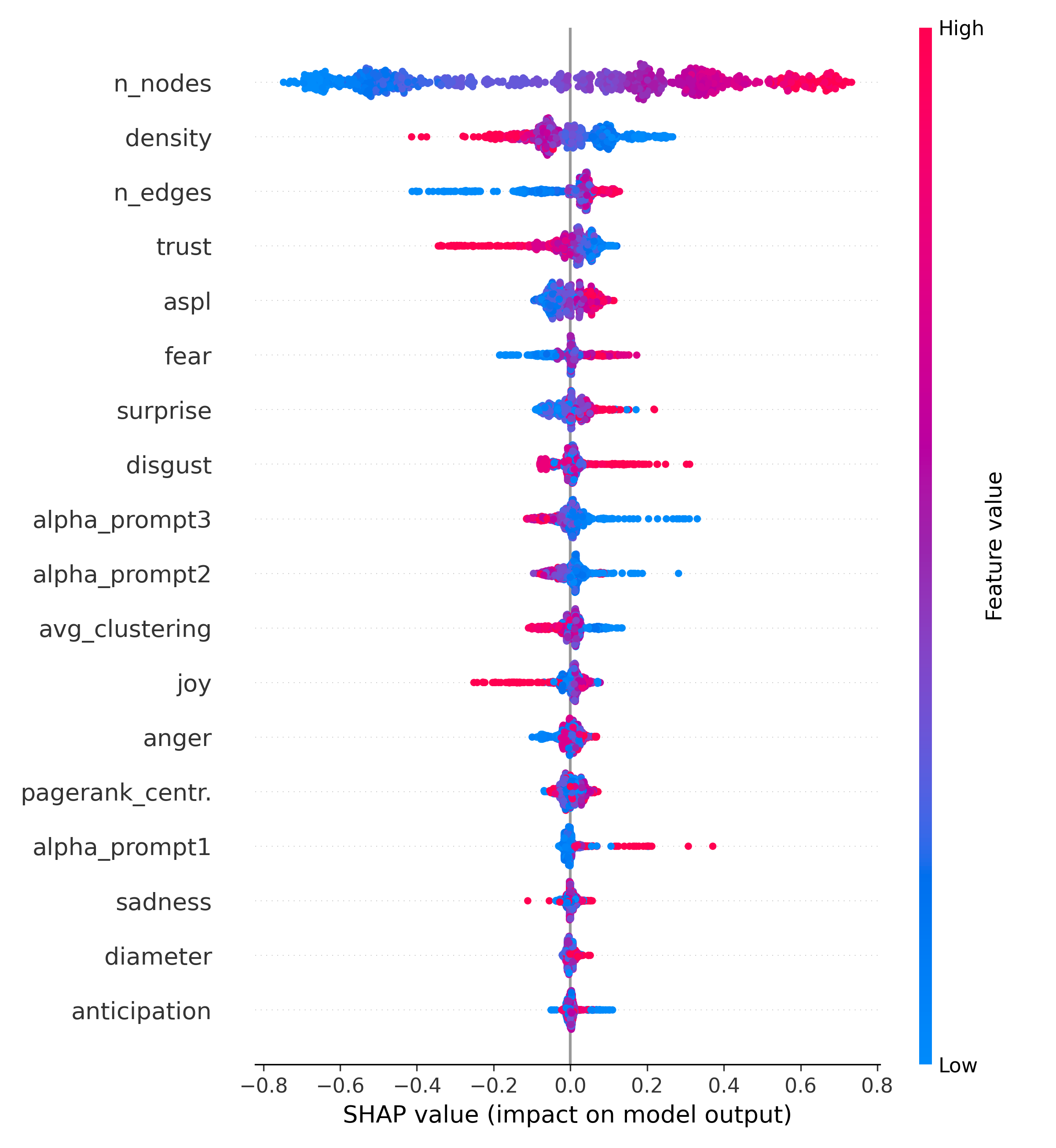}
\end{subfigure}
\hfill
\begin{subfigure}[b]{0.48\linewidth}
\centering
\includegraphics[width=\linewidth]{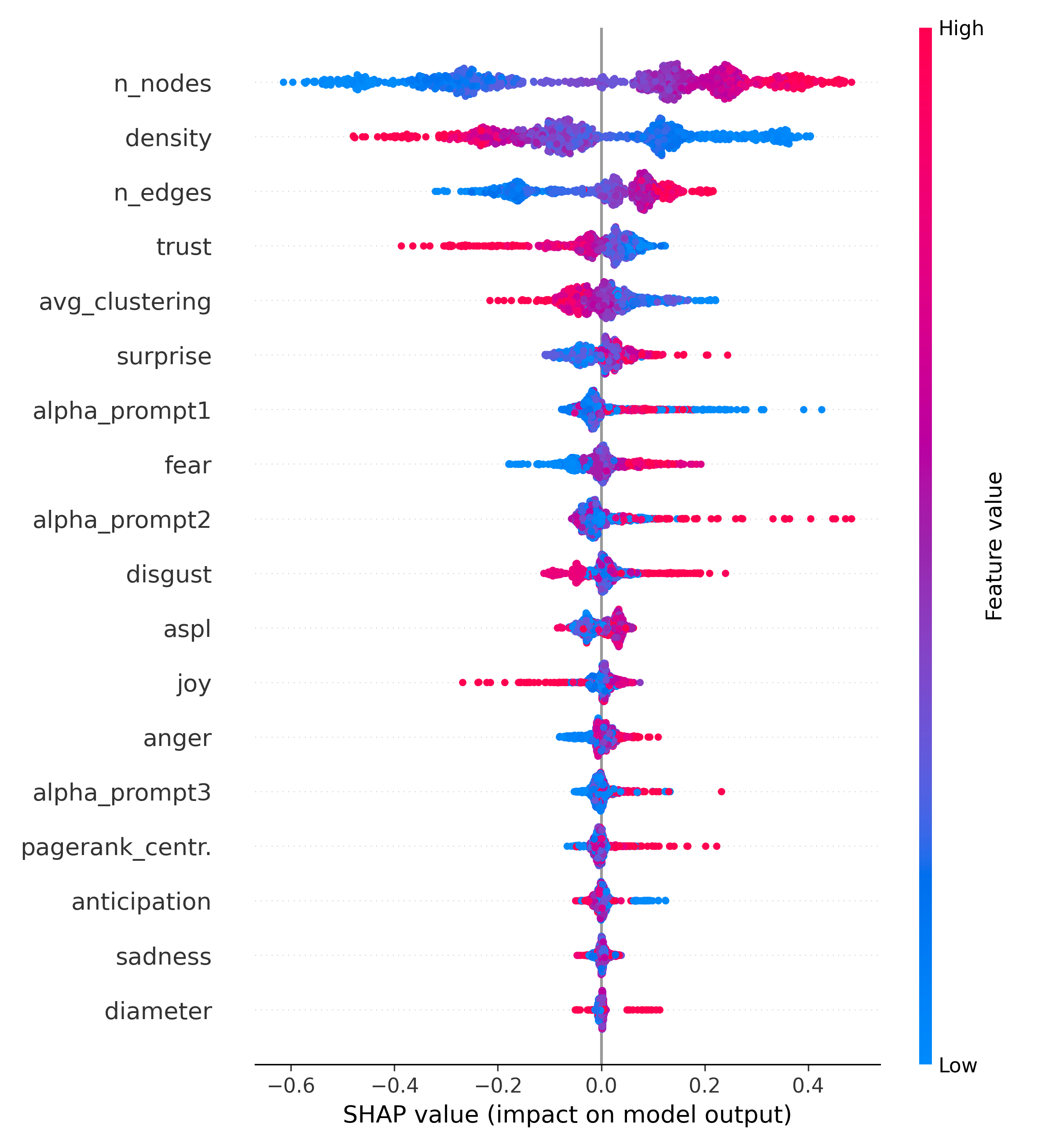}
\end{subfigure}

\caption{\textit{SHAP beeswarm plots for the best-performing models.}
\newline\textit{Note.} The left panel corresponds to the Gradient Boosting model trained on the TFMN representation. The right panel corresponds to the XGBoost model trained on the pronoun-inclusive co-occurrence network with window size 4. Colours indicate feature values, and horizontal position reflects SHAP impact on predicted creativity scores.}
\label{fig:shap_beeswarm}
\end{figure}

\subsection{Feature importance and model interpretability}
\label{subsec:SHAP}
To clarify how individual predictors influence the models, we inspected SHAP values for the two best regressors (XGB for \texttt{coocc\_p}\_WS4 and GB for TFMN, see Figure \ref{fig:shap_beeswarm}). In both cases, the dominant effects come from basic structural quantities, consistent with the performance patterns reported above. Larger $n_{\text{nodes}}$ and higher $n_{\text{edges}}$ systematically push predictions towards higher creativity, while lower density $d$ has the same effect, suggesting that large but relatively sparse networks are most strongly associated with high predicted scores. Longer average shortest path length ($ASPL$) also tend to increase predicted creativity, whereas average $CC$, PageRank centrality $C_{\mathrm{PR}}$ and diameter $D$ have only small and mostly symmetric impacts around zero. Emotion features provide a secondary modulation. The clearest signal comes from \textit{trust}, for which lower values increase predicted creativity, followed by \textit{surprise} and \textit{fear}, whose higher values are associated with higher predictions. Spreading-activation features (alpha levels) over the prompt seeds show modest importance and do not qualitatively change this picture, confirming that network order (number of nodes), network size (number of edges) and connectivity form the primary substrate of predictive power. A restricted subset of emotions, especially \textit{trust}, can improve the model only moderately. 

These patterns are generally observable for all network models (Fig. \ref{fig:shap_beeswarm}), but some differences in feature importance can be observed between TFMNs (left panel) and pronoun-inclusive co-occurrence networks (right panel). For co-occurrence networks (including pronouns, WS4) static spreading activation values (alpha prompt1-3) play a slightly more important role in predicting creativity ratings than for TFMNs. In contrast, ASPL values have higher feature importance in the TFMN model, whereas they only have medium importance in the co-occurrence network models. 

\begin{figure}[!htbp]
\centering
\begin{subfigure}[!htbp]{0.49\linewidth}
  \centering
  \includegraphics[width=\linewidth]{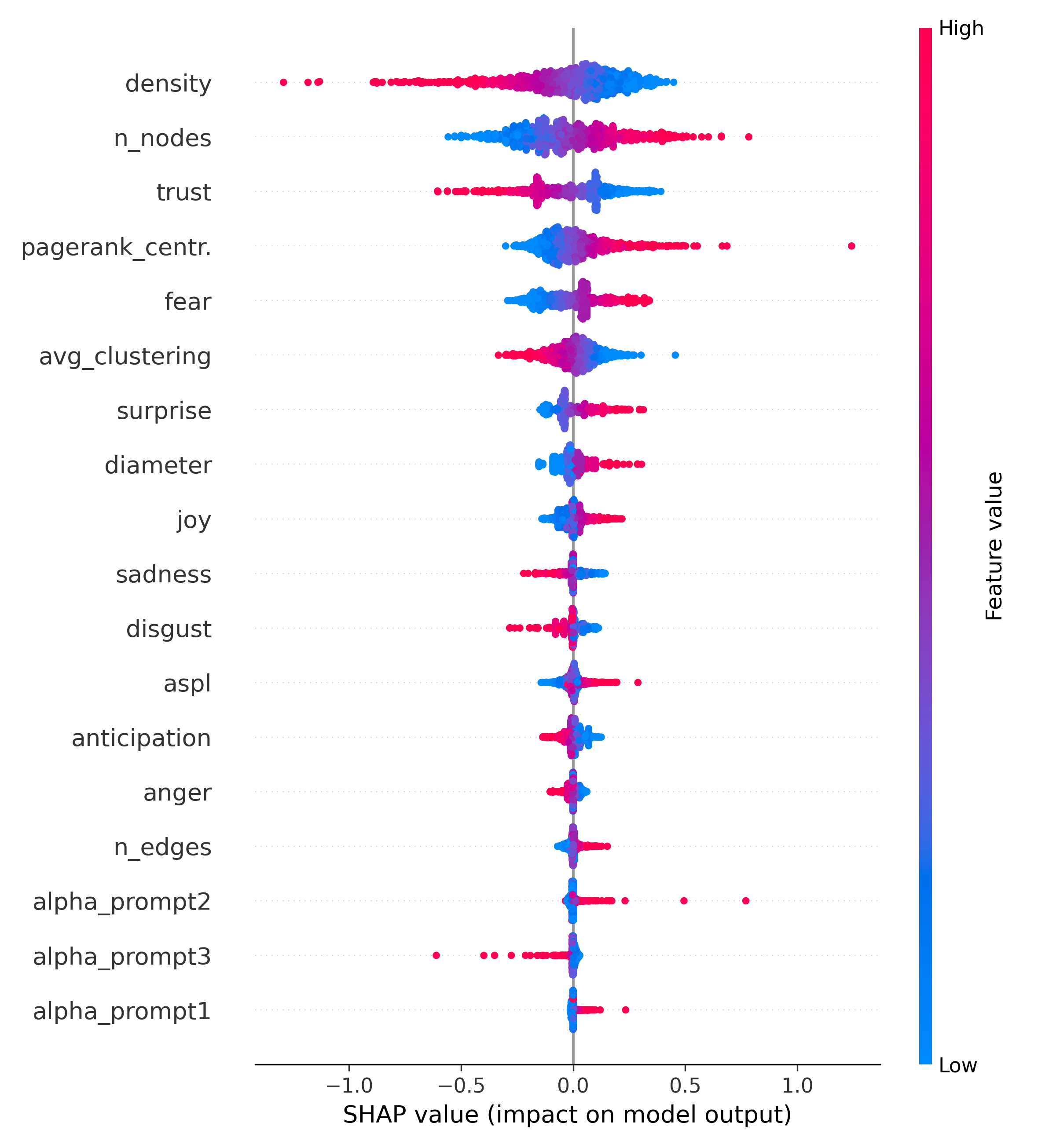}
\end{subfigure}
\hfill
\begin{subfigure}[!htbp]{0.49\linewidth}
  \centering
  \includegraphics[width=\linewidth]{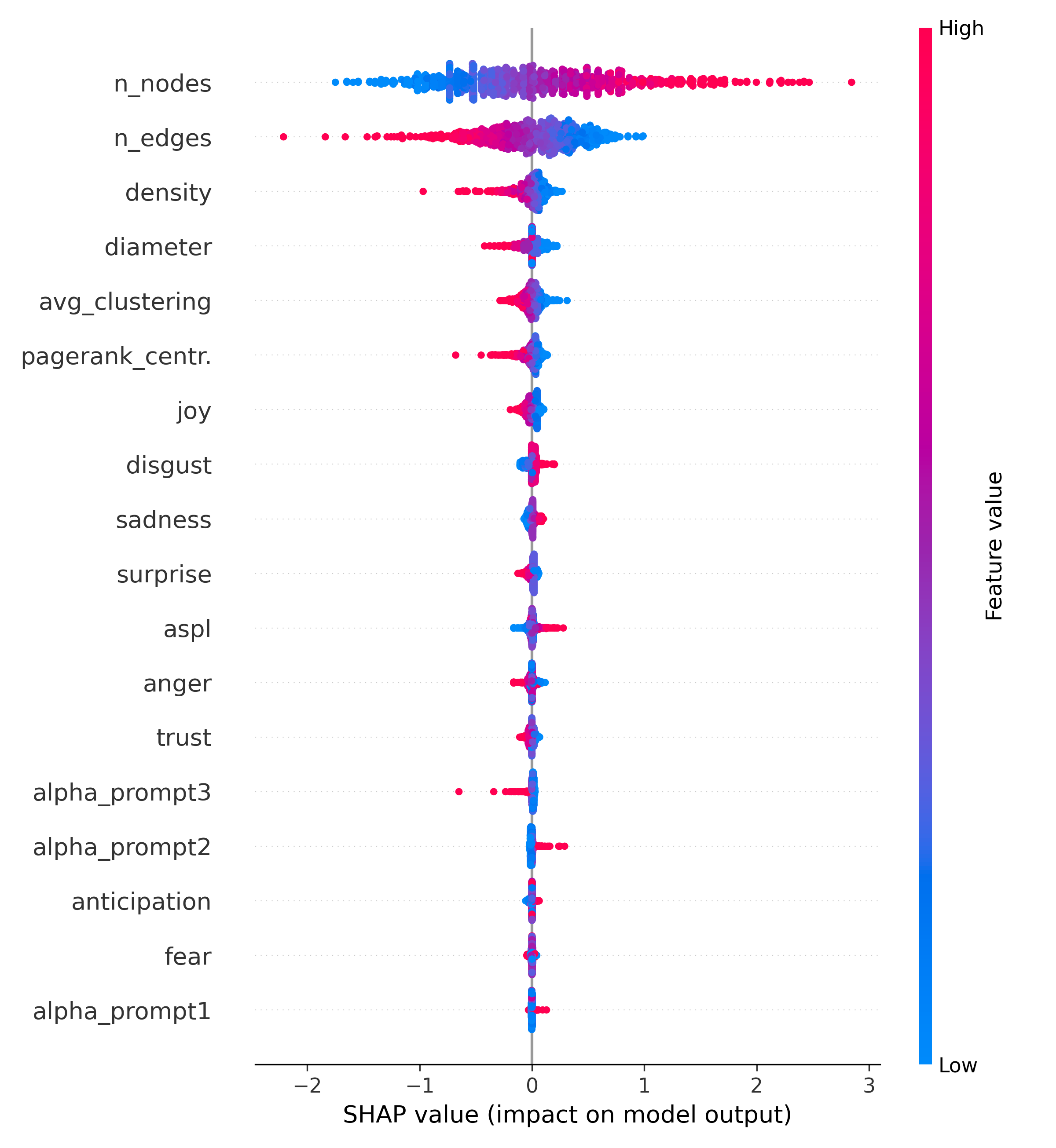}
\end{subfigure}

\vspace{0.6em}

\begin{subfigure}[!htbp]{0.49\linewidth}
  \centering
  \includegraphics[width=\linewidth]{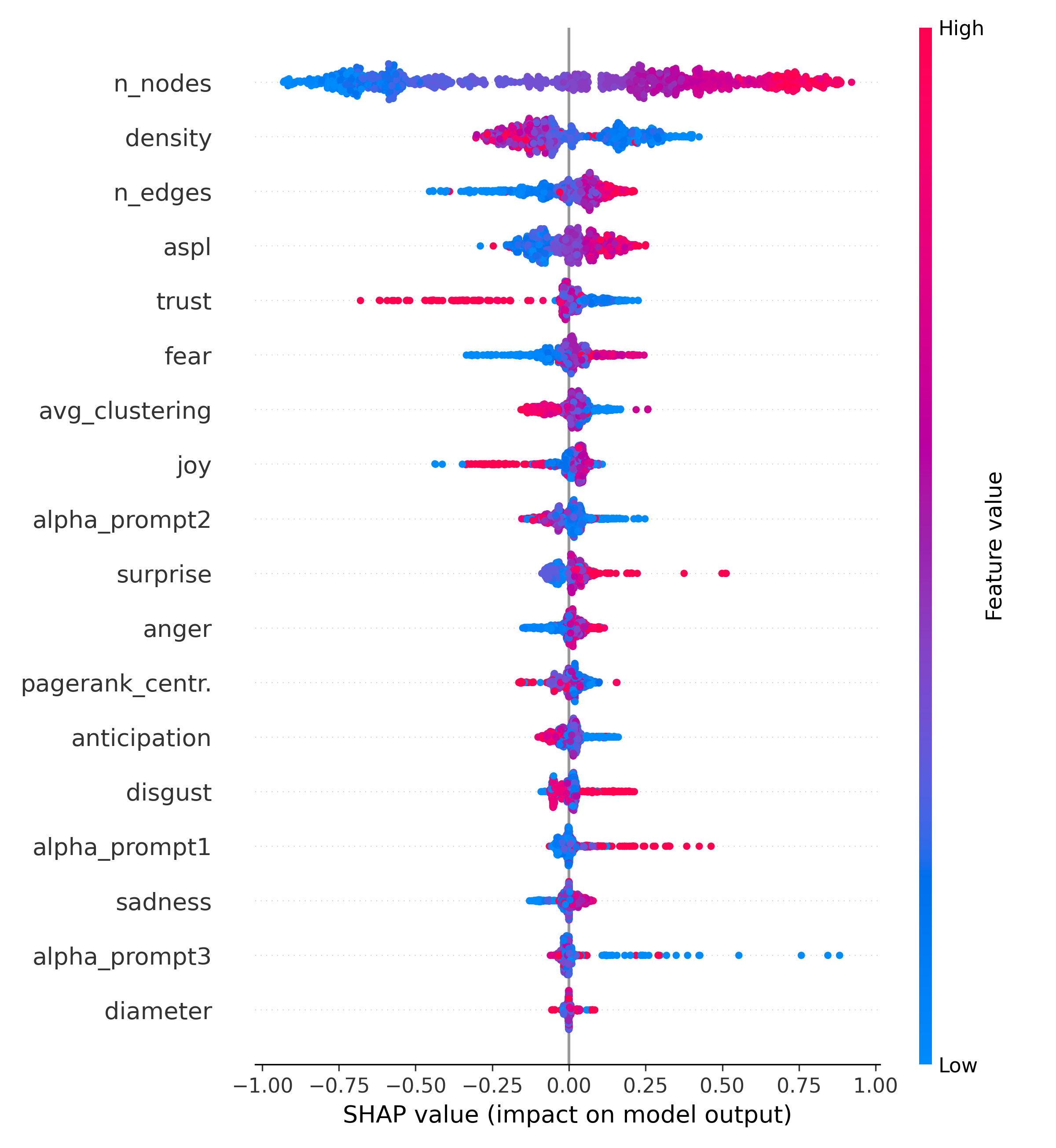}
\end{subfigure}
\hfill
\begin{subfigure}[!htbp]{0.49\linewidth}
  \centering
  \includegraphics[width=\linewidth]{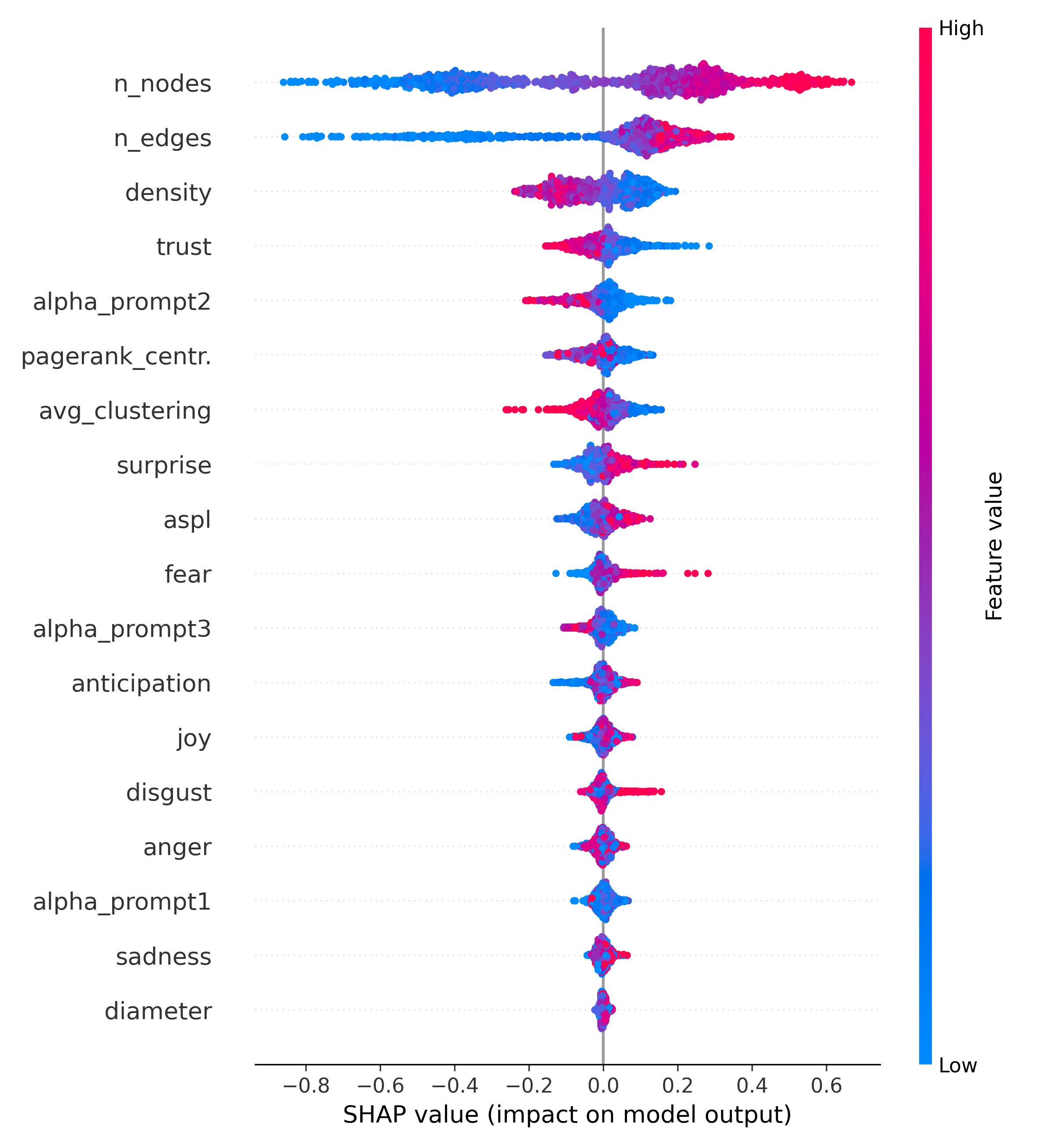}
\end{subfigure}

\caption{\textit{Rater-specific SHAP beeswarm plots for the best-performing model--builder pair per rater.} Panels show H (top-left; \texttt{coocc\_p\_WS4} with SGD regression), J (top-right; \texttt{coocc\_p\_WS4} with linear regression), K (bottom-left; TFMN with Gradient Boosting), and N (bottom-right; TFMN with Random Forest). Colours indicate feature values, and horizontal position reflects SHAP impact on predicted creativity scores.}
\label{fig:shap_raters}
\end{figure}

\subsection{Rater-specific SHAP profiles}

We can use XAI and network models to investigate individual differences in creativity judgements. To this aim, we examined SHAP feature-attribution profiles for models trained to predict the ratings of each evaluator separately (raters: H, J, K, N; Figure~\ref{fig:shap_raters}). Despite differences in the selected configurations (H/J: \texttt{coocc\_p\_WS4} with SGD/linear regression; K/N: TFMN with Gradient Boosting/Random Forest), the beeswarm plots reveal a robust shared core across raters: larger networks (higher $n_{\text{nodes}}$) consistently increase predicted creativity, whereas higher density $d$ tends to decrease it, yielding the same ``large-but-sparse'' signature observed in the mean-rating models. A second consistent pattern concerns affect: \textit{trust} shows a clear negative direction in three of the four raters (H, K, N), with higher \textit{trust} pushing predictions downward and lower \textit{trust} pushing them upward, while \textit{fear} and \textit{surprise} generally contribute in the opposite direction (higher values associated with higher predictions), most clearly for H, K and N. In contrast, emotion features play a markedly reduced role in the model fitted to rater J: The six most influential predictors in this case are exclusively structural network measures, indicating that this rater’s judgements are captured almost entirely by variations in network size and connectivity, with affective content contributing little to the predictive signal. 

Raters also differ in how connectivity-related quantities are weighted: in J, $n_{\text{edges}}$ exhibits a pronounced negative direction (more edges $\rightarrow$ lower predictions), consistent with this feature acting as a density proxy once $n_{\text{nodes}}$ is accounted for, whereas in K and N $n_{\text{edges}}$ is weakly positive. 

Spreading-activation features remain secondary across all raters, with the exception of $\alpha_{\text{prompt2}}$ for K and N, which show more systematic negative contributions (higher $\alpha$ associated with lower predictions), suggesting that these variables may partly track seed availability rather than clear psychologically interpretable patterns. 

Overall, rater-specific SHAP profiles support the conclusion that the most stable predictive substrate is network order (total number of nodes) and sparsity, while individual differences in judgement appear mainly in the secondary modulation by emotion features.

\vspace{1cm}

\section{Discussion}

This paper provides both a tutorial introduction and an empirical comparison of two approaches for constructing text-based semantic networks. We compare standard word co-occurrence networks \citep{amancio2015complex} with syntactically grounded textual forma mentis networks (TFMNs) \citep{stella2020text,semeraro2025emoatlas} as representations of associative structure in creative stories. In the following Subsections we discuss how our tutorial contributes to enhancing transparency and accessibility of these two methods, and provide an empirical evaluation of how these network construction choices affect predictive modelling of creativity ratings.

\vspace{1cm}

\subsection{Tutorial}
The Tutorial Section of this paper provides a practical, reproducible workflow for constructing text-based networks, while highlighting the underlying methodological choices that shape network structure. Methodological decisions such as whether to retain pronouns, how to select window sizes in co-occurrence networks, or whether to link words based on surface adjacency (as in co-occurrence networks) or syntactic dependency (as in TFMNs) can substantially alter network topology. Yet, these choices are often embedded in software details or reported only briefly, making it difficult to adapt and reproduce methods. This tutorial addresses this by providing a transparent workflow that discusses each decision while relying on open, well-documented tools. 

Furthermore, our tutorial aimed to lower the barrier to entry for researchers with limited experience in programming or network science. We do this by clearly separating what users must actively specify in the code from what is automatically handled by libraries such as \texttt{spaCy}, \citep{montani2023explosion} \texttt{NetworkX} \citep{hagberg2008exploring}, and \texttt{EmoAtlas} \citep{semeraro2025emoatlas}. To support reusing the methods described here, all code and notebooks are openly available on our \href{https://osf.io/5cn2y/overview}{OSF page}. 

At the same time, this tutorial is designed to be informative for more experienced users who wish to better understand the assumptions underlying the network construction strategies discussed here. Making informed choices about parameter settings helps in adapting and extending these methods to specific datasets and new research questions.

Finally, we illustrate how predictive modelling can be used not only to assess whether creativity ratings can be predicted from network features, but also to examine which structural and emotional properties contribute most strongly to those predictions. Thus, we show how to use predictive modelling not only as black-box predictors, but as informative and interpretable models explaining observed patterns between features and creativity rating.

However, our workflow is not exclusive to this specific domain of predicting creativity judgements. The same pipeline of network construction and feature extraction can be used to compare semantic networks across groups or in a longitudinal design, to identify central concepts and their neighbourhoods in a text, or to quantify affective tone. 

Taken together, this tutorial offers a reusable methodological template that advances accessibility, transparency, and interpretability in text-based network research.

\vspace{1cm}
\subsection{Empirical results}
Building on this workflow, we evaluated how these network representations perform when used to predict human creativity ratings. Using 1029 short narratives, we trained regression models on static network measures \citep{stella2020text}, prompt-seeded spreading-activation indices \citep{citraro2025spreadpy}, and emotion z-scores \citep{semeraro2025emoatlas} to predict creativity ratings.

In the full-predictor setting, TFMNs showed the strongest overall performance when aggregating across algorithms (mean MAE $\approx0.581$, mean Spearman $\rho\approx0.644$). Co-occurrence builders, instead, formed a tight cluster: variation across window sizes (WS2--WS4) was negligible, and the pronoun-handling manipulation did not yield a stable performance tier. Averaged across window sizes, \texttt{coocc\_p} which includes pronouns showed a marginally lower MAE (mean $\approx 0.589$, range $0.588$--$0.589$) but also a marginally lower rank correlation (mean $\rho \approx 0.627$, range $0.626$--$0.628$) than \texttt{coocc} which excludes pronouns (mean MAE $\approx 0.591$, range $0.590$--$0.592$; mean $\rho \approx 0.629$, range $0.628$--$0.629$). In practice, these differences between co-occurrence builders are minimal and direction-dependent across evaluation criteria. The dominant separation is found between syntactically grounded TFMNs and all window-based co-occurrence builders, including the conventional pronoun-filtered co-occurrence baseline.

Static network topology emerged as the dominant source of predictive signal (NetStr-only: mean MAE $\approx 0.599$, mean $\rho \approx 0.613$ across builders), strongly outperforming both emotion-only models (MAE $\approx 0.711$, $\rho\approx 0.376$) and spreading activation-only models (mean MAE $\approx 0.778$, mean $\rho \approx 0.066$). Adding emotional features to the topology model provided a small but reliable gain over topology-only ($\Delta$MAE $= -0.0106$, $p = 0.016$), whereas spreading-activation features did not contribute beyond topology ($\Delta$MAE $= -0.0004$, $p = 0.812$). Adding spreading activation features on top of NetStr+Emo yielded no further improvement ($\Delta$MAE $= -0.0001$, $p = 0.984$). 

Higher creativity ratings were associated with networks that were larger, structurally sparser, and more globally distributed, combined with emotional profiles characterised by lower trust and elevated tension (fear/surprise). These patterns are consistent with the individual-rater analysis. They further suggest that, in short narrative writing, creativity is better captured by network structure, specifically broad conceptual expansion and global coherence, than by dense local associations or affect alone. 

A key question is why TFMNs outperform co-occurrence networks even when the latter can reach comparable average sizes and densities (Table~\ref{tab:mean_network_metrics}). For these 4–6 sentence long stories, all builders produce relatively small networks (averaging the per-builder means: $n_{\text{nodes}} \approx 25.4$, $n_{\text{edges}} \approx 47.3$), and pronoun-inclusive co-occurrence variants can match or exceed TFMNs in node and edge counts (nodes: $27$ vs.\ $25$, $\Delta=+2$; edges: $72$ vs.\ $61$ in \texttt{coocc\_p}\_WS4, $\Delta=+11$). This indicates that the gap in predictive performance cannot be attributed to simple scalar properties such as network size or density. Instead, it reflects systematic differences in how each representation distributes edges and encodes relational structure. 

Co-occurrence builders link words that fall within a sliding window, which tends to produce less fragmented (in other words more connected) networks when pronouns are retained than when they are removed (mean number of components: \texttt{coocc\_p}\_WS2/WS3/WS4 $= 1.395$ vs.\ \texttt{coocc}\_WS2/WS3/WS4 $= 2.552$ vs. TFMN: $= 1.458$). A higher number of components indicates stronger fragmentation into disconnected subnetworks, whereas a single component indicates a fully connected network. In contrast, TFMNs connect words along syntactic dependencies, linking subjects, predicates, modifiers, and objects even when they are distant in surface order (Figure \ref{fig:graphs}). Through this approach, TFMNs typically are fully connected graphs with no disconnected components. This difference suggests a useful asymmetry: window-based co-occurrence networks can either under-connect key concepts (fragmenting the narrative scaffold) or over-concentrate connections in local clusters, whereas TFMNs more consistently link conceptually relevant elements via grammatically licensed relations. SHAP beeswarm analyses (Figure~\ref{fig:shap_beeswarm}) confirm this: stories judged as more creative are those whose networks combine richer elaboration (more nodes and edges) with more globally distributed connectivity and reduced local clustering, in line with evidence that elaboration and conceptual diversity support originality judgements \citep{haim2024forma,forthmann2019application}. At the same time, our models associate higher creativity with longer average path lengths. In short narratives, longer paths may reflect broader thematic reach and more complex syntactic organisation rather than inefficient connectivity, consistent with the global, grammar-constrained integration captured by TFMNs. Taken together, these observations indicate that TFMNs instantiate a configuration of structural properties that is particularly well aligned with the network topology our models associate with creativity.

\vspace{1cm}

\subsection{Limitations and future research}

A few limitations of this study warrant consideration. First, the narratives analysed were short (4–6 sentences), reflecting a common design in the creativity literature where rated story corpora often rely on the five-sentence creative story task \citep{johnson2023divergent,bianchi2025creativity,ismayilzada2024evaluating}. This short-text regime constrains the structural properties that co-occurrence networks can capture. In particular, removing pronouns from short texts substantially reduces node counts and increases fragmentation of co-occurrence networks. This provides a disadvantage for pronoun-excluding co-occurrence networks relative to pronoun-including variants and TFMNs. Future work should explicitly test whether the observed sparsity and performance gap for pronoun-excluding co-occurrence networks narrows when investigating longer narratives. 

Second, our conclusions are tied to the specific feature set examined. We focused on standard topological descriptors used in previous creativity research \citep{haim2024forma}, for the sake of comparability. However, alternative network measures, such as modularity or assortativity, could capture different aspects of narrative organisation and might modulate the relative contributions of each network builder \citep{newman2002assortative,newman2004finding,fortunato2007resolution,fortunato2010community}. Future studies could explore the influences of a broader or different set of predictive features, aligned with the specific research questions. 

Third, regression performance reflects correlations present in this specific corpus and annotation protocol. Creativity ratings were obtained from four raters, instructed to assess the creativity, vividness, and emotional engagement of the stories \citep{johnson2023divergent}. Different text genres or rating schemes may result in different relationships between network structure and perceived creativity \citep{patterson2025cap,amabile1982social,hennessey1994consensual}. Accordingly, future studies should apply the workflow presented here to alternative datasets. 

Future work could extend these analyses to broader creative contexts. Longer narratives, unconstrained story-writing tasks, and multi-author formats such as a paired Woseco sentence-chain paradigm would allow evaluation of whether the structural patterns observed here generalise to texts with different lengths, stylistic variability, and collaborative dynamics \citep{haim2024word}. In parallel, it would be valuable to examine whether the advantages of TFMNs observed in creativity prediction also transfer to other semantic-network tasks. In addition, our current approach relies on manually selected network features, chosen to balance interpretability with predictive power. An alternative direction is to apply graph neural networks (GNNs) directly to the different network topologies, allowing models to learn task-relevant representations without requiring predefined measures \citep{yuan2022explainability,wu2020comprehensive,ying2019gnnexplainer}. Although such approaches may reduce interpretability, they offer a promising route for discovering higher-order structural patterns that traditional feature engineering might overlook.

\vspace{1cm}

\section{Conclusion}
This paper provided a tutorial and an empirical evaluation of co-occurrence  and textual forma mentis networks to support transparent and interpretable use of semantic networks. 
We presented a step-by-step workflow for constructing text-based semantic networks from short narratives. We further guided through the steps for extracting structural, spreading-activation, and emotional features from the stories, and show how to integrate these features into machine learning models for predicting creativity levels. By explicitly contrasting co-occurrence networks and TFMNs at each stage, the tutorial provides a framework that can be readily adapted to new datasets and research questions.

Overall, the results suggest that, in short creative writing, creativity judgments depend primarily on global patterns of conceptual organisation. Dependency-based TFMNs capture these patterns more reliably than surface co-occurrence models, making them a particularly suitable representation for the analysis of creative texts. Higher creativity ratings were associated with stories relative to larger and more globally distributed networks of semantic/syntactic associations, rich in emotions like trust, fear and anticipation.

\section{Declarations}

\begin{itemize}
    \item Funding: This work received no external funding.
    \item Conflicts of interest: MS is an associate guest editor of Behavior Research Methods but had no involvement in the review process or the editorial decision for this manuscript. The other authors disclose no conflict of interest.
    \item Ethics approval and consent to participate: Exempt. This work was based on secondary data analyses of fully anonymised datasets acknowledged in the Methods.
    \item Consent for publication: The authors agreed on the current manuscript version.
    \item Availability of data and materials: The data and materials used for the applied part of this tutorial are available online in an \href{https://osf.io/5cn2y/overview}{OSF repository}, see Methods.
    \item Code availability: The code developed in Python for the applied part of this tutorial is available online in an \href{https://osf.io/5cn2y/overview}{OSF repository}, see Methods.
    \item Authors' contributions: Study design - All authors; Data Analysis - RP; Data Visualisation: RP and EH; Formal analysis - EH; Validation - RP and EH; Supervision - MS; Write up - All authors.
\end{itemize}

\printbibliography

\appendix
\section{Distributions of network features}

\begin{figure}
\centering

\includegraphics[width=0.40\textwidth]{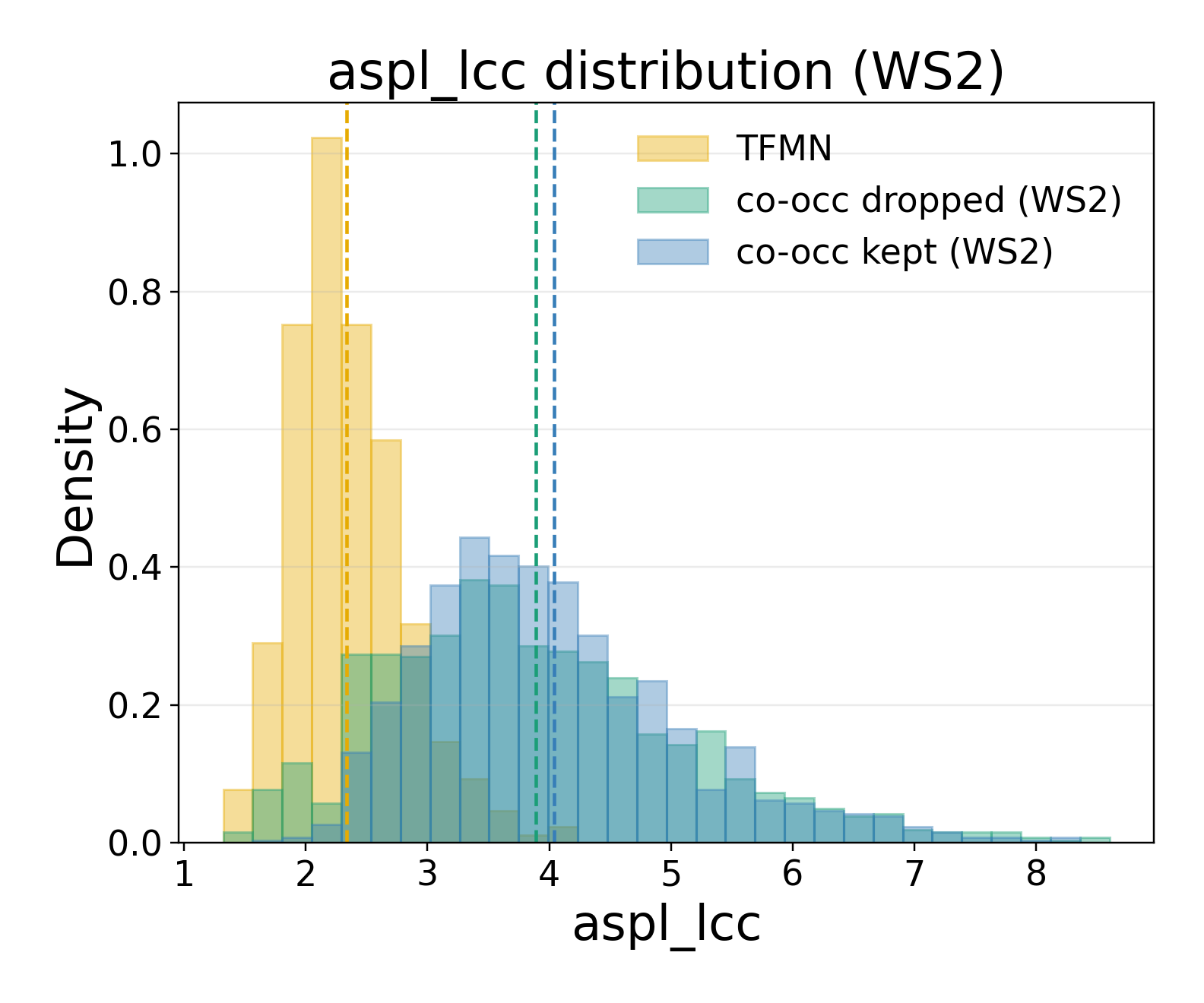}\hfill
\includegraphics[width=0.40\textwidth]{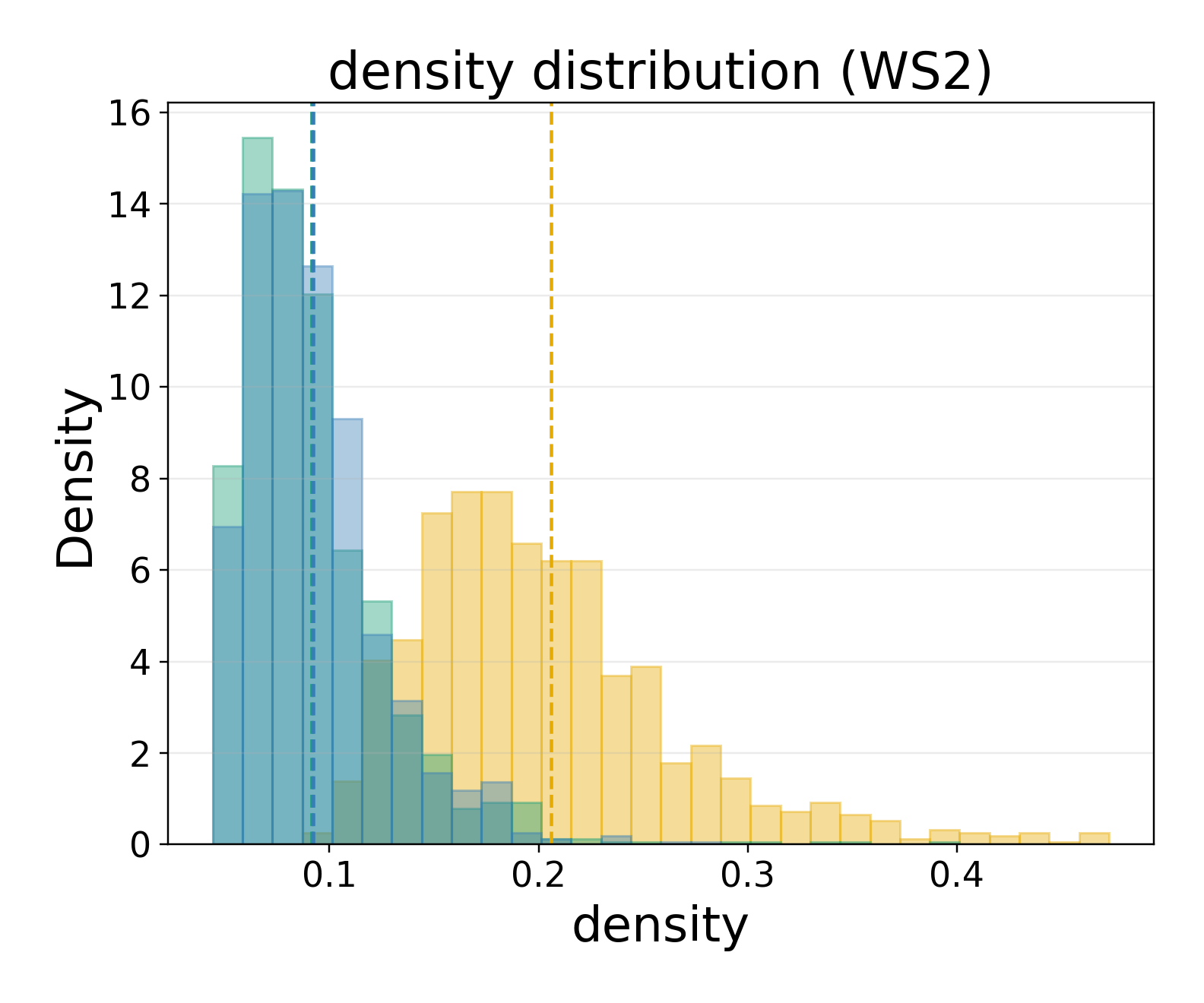}

\includegraphics[width=0.40\textwidth]{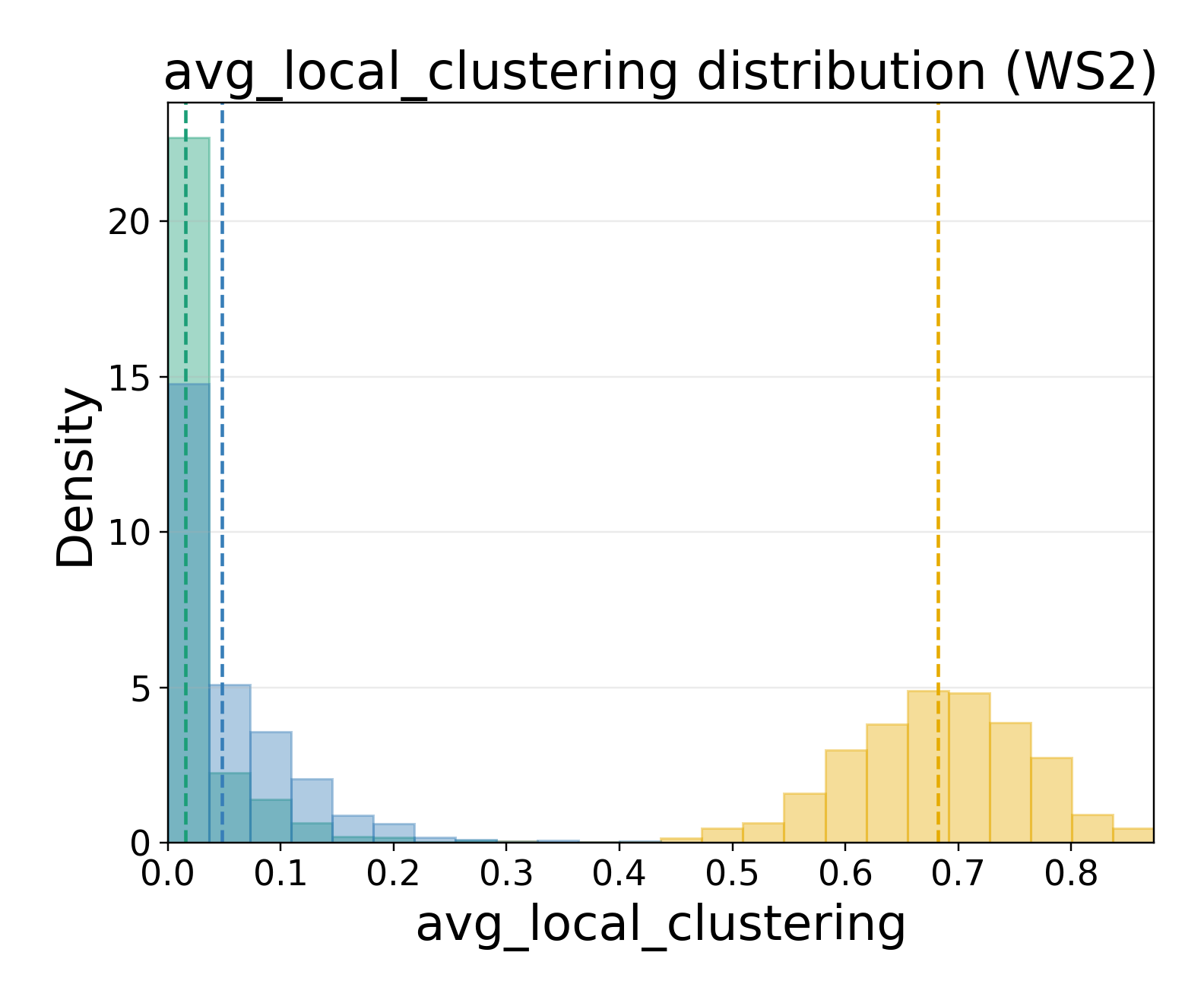}\hfill
\includegraphics[width=0.40\textwidth]{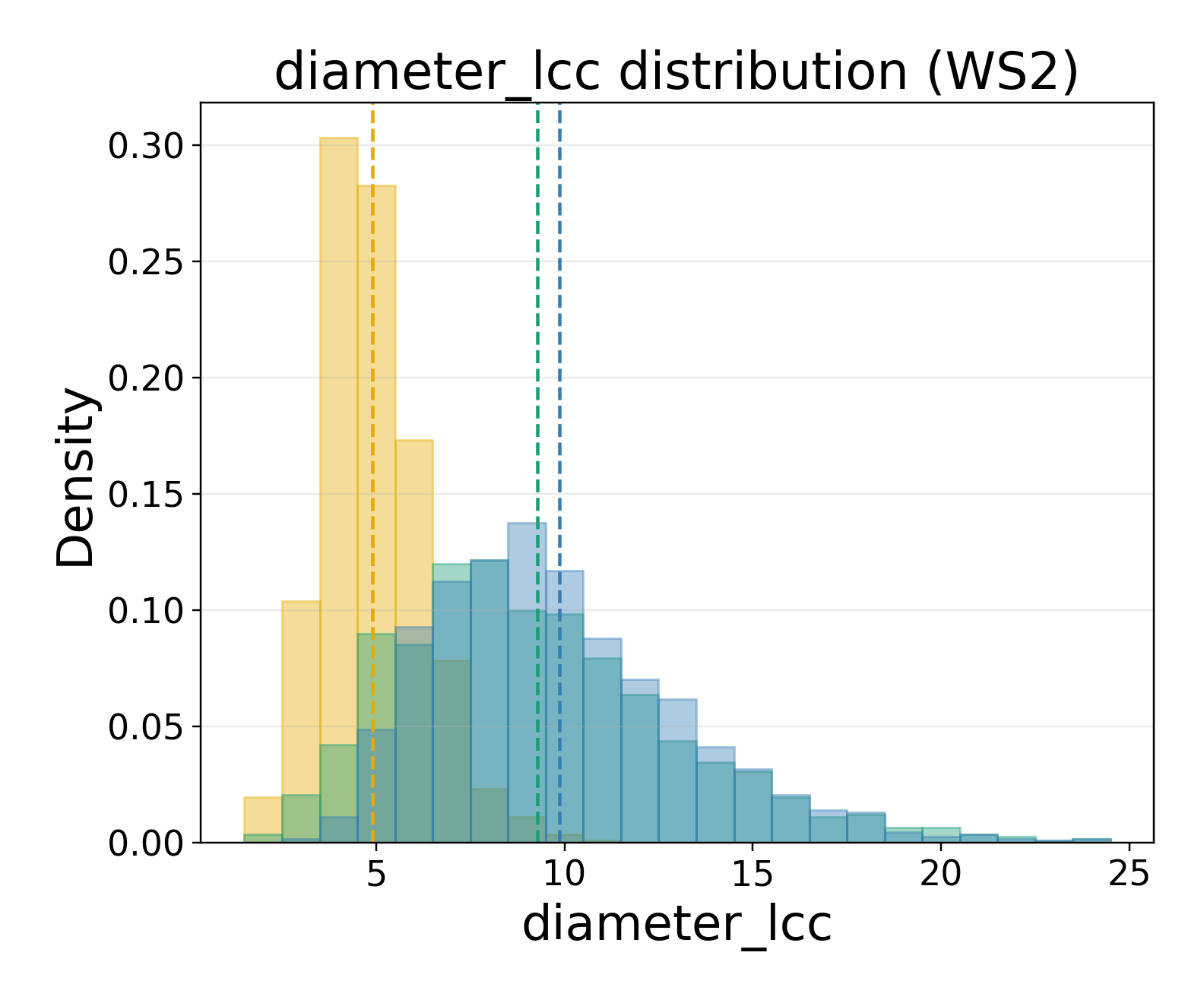}

\includegraphics[width=0.40\textwidth]{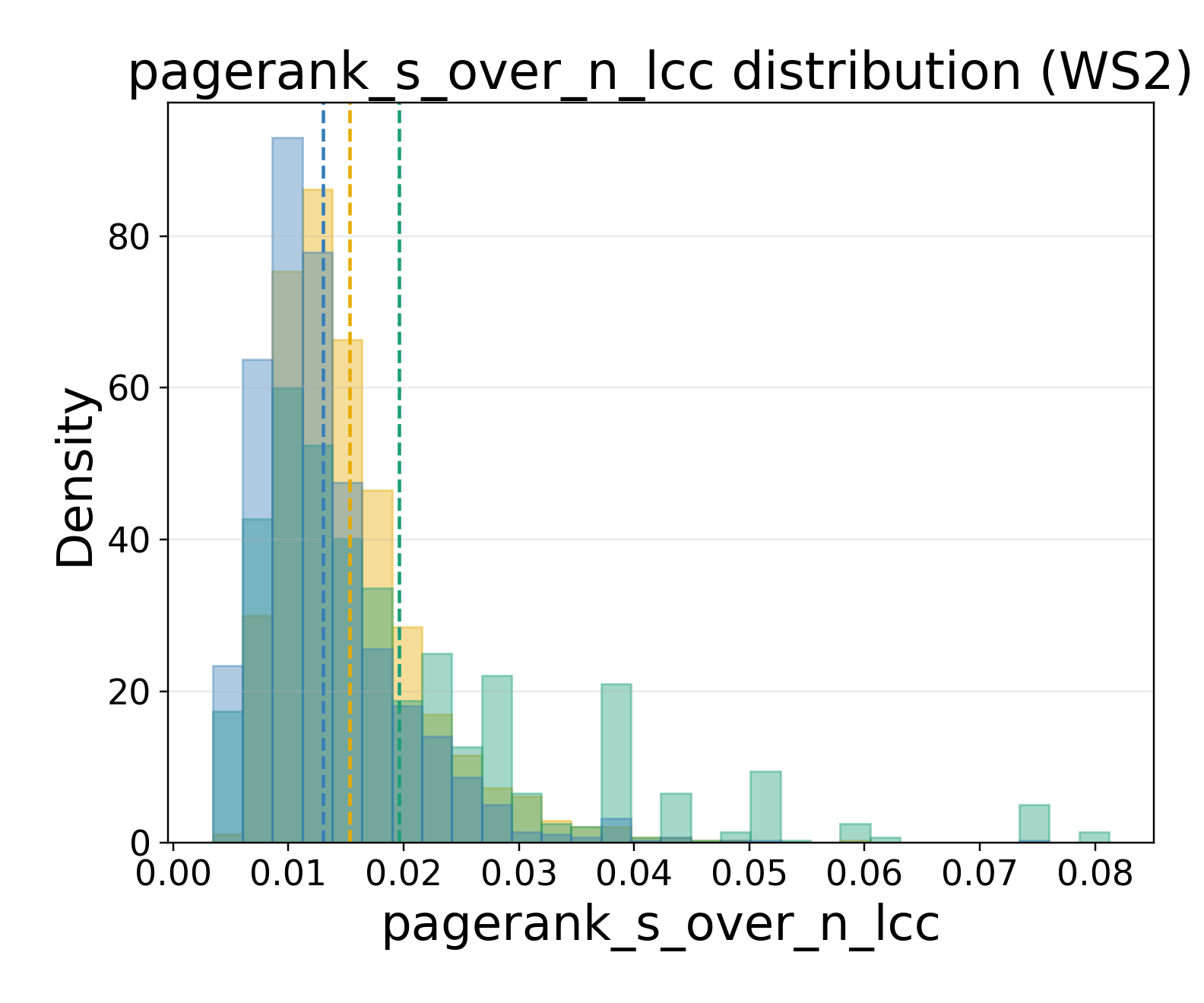}

\caption{
Distributions of key network-structural features for the WS2 co-occurrence setting.
For each feature, distributions are shown for TFMN networks, co-occurrence networks with pronouns retained, and co-occurrence networks with pronouns removed; dashed vertical lines indicate the corresponding mean values.
}
\label{fig:ws2_network_distributions}
\end{figure}

\begin{figure}
\centering

\includegraphics[width=0.40\textwidth]{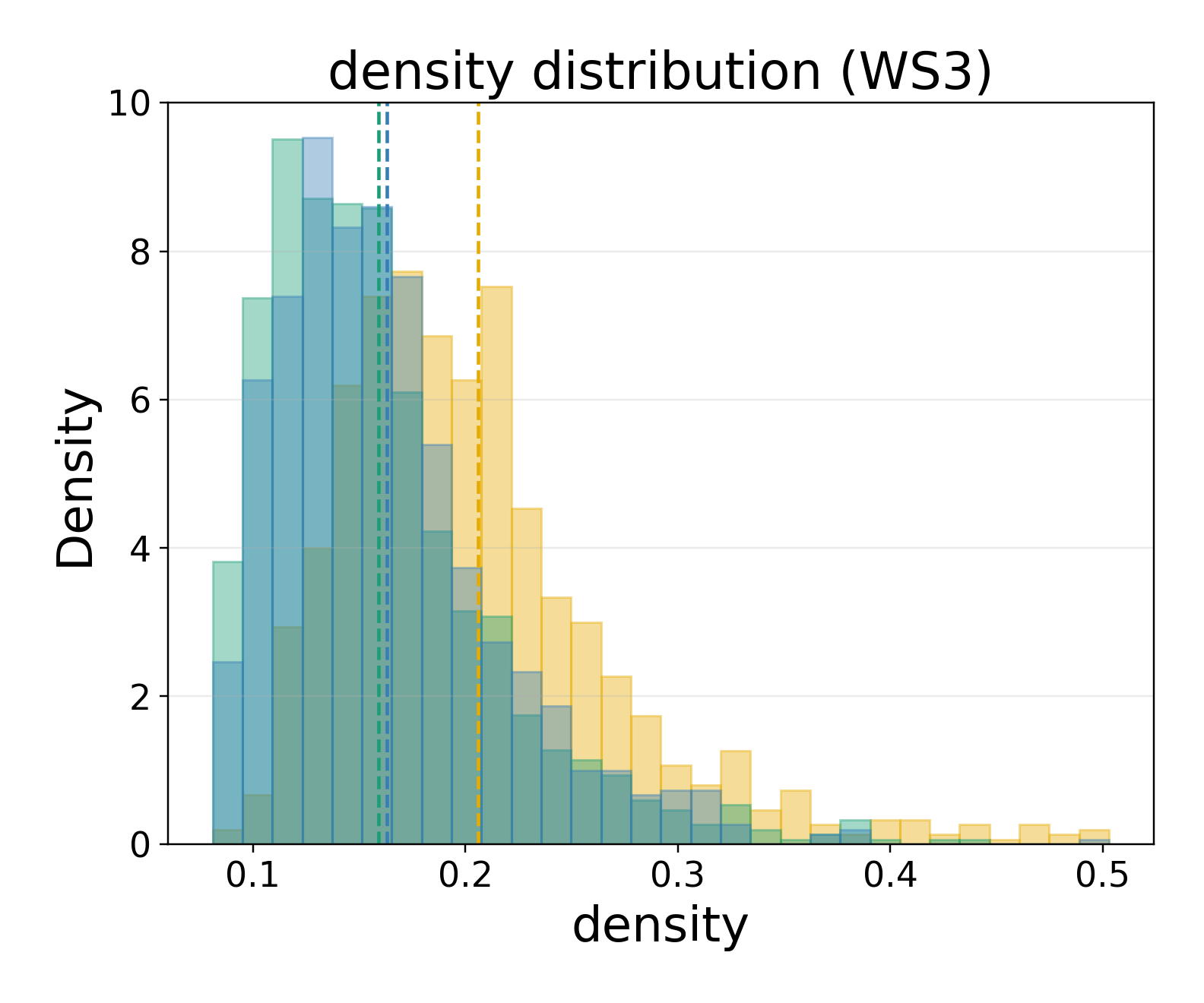}\hfill
\includegraphics[width=0.40\textwidth]{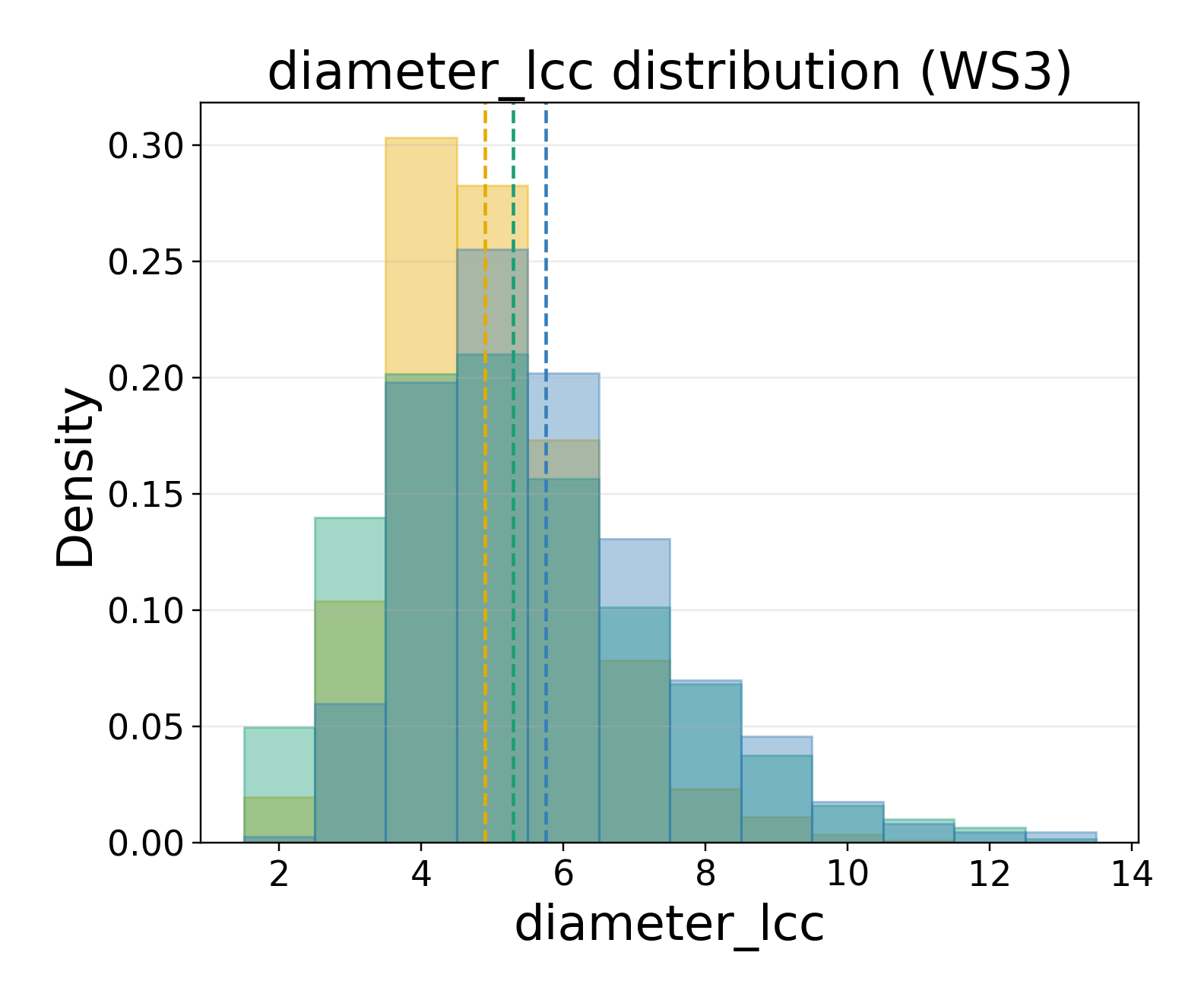}

\includegraphics[width=0.40\textwidth]{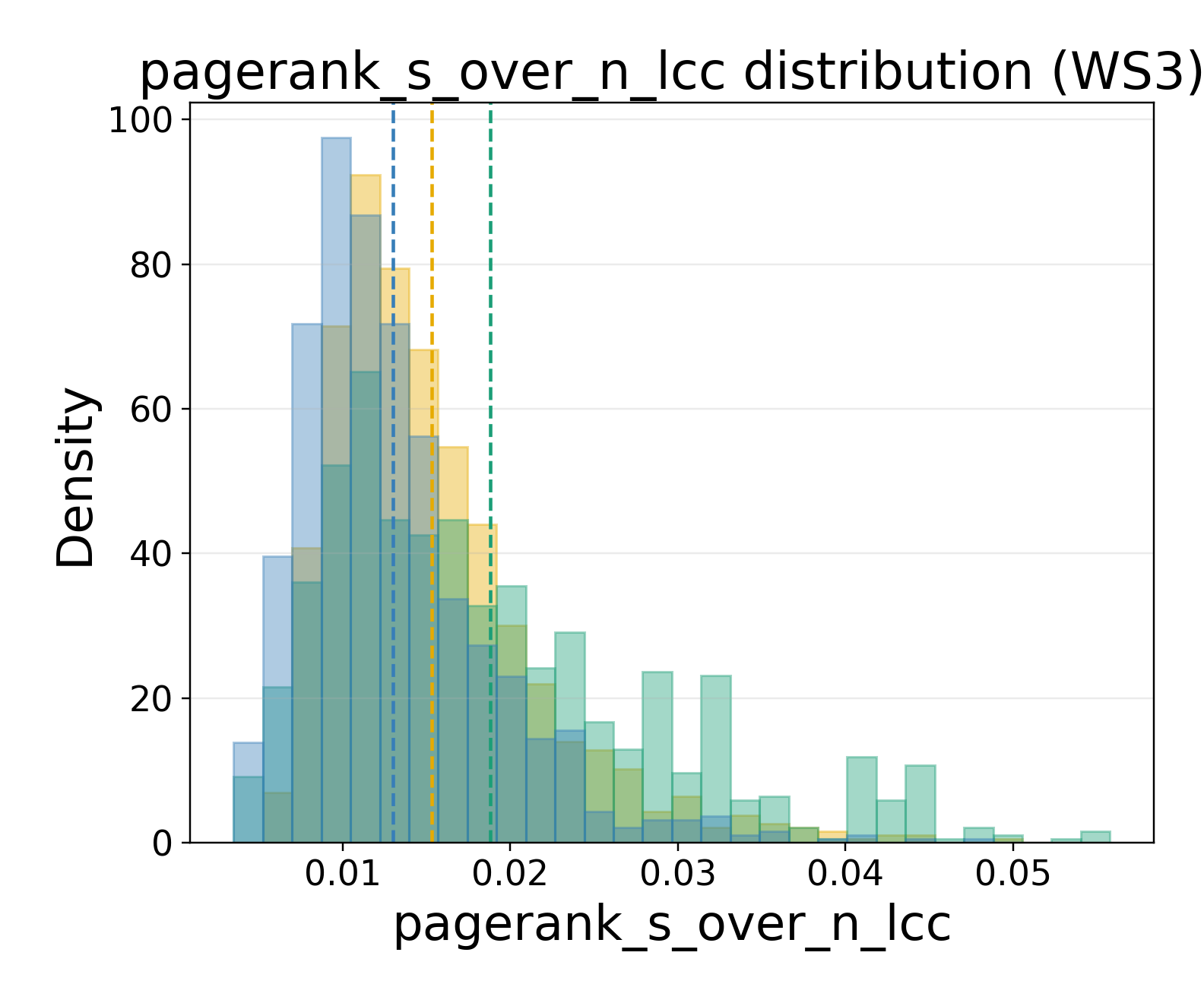}

\caption{
Distributions of key network-structural features for the WS3 co-occurrence setting.
For each feature, distributions are shown for TFMN networks, co-occurrence networks with pronouns retained, and co-occurrence networks with pronouns removed; dashed vertical lines indicate the corresponding mean values.
}
\label{fig:ws3_network_distributions}
\end{figure}

\begin{figure}
\centering

\includegraphics[width=0.40\textwidth]{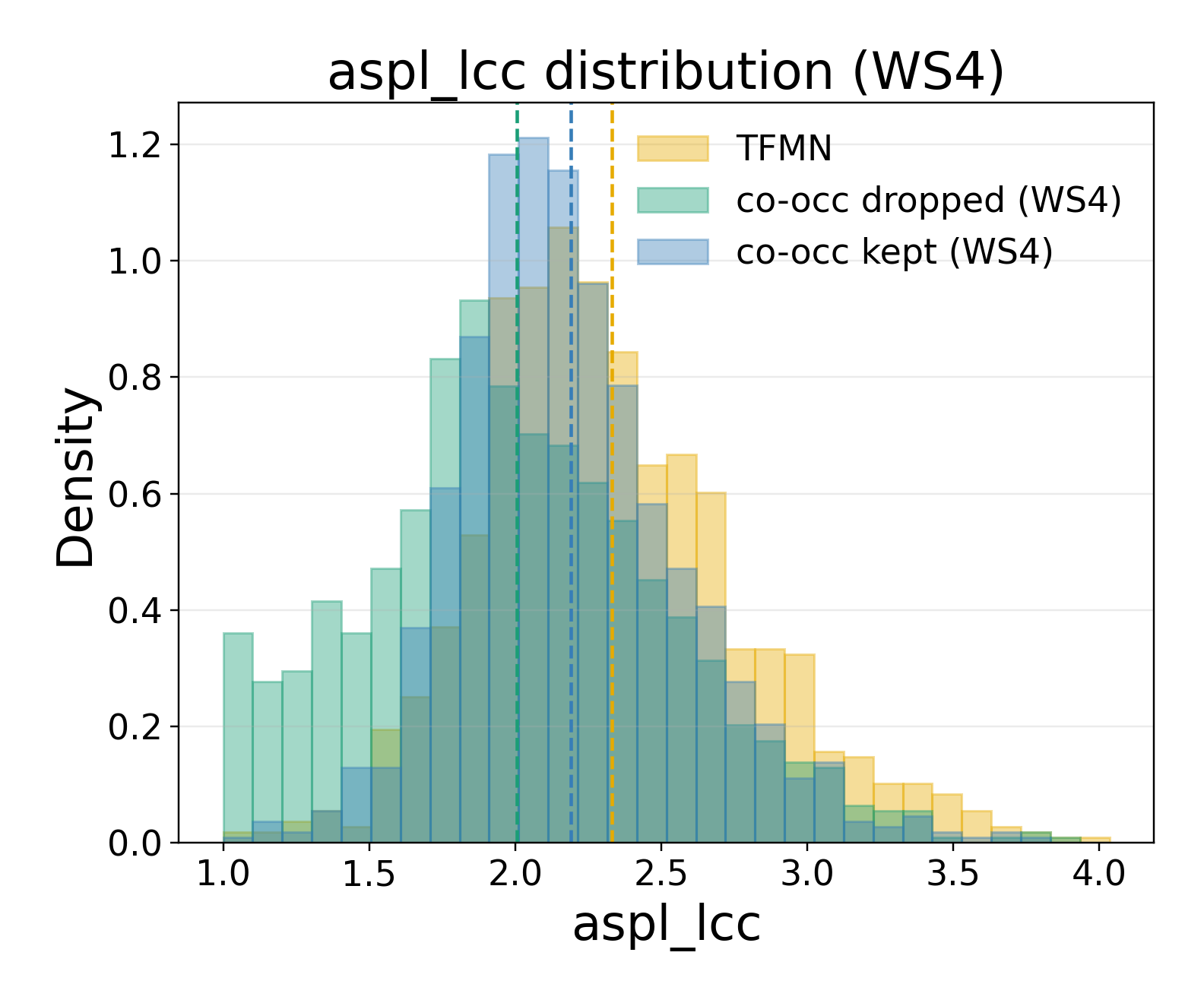}\hfill
\includegraphics[width=0.40\textwidth]{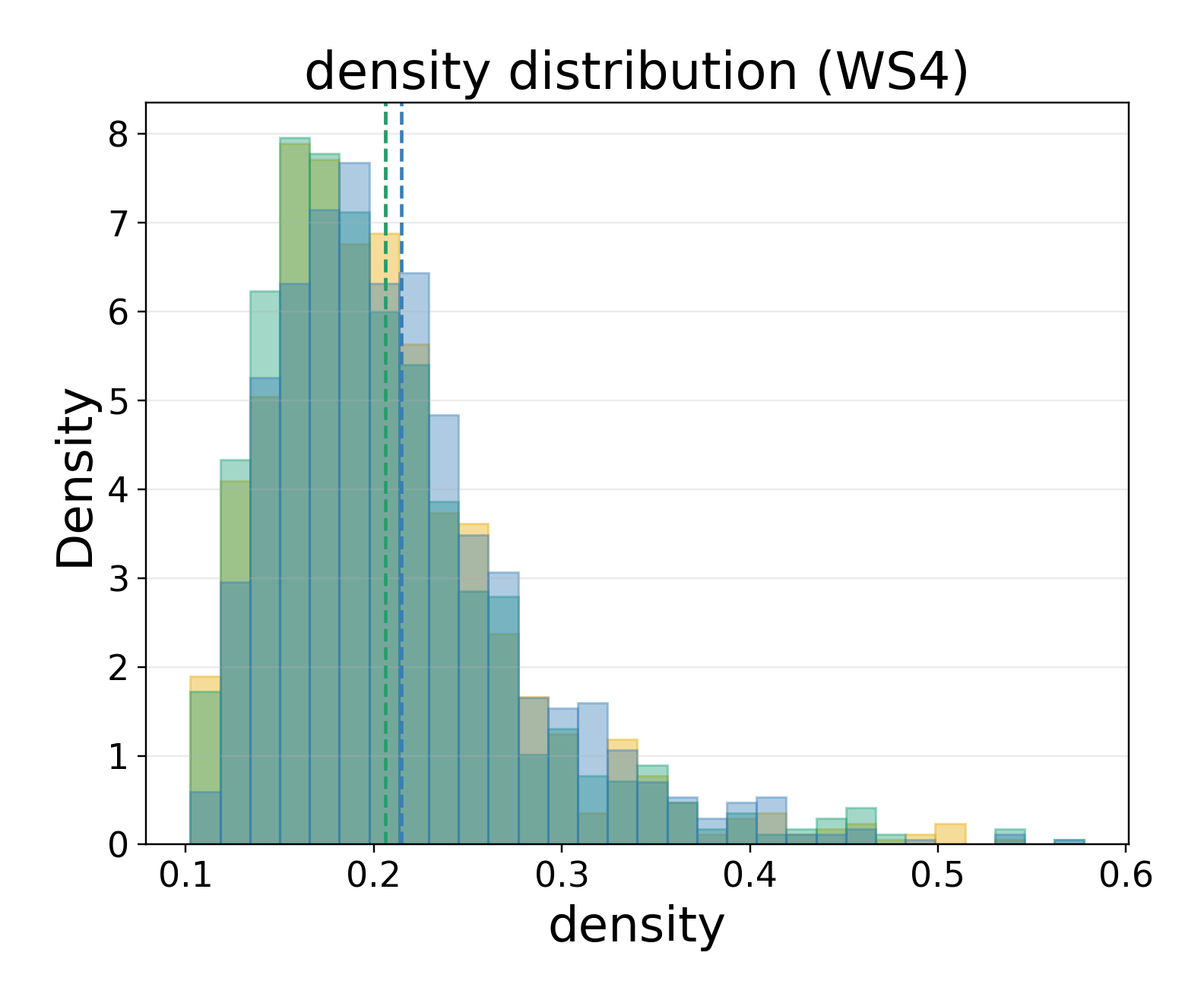}

\includegraphics[width=0.40\textwidth]{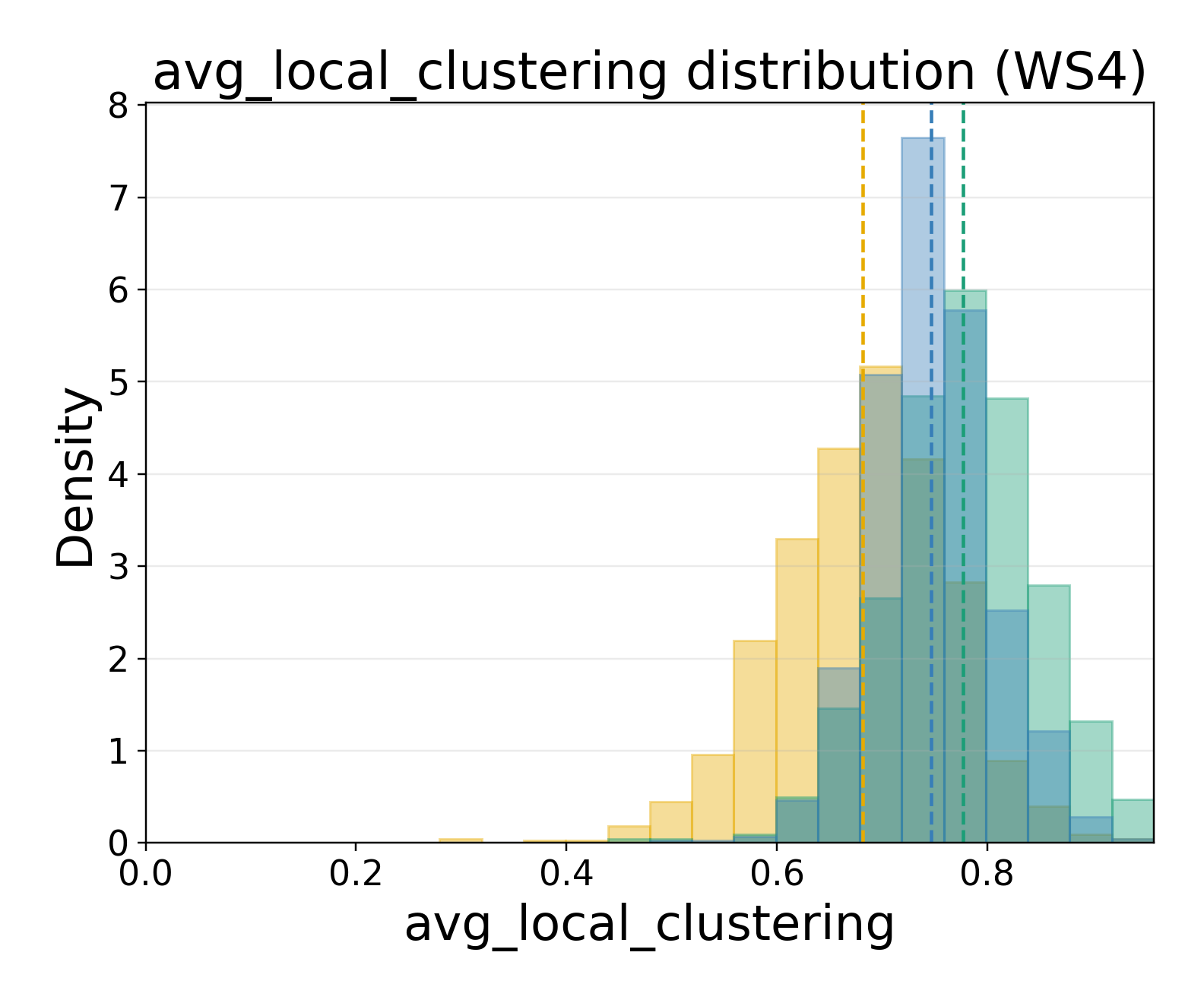}\hfill
\includegraphics[width=0.40\textwidth]{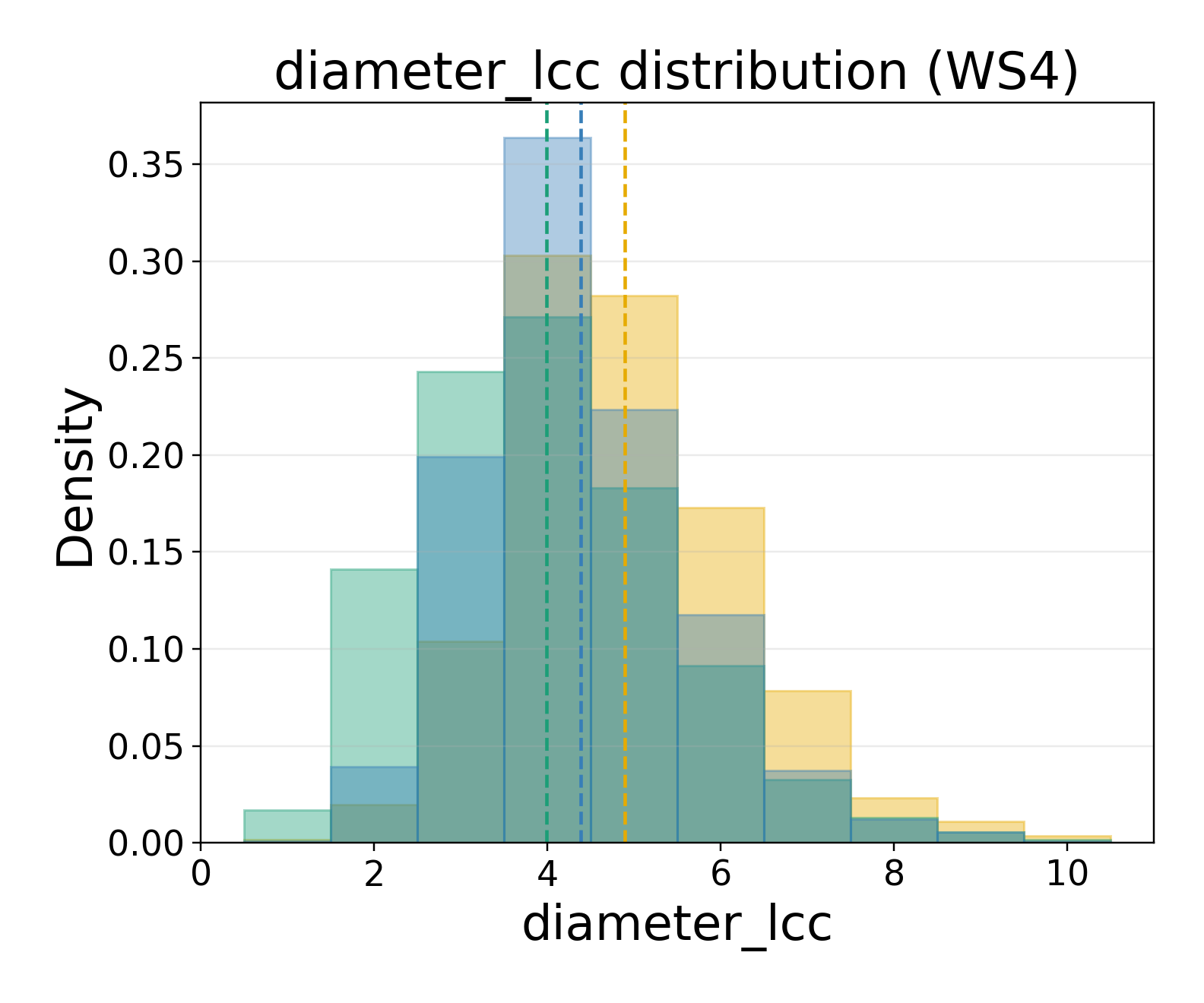}

\includegraphics[width=0.40\textwidth]{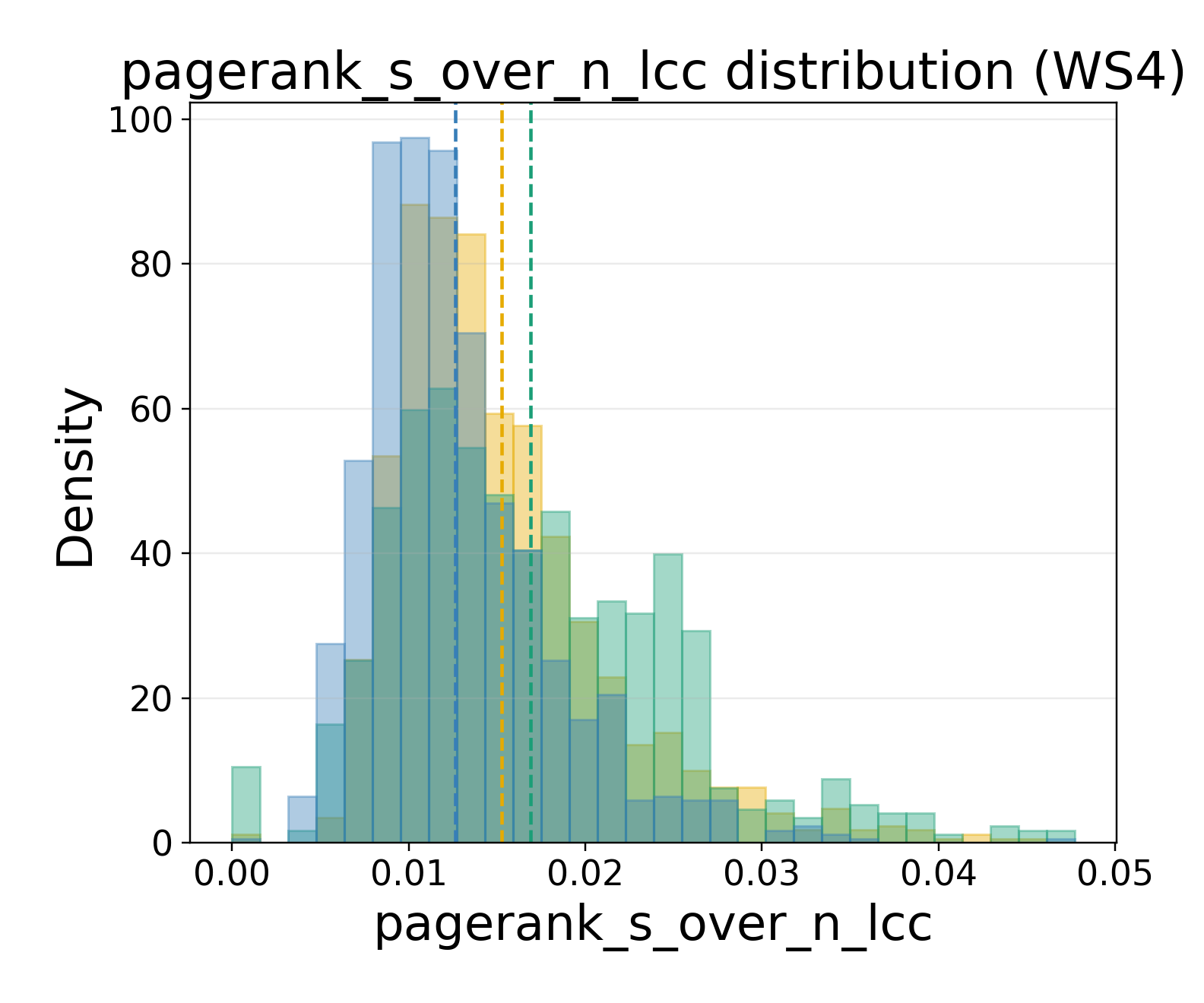}

\caption{
Distributions of key network-structural features for the WS4 co-occurrence setting.
For each feature, distributions are shown for TFMN networks, co-occurrence networks with pronouns retained, and co-occurrence networks with pronouns removed; dashed vertical lines indicate the corresponding mean values.
}
\label{fig:ws4_network_distributions}
\end{figure}

\section{Spreading activation dynamics}

\begin{figure}[H]
\centering
\includegraphics[width=0.40\textwidth]{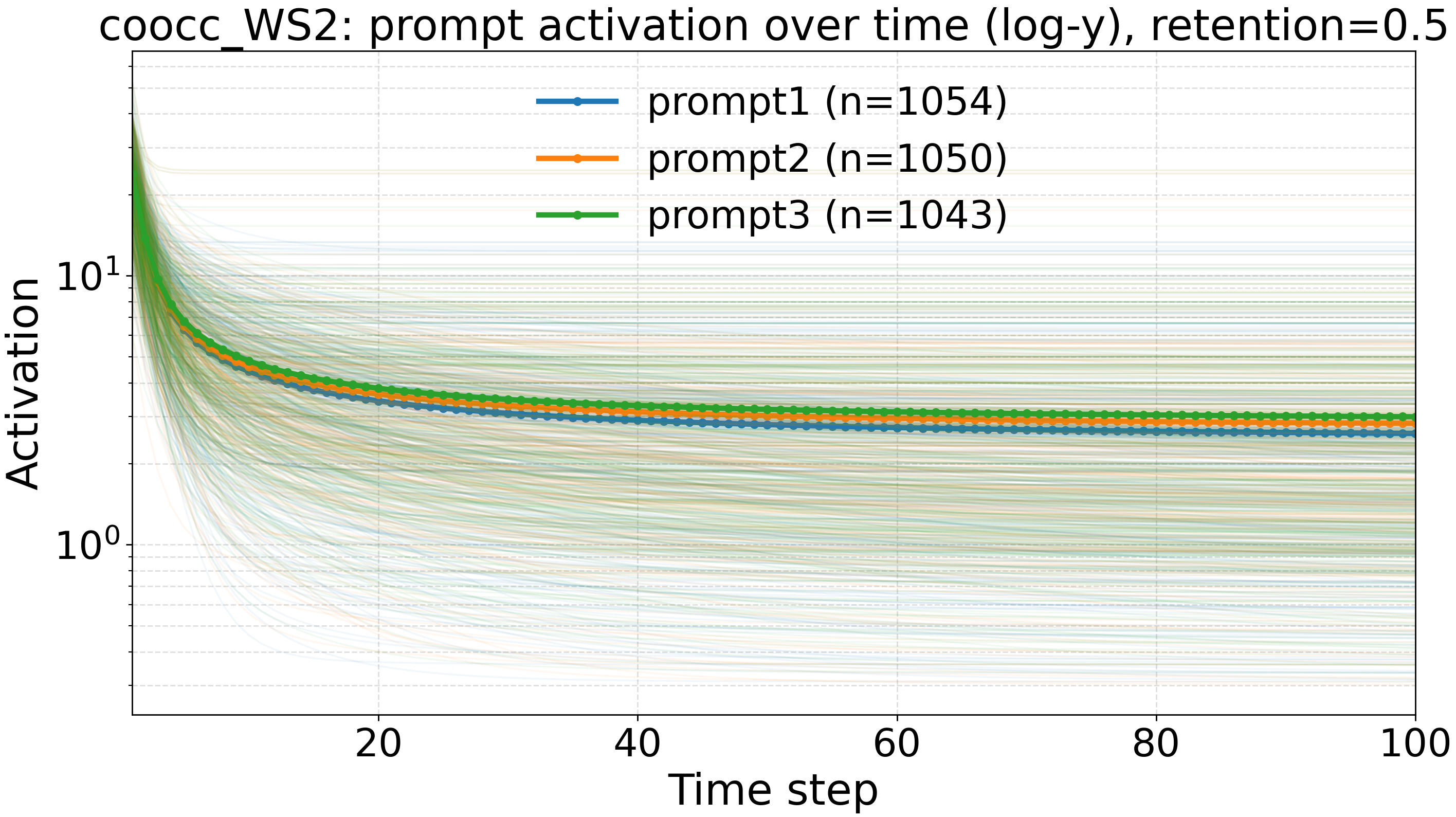}\hfill
\includegraphics[width=0.40\textwidth]{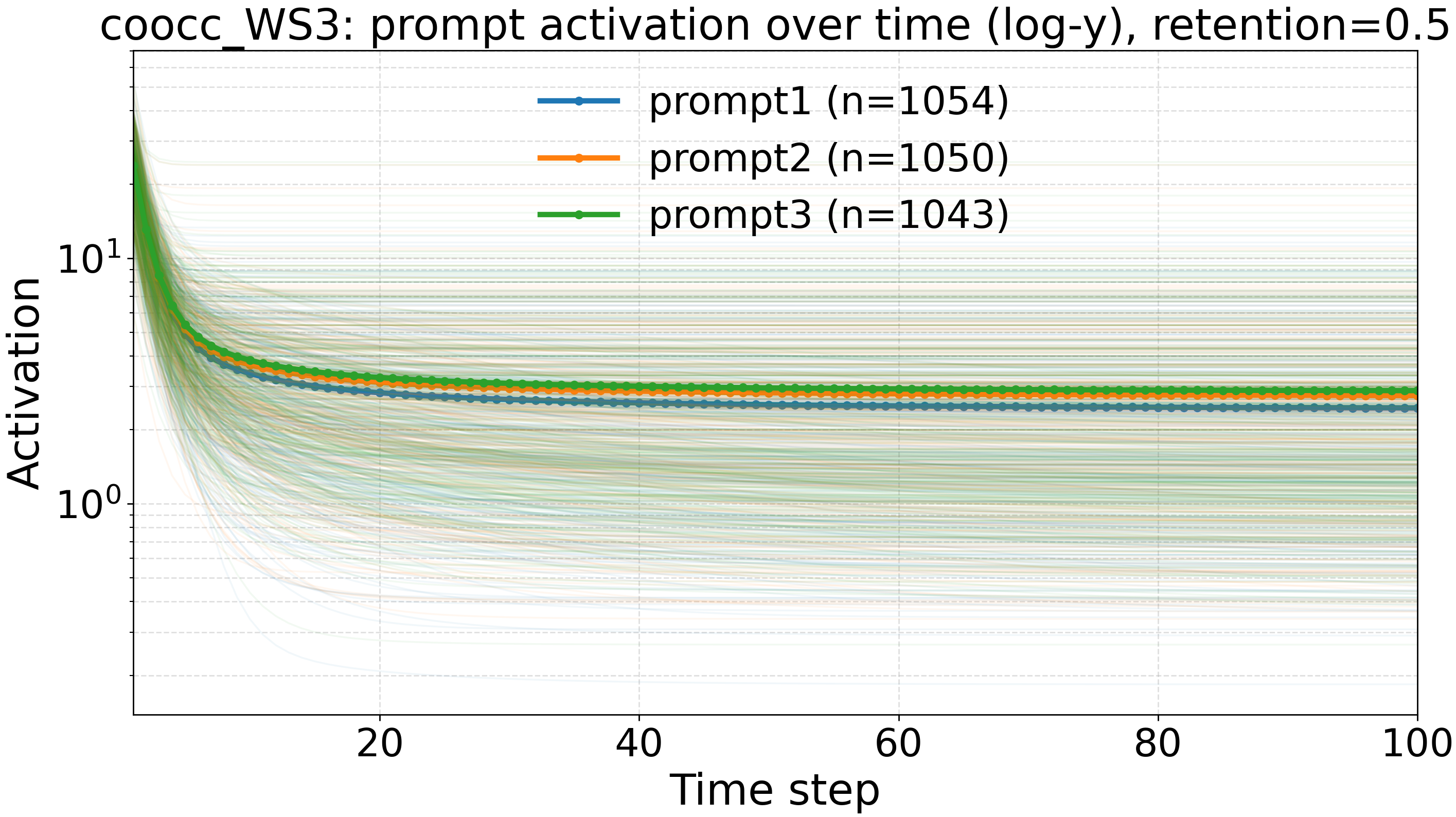}

\includegraphics[width=0.40\textwidth]{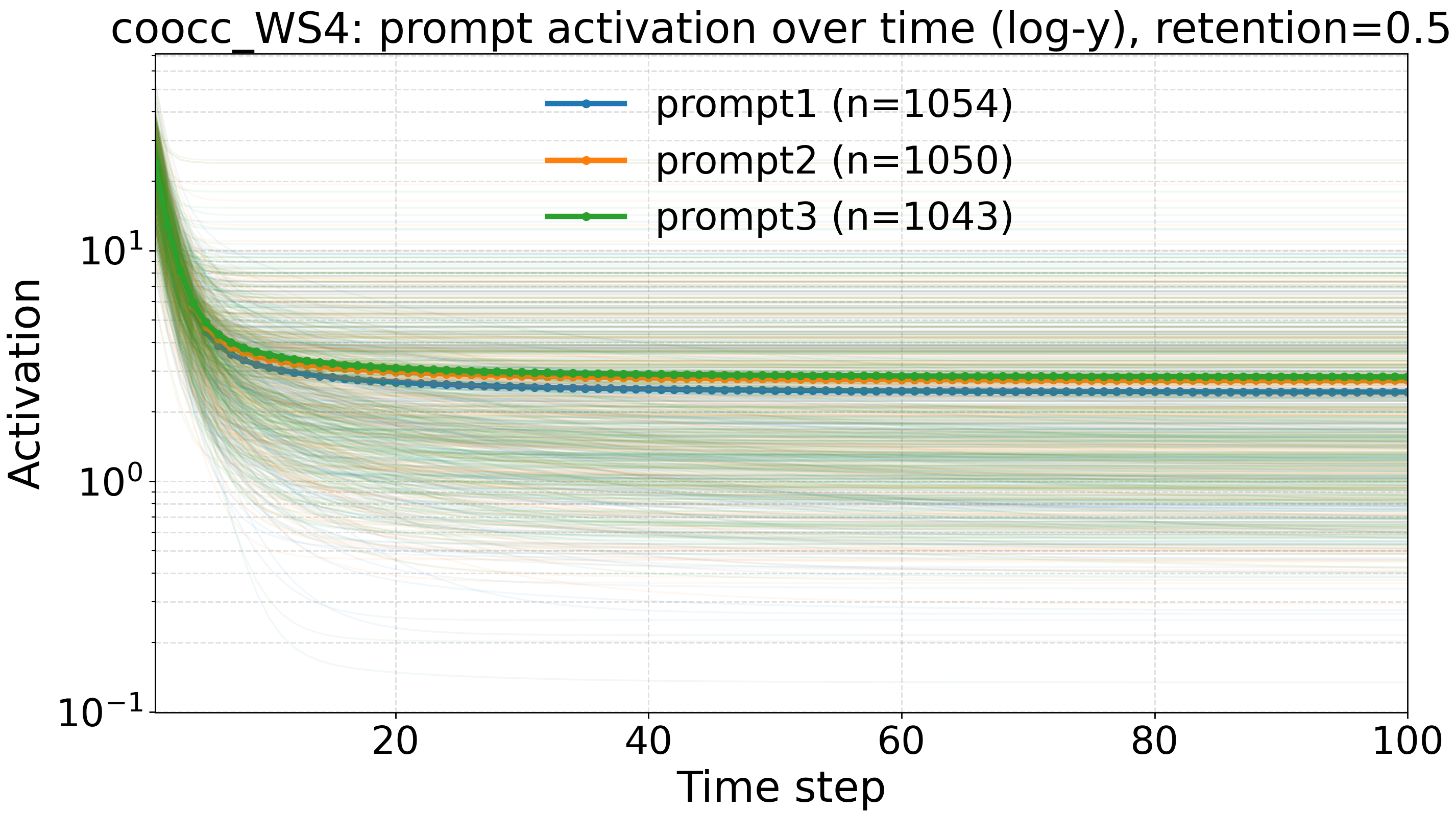}\hfill
\includegraphics[width=0.40\textwidth]{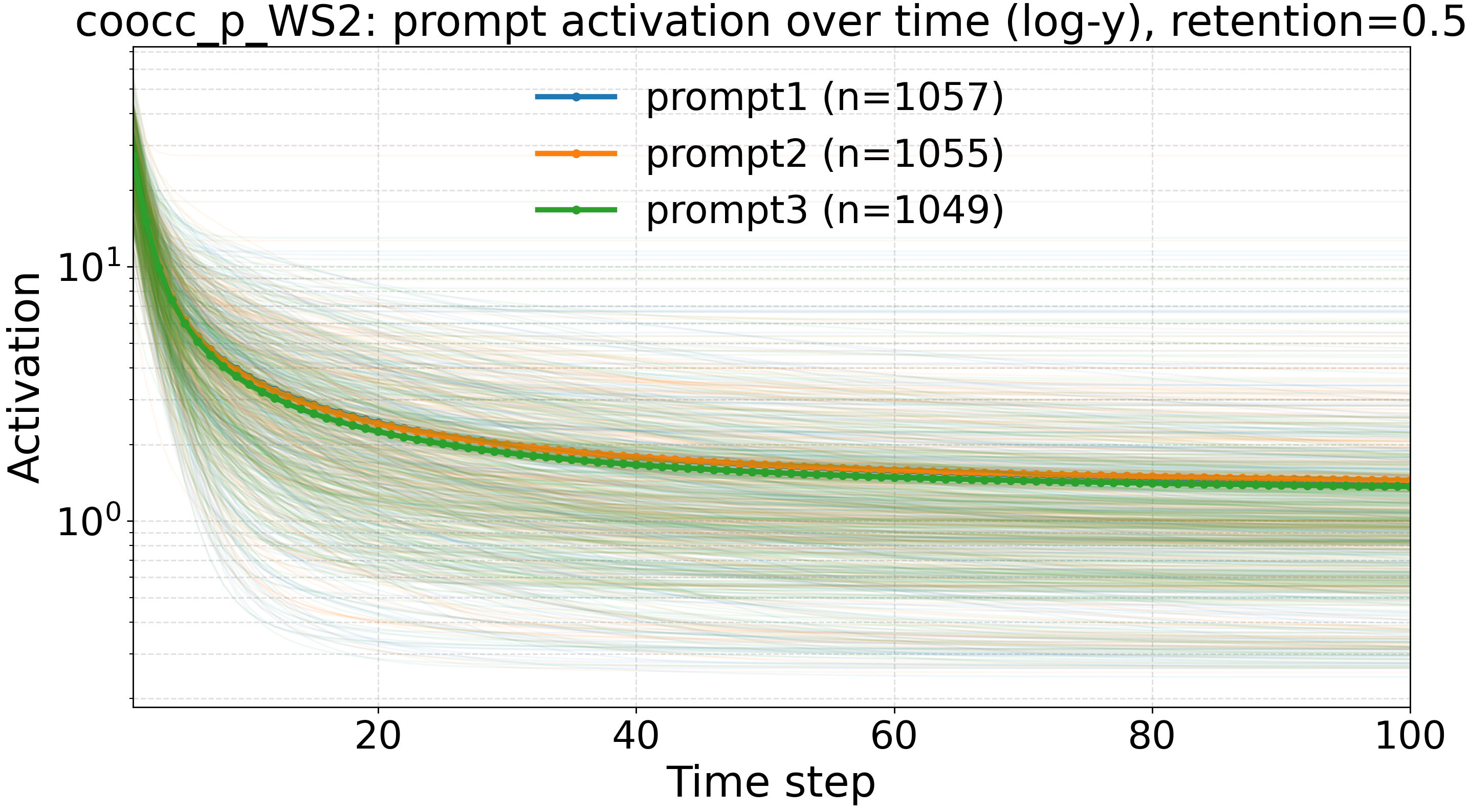}

\includegraphics[width=0.40\textwidth]{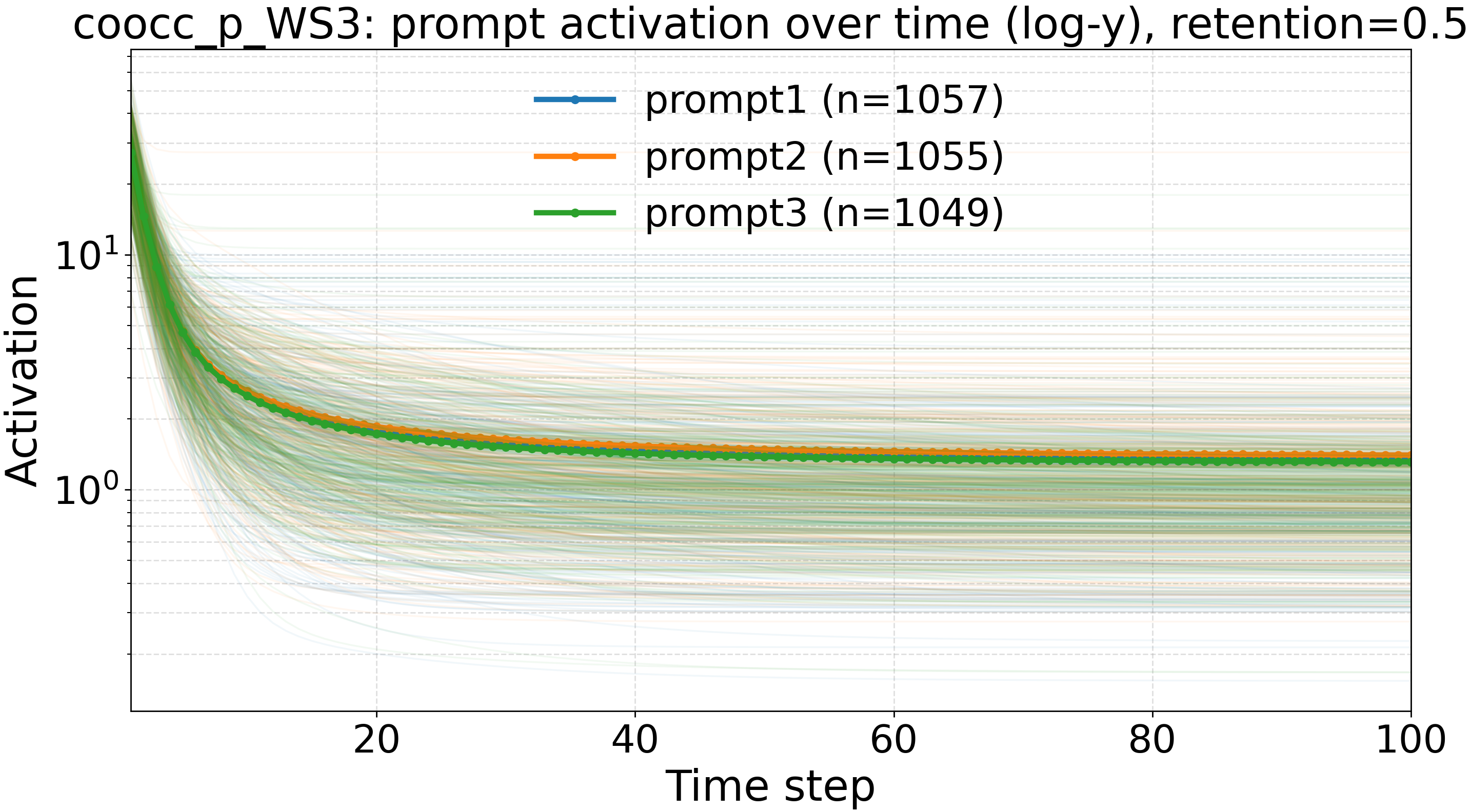}\hfill
\includegraphics[width=0.40\textwidth]{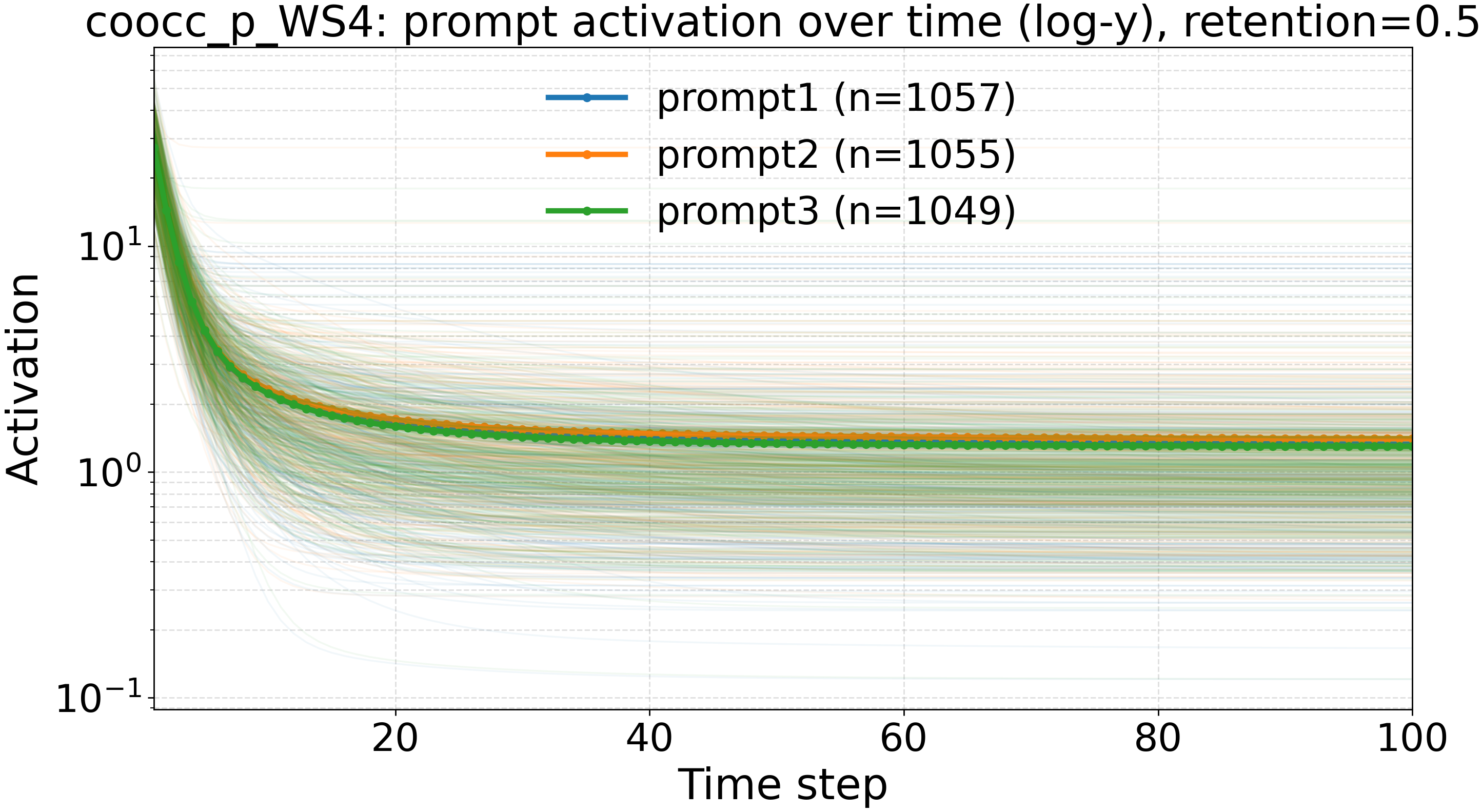}

\includegraphics[width=0.40\textwidth]{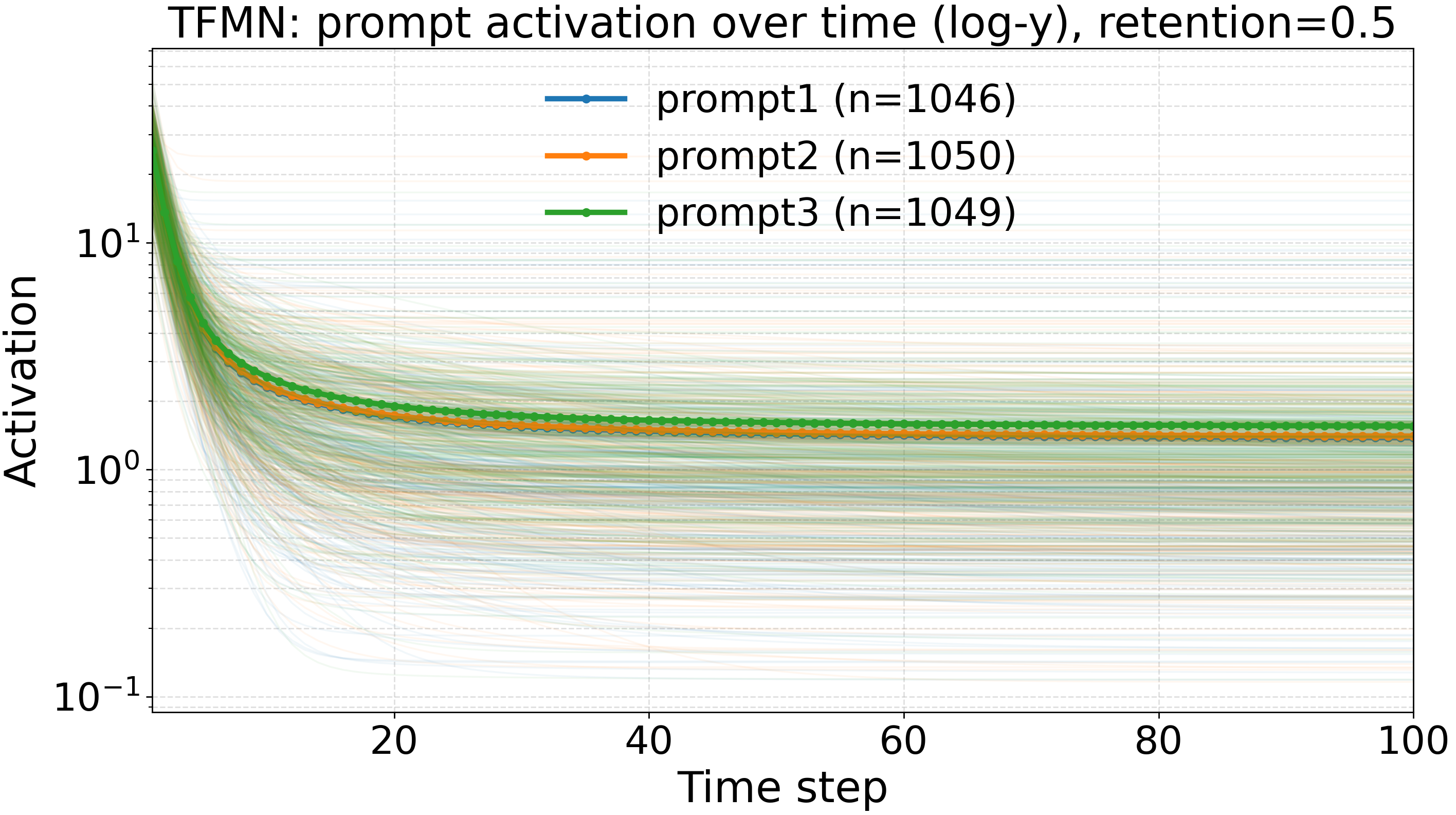}

\caption{Prompt-seeded activation trajectories over time (log-scaled y-axis; retention $=0.5$).
Thin lines denote individual-story trajectories and thick lines denote mean trajectories per prompt word.
Trajectories are shown for a fixed horizon of 100 time steps, whereas reported activation values correspond to stationary seed-node activation, defined by an absolute change $<10^{-9}$ between successive iterations.}
\label{fig:activation_dynamics_all_builders}
\end{figure}

\section{Hyperparameters used for the \textit{All} regression setup}

\begin{table}
\centering
\footnotesize
\begin{tabular}{lp{12cm}}
\toprule
Regressor & Selected hyperparameters \\
\midrule
XGB & \texttt{subsample=0.9, reg\_lambda=1, reg\_alpha=0.1, n\_estimators=600, min\_child\_weight=1, max\_depth=3, learning\_rate=0.01, gamma=0.3, colsample\_bytree=0.6} \\
RF  & \texttt{n\_estimators=500, min\_samples\_leaf=5, max\_features=0.7, max\_depth=None, bootstrap=True} \\
GB  & \texttt{subsample=0.7, n\_estimators=800, min\_samples\_leaf=3, max\_depth=2, learning\_rate=0.01} \\
DT  & \texttt{min\_samples\_leaf=3, min\_impurity\_decrease=0.001, max\_depth=4} \\
KNN & \texttt{weights=distance, p=1, n\_neighbors=15} \\
MLP & \texttt{solver=adam, learning\_rate\_init=0.001, learning\_rate=adaptive, hidden\_layer\_sizes=(256,), batch\_size=128, alpha=0.0003, activation=tanh} \\
BAG & \texttt{n\_estimators=200, max\_samples=0.8, max\_features=0.6, bootstrap\_features=False, bootstrap=True, base\_min\_samples\_leaf=5, base\_max\_depth=12} \\
\bottomrule
\end{tabular}
\caption{Selected hyperparameters for each regression model, obtained via random-search cross-validation (CV4). For each regressor, hyperparameters were tuned independently on each network builder configuration, and the reported setting corresponds to the configuration yielding the best individual performance for that regressor (minimum MAE, with Spearman’s $\rho$ used as a tie-break criterion).}
\label{tab:hyperparams_cv4}
\end{table}

\end{document}